\documentclass[final,5p,authoryear]{elsarticle}

\usepackage{amsmath,amssymb,amsthm,amsfonts,amsxtra,amsthm,mathrsfs}
\usepackage{multirow} 

\usepackage{lmodern}

\usepackage{pifont} 
\usepackage{siunitx}

\usepackage{subcaption}

\usepackage[ruled,lined,boxed,commentsnumbered]{algorithm2e}

\usepackage{tikz}
\usepackage{pgfplots}
\usetikzlibrary{arrows,decorations.pathmorphing,backgrounds,positioning,fit,plotmarks,shadows}
\pgfplotsset{compat=1.3,every axis/.append style={font=\scriptsize}, every legend/.append style={font=\scriptsize}}
\usepackage{color}
\usepackage{xcolor}

\newcommand{\EEqref}[1]{\text{Eq.}~\eqref{#1}}

\usepackage{amsmath}

\DeclareMathOperator*{\argmax}{arg\,max}

\newcommand{\rmd}{\mathrm{d}}  

\newcommand{\uu}{\ensuremath{\mathbf{u}}}

\newcommand{\mathbi}[1]{\ensuremath{\textbf{\textit{#1}}}}
\newcommand{\VV}[1]{\ensuremath{\textbf{\textit{#1}}}}

\newcommand{\Vx}{\mathbi{x}}
\newcommand{\Vy}{\mathbi{y}}

\newcommand{\Vv}{\mathbi{v}}
\newcommand{\Vn}{\mathbi{n}}

\newcommand{\Vw}{\mathbi{w}}

\newcommand{\BM}[1]{\mathbf{#1}}

\usepackage[hyphens]{url}

\usepackage[bookmarks=true, 
            final=true,
            hyperfootnotes=true,
            ]{hyperref}

\usepackage{graphicx}
\usepackage{grffile} 
\graphicspath{{pics/}{figs/}}

\usepackage{stmaryrd}

\usepackage[colorinlistoftodos, textwidth=18mm, shadow]{todonotes}

\theoremstyle{plain}
\newtheorem{definition}{Definition}

\newtheorem{proposition}{Proposition}

\emergencystretch=\hsize
\usepackage{rotating}

\journal{Medical Image Analysis}

\makeatletter
\def\ps@pprintTitle{%
 \let\@oddhead\@empty
 \let\@evenhead\@empty
 \def\@oddfoot{}%
 \let\@evenfoot\@oddfoot}
\makeatother

\begin{document}

\begin{frontmatter}

\title{Director Field Analysis (DFA): Exploring Local White Matter Geometric Structure in Diffusion MRI}

\author[STBB]{Jian Cheng\corref{cor}}
\ead{jian.cheng@nih.gov}

\author[STBB]{Peter J. Basser\corref{cor}}
\ead{pjbasser@helix.nih.gov}

\cortext[cor]{Corresponding author}

\address[STBB]{SQITS, NIBIB, NICHD, National Institutes of Health, United States}

\begin{abstract}

In Diffusion Tensor Imaging (DTI) or High Angular Resolution Diffusion Imaging (HARDI), a tensor field or a spherical function field (e.g., an orientation distribution function field), can be estimated from measured diffusion weighted images. 
In this paper, inspired by the microscopic theoretical treatment of phases in liquid crystals, 
we introduce a novel mathematical framework, called Director Field Analysis (DFA), to study local geometric structural information of white matter based on the reconstructed tensor field or spherical function field:  
1) We propose a set of mathematical tools to process general director data, which consists of dyadic tensors that have orientations but no direction. 
2) We propose Orientational Order (OO) and Orientational Dispersion (OD) indices to describe the degree of alignment and dispersion of a spherical function in a single voxel or in a region, respectively;
3) We also show how to construct a local orthogonal coordinate frame in each voxel exhibiting anisotropic diffusion; 
4) Finally, we define three indices to describe three types of orientational distortion (splay, bend, and twist) in a local spatial neighborhood, and a total distortion index to describe distortions of all three types. 
To our knowledge, this is the first work to \emph{quantitatively} describe orientational distortion (splay, bend, and twist) in general spherical function fields from DTI or HARDI data. 
The proposed DFA and its related mathematical tools can be used to process not only diffusion MRI data but also general director field data, 
and the proposed scalar indices are useful for detecting local geometric changes of white matter for voxel-based or tract-based analysis in both DTI and HARDI acquisitions. 
The related codes and a tutorial for DFA will be released in DMRITool. 

\end{abstract}

\begin{keyword}

Diffusion MRI \sep Diffusion Tensor \sep Orientation Distribution Function \sep Distortion \sep Dispersion \sep Director Field Analysis \sep Local Orthogonal Frame \sep Liquid Crystals \sep dyadic 
\end{keyword}

\end{frontmatter}

\section{Introduction}
\label{sec:1}

Diffusion MRI is a powerful non-invasive imaging technique widely used to explore white matter in the human brain. 
Diffusion Tensor Imaging (DTI)~\citep{Basser1994} is used to reconstruct a tensor field from diffusion weighted images (DWIs). 
High Angular Resolution Diffusion Imaging~\citep{TuchMRM2002,frank_MRM2002,Descoteaux2007,tournier_NI2007,Cheng_PDF_MICCAI2010,cheng_MICCAI2015,cheng_NI2014,ozarslan_NI13}, 
which makes no assumption of a 3D Gaussian distribution of the diffusion propagator, is used to reconstruct a general function field from DWIs, 
(e.g., an Orientation Distribution Function (ODF) or Ensemble Average Propagator (EAP) field). 
Both the ODF and the EAP fields with a given radius are spherical function fields. 
Exploring microstructural information from the reconstructed tensor field or spherical function field is of interest in many biological and clinical application areas, 
which makes diffusion MRI a powerful means to study white matter. 
For example, in an voxel exhibiting anisotropic diffusion, local peaks of the reconstructed spherical function or the principal eigenvector of the reconstructed 2nd-order diffusion tensor normally prescribe the fiber directions in that voxel. 

Some scalar indices have been proposed to be estimated voxel-wise from tensors/ODFs/EAPs. 
For DTI, well-established tensor scalar indices, including the mean diffusivity and Fractional Anisotropy (FA), are widely used as biologically meaningful descriptors~\citep{Pierpaoli1996}. 
\cite{Kindlmann_TMI2007} proposed two sets of scalar indices (three scalar indices per set), which are orthogonal in terms of tensor changes and the Euclidean inner product. 
For High Angular Resolution Diffusion Imaging (HARDI), the generalized FA~\citep{Tuch2004}, Orientation Dispersion index (OD)~\citep{zhang_NODDI_NI2012}, return-to-origin probability~\citep{helmer_MRM2003,Wu_NI2007}, and mean-squared displacement~\citep{basser_MRM2002,Wu_NI2007} 
were all proposed for ODFs and EAPs.
These indices indicate some information inside a voxel, but cannot describe local geometric or topological information, including fiber crossing, fanning, bending, and twisting, in a local spatial neighborhood.

Some previous works have extracted local geometric information by considering the local spatial change of tensor fields or ODF fields. 
\cite{pajevic_JMR2002} demonstrated that the norm of the spatial gradient of the tensor's isotropic and anisotropic parts can detect boundaries between white matter, CSF, and gray matter. 
\cite{Kindlmann_TMI2007} proposed tangents of scalar invariants and rotation tangents, which are 2nd-order tensors, 
and also proposed projecting the 3rd-order spatial gradient tensor onto these 2nd-order tangents to obtain the spatial direction with the largest change of scalar indices or rotation of tensors. 
Based on the rotation tangents of tensors, \cite{Savadjiev_NI2010} proposed fiber curving and fiber dispersion indices. 
\cite{tax_NI2016} proposed a sheet probability index to quantify the local sheet structure by using spatial changes of ODF peaks. 
\cite{duits:2009,duits:IJCV2011,portegies:PLOSOne2015} proposed spatial and spherical smoothing to enhance an ODF field in a PDE framework, preserving crossing structures. 
\cite{reisert_TMI11,JianCheng_NNSD_ISBI13,michailovich:TMI2011} considered spatial coherence in ODF estimation. 

The terms ``splay'', ``bend'' and ``twist'' have been used to qualitatively describe complex local white matter structural configuration in literature of diffusion MRI for about 20 years~\citep{basser_ANYAS1997,pajevic_JMR2002,dMRI_book2009}. 
However, to our knowledge, there is no existing work that \emph{quantitatively} describes the degree of local orientational change of white matter, 
including splay, bend, and twist, from general ODF fields in dMRI, 
although the fiber curving and dispersion indices in~\cite{Savadjiev_NI2010} can be seen to quantify ``splay'' and ``bend'' for a tensor field in DTI.

\cite{basser_ANYAS1997} discussed the initial idea to study the torsion and curvature of a fiber tract by using the Frenet frame~\footnote{\href{https://en.wikipedia.org/wiki/Frenet-Serret_formulas}{https://en.wikipedia.org/wiki/Frenet-Serret\_formulas}} along the tract. 
Torsion and curvature from the Frenet frame were later used in diffusion data analysis in~\cite{batchelor_MRM2006}. 
\cite{savadjiev_ICCV2007} used the Frenet frame as a prior in the relaxation labeling algorithm to regularize the data and estimate ODFs in voxels. 
These works on the Frenet frame studied geometric information along a single tract. 
However, tractography is known to be sensitive to a large number of parameters, 
and any flaws in the reconstructed tracts due to noise or parameter selection will inevitably be reflected in the geometric information that is extracted subsequently. 
\cite{piuze_PAMI2015} proposed moving frames determined by the geometry of cardiac data, and calculated Maurer-Cartan connections. 
However this method is not applicable to general diffusion MRI data, and does not consider the sign ambiguity in the frame.

There exist some connections between diffusion MRI data analysis and liquid crystals. 
Orientational order parameter is well-established to describe the degree of alignment in liquid crystals~\cite{andrienko_2006}. 
\cite{lasivc:FrontiersInPhysics2014,szczepankiewicz:NI2015} calculated the order parameter map by estimating variance of microscopic diffusion parameters from the contrast between diffusion signals measured by directional and isotropic diffusion encoding. 
However, it cannot be used for general DTI and HARDI data.  
\cite{topgaard:PCCP2016} used a diffusion tensor method to estimate the director orientations of a lyotropic liquid crystal as a spatially resolved field of Saupe order tensors. 

\begin{figure*}[t!]
  \centering
\includegraphics[width=1.0\textwidth]{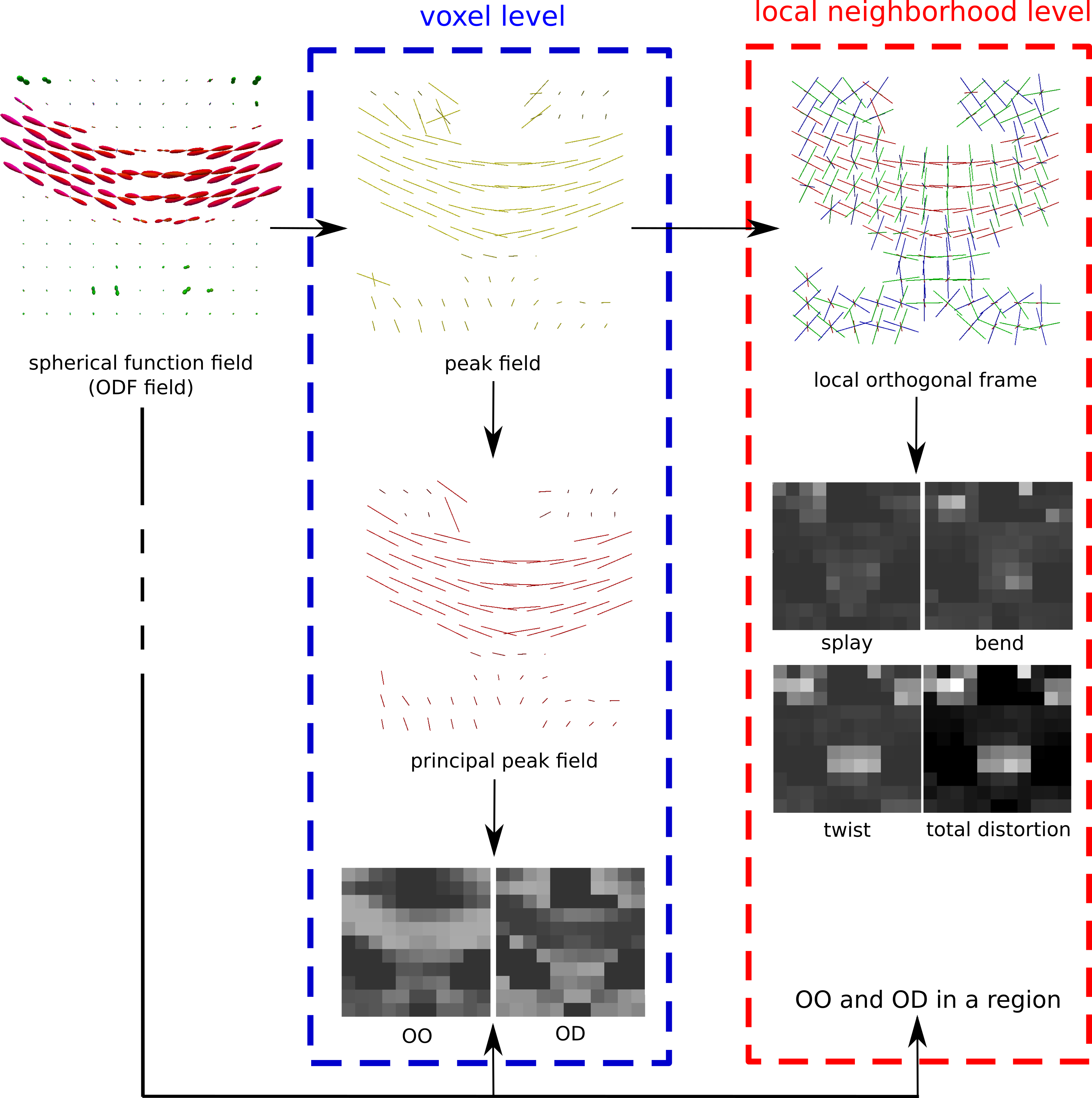}
  \caption{\label{fig:DFA}Director Field Analysis (DFA) pipeline for an ODF field obtained from DTI or HARDI. 
  DFA provides total six scalar indices calculated from a spherical function field at the voxel level and at the local neighborhood level.}
\end{figure*}

In this paper, inspired by orientation and distortion analyses applied to liquid crystals, 
we propose a unified framework, called Director Field Analysis (DFA), to study the local geometric information of white matter from the reconstructed spherical function field. 
DFA works both for tensor fields obtained from DTI and for spherical function fields from HARDI. 
At the voxel level,   
1) the Orientational Order index (OO) and Orientational Dispersion index (OD) are defined for the spherical function in a voxel with a given axis (e.g., the ODF with its principal direction); 
and 2) the principal direction is extracted from the spherical function in such a voxel exhibiting anisotropic diffusion. 
At a local neighborhood level, 
1) an orthogonal coordinate frame is defined for each voxel with anisotropic diffusion, where the first axis is the extracted principal direction;
2) OO is defined for spherical functions in a local neighborhood with the given principal direction; 
and 3) three distortion indices (splay, bend, twist) and a total distortion index are defined based on the spatial directional derivatives of the principal direction. 
An overview of the DFA pipeline for a spherical function field is shown in Fig.~\ref{fig:DFA}. 

This paper is organized as follows. 
Section~\ref{sec:method_tensor} provides a unified overview of existing works on tensor field analysis for exploring local geometric information~\citep{pajevic_JMR2002,Kindlmann_TMI2007,Savadjiev_NI2010}, 
which is also a motivation for the proposed DFA. 
Section~\ref{sec:method} proposes the DFA framework that works for both diffusion tensor fields and ODF fields. 
Section~\ref{sec:experiments} demonstrates some results of synthetic and real data experiments by using DFA. 
Section~\ref{sec:discussion} discusses some issues on implementing DFA. 

\section{Tensor Field Analysis}
\label{sec:method_tensor}

This section provides an overview of existing works exploring the local geometric features of a 2nd-order diffusion tensor field by using the spatial gradient of tensors~\citep{pajevic_JMR2002,Kindlmann_TMI2007,Savadjiev_NI2010} in a unified framework. 
It also proposes a new 4th-order structure tensor applied to 2nd-order tensor data that generalizes the conventional 2nd-order structure tensor applied to scalar fields.

\textbf{The 2nd-order diffusion tensor field}. 
In a diffusion tensor field denoted as $\BM{D}$, there is a diffusion tensor $\BM{D}(\Vx)$ at the voxel $\Vx$, where $\BM{D}(\Vx) \in S_+^3$, 
and $S_+^3$ is the set of $3\times 3$ symmetric positive definite matrices. 
The diffusion tensor is symmetric with six unique (i.e., independent) elements.

\textbf{The 3rd-order spatial gradient of the diffusion tensor}. 
For a tensor field denoted as $\BM{D}$ with elements $[D_{ij}(\Vx)]$ at the voxel $\Vx$, its spatial gradient at voxel $\Vx$, denoted as $\nabla_\Vx \BM{D}(\Vx)$, is a 3rd-order tensor with elements $[D_{ij,k}(\Vx)=\frac{\partial D_{ij}}{\partial x_k}(\Vx)]$, where $i,j,k\in \{1,2,3\}$. 
Since the diffusion tensor is symmetric with six unique elements, the 3rd-order spatial gradient has 18 unique elements. 

\textbf{Mapping the 3rd-order spatial gradient to a vector}. 
Let $\BM{W}=[W_{ij}]$ be a designed 2nd-order weighting tensor in tensor space, 
then the tensor inner product 
\begin{footnotesize}
\begin{equation}\label{eq:tensor2vector}
 \BM{W} : \nabla_\Vx \BM{D} =\sum_{ij} W_{ij} D_{ij,k} = \sum_{ij} W_{ij} \frac{\partial D_{ij}}{\partial x_k} = \frac{\partial \sum_{ij}W_{ij} D_{ij}}{\partial x_k} 
\end{equation}
\end{footnotesize}%
produces a vector in the image space that is the spatial gradient of the scalar field $\sum_{ij}W_{ij} D_{ij}$ at the voxel $\Vx$. 
Note that the inner product in~\EEqref{eq:tensor2vector} is performed at voxel $\Vx$, and the $\Vx$ dependency is omitted in the notation if there is no ambiguity. 
There are several ways to design a physically meaningful weighting tensor, $\BM{W}$. 
$\BM{W}$ could be a constant independent of spatial position, $\Vx$, or a function of $\Vx$. 
1) If $\BM{W}=\frac{1}{3} \BM{I}$, then $\sum_{ij}W_{ij} D_{ij}$ is the mean diffusivity field, and~\EEqref{eq:tensor2vector} is its spatial gradient. 
2) If $f: \BM{D}\in S_+^3 \mapsto f(\BM{D})\in \mathbb{R}^1 $ is a scalar function that maps $\BM{D}$ to a scalar value, then $\frac{\partial f(\BM{D})}{\partial \BM{D}}$ is the gradient of the scalar function, which is also a 2nd-order tensor with elements $[\frac{\partial f(\BM{D})}{\partial D_{ij}}]$. 
If we set $\BM{W}=\frac{\partial f(\BM{D})}{\partial \BM{D}}$, then the vector in~\EEqref{eq:tensor2vector} is the spatial gradient of the scalar field $f(\BM{D}(\Vx))$, 
because of $\frac{\partial f}{ \partial \Vx} = \frac{\partial f}{\partial \BM{D}} : \frac{\partial \BM{D}}{\partial \Vx}$  by the chain rule. 
If $f(\BM{D})$ is the mean diffusivity function, then $\frac{\partial f}{\partial \BM{D}}=\frac{1}{3}\BM{I}$. 
We can also use other scalar invariants of tensors, e.g., FA. 
3) If we choose $\BM{W}$ as the rotation tangent $\Phi_p(\BM{D})$~\citep{Kindlmann_TMI2007} defined as the change of tensor value due to infinitesimal rotations
around the $p$-th eigenvector, then~\EEqref{eq:tensor2vector} denotes the direction in which the tensor orientation
around the $p$-th eigenvector varies the fastest.

\textbf{Mapping the 3rd-order spatial gradient to a scalar value}. 
Let $\BM{W}=[W_{ij}]$ be a 2nd-order weighting tensor in tensor space, and let $\Vv$ be a vector, 
then the tensor inner product 
\begin{align}
  \BM{W} : \nabla_\Vx \BM{D} : \Vv &= \sum_{ijk} W_{ij}v_k \frac{\partial D_{ij}}{\partial x_k}  \nonumber \\
  &= \frac{\partial \sum_{ij}W_{ij} D_{ij}}{\partial \Vv} = \sum_{ij} W_{ij}  \frac{\partial D_{ij}}{\partial \Vv} \label{eq:tensor2scalar}
\end{align}
produces a scalar value 
that is the directional derivative of the scalar field $\sum_{ij}W_{ij} D_{ij}$ at voxel $\Vx$ along the vector $\Vv$, 
and is also the weighted mean of the directional derivative of $\BM{D}(\Vx)$ along the vector $\Vv$. 
1) If we set $W_{ij}v_k = D_{ij,k}$, then~\EEqref{eq:tensor2scalar} is the squared norm of the tensor gradient, 
which is useful for detecting boundaries of a tensor field~\citep{pajevic_JMR2002}. 
2) By choosing $\Vv$ as the three eigenvectors of $\BM{D}$, and $\BM{W}$ as three rotation tangents around three eigenvectors, respectively, 
we have total 9 scalar values to distinguish 9 configurations of tensor fields~\citep{Savadjiev_NI2010}.
3) The above 9 scalar indices can be combined to devise the fiber curving and fiber dispersion indices~\citep{Savadjiev_NI2010}.

\textbf{The 4th-order structure tensor}. 
We propose a new 4th-order structure tensor with elements 
$D_{ij,kl}= \frac{\partial D_{ij}}{\partial x_k} \frac{\partial D_{ij}}{\partial x_l}$, 
which is analogous, but generalizes the structure tensor of a scalar field~\footnote{\href{https://en.wikipedia.org/wiki/Structure_tensor}{https://en.wikipedia.org/wiki/Structure\_tensor}}. 
The above 4th-order structure tensor is minor symmetric~\citep{moakher_2009}, i.e., $D_{ij,kl}=D_{ji,kl}$, $D_{ij,kl}=D_{ij,lk}$. 
Thus, there are 36 unique elements out of a possible total of $81$ elements, and there is one-to-one mapping between this 4th-order tensor and a  2nd-order tensor (i.e., a $6\times 6$ matrix). 
However, since the 4th-order tensor is minor symmetric, the corresponding $6\times 6$ matrix is not symmetric in general. 
Thus, eigenvalues and the 2nd-order left and right eigenvectors can be calculated based on eigen-decomposition of the non-symmetric $6\times 6$ matrix. 
We may define some scalar invariants from these six eigenvalues of the 4th-order structure tensor,  which can be used as features in this high dimensional space. 
We can also contract the 4th-order structure tensor to a scalar value by using the tensor inner product
$\sum_{ijkl} W_{ij} D_{ij,kl} v_kv_l$, 
which is the weighed mean of the squared spatial directional derivative along vector $\Vv$.
When setting $\Vv$ as three eigenvectors and corresponding weighting $\BM{W}$ as rotation tangents divided by the spatial gradient, 
then the tensor inner product produces nine scalar indices that are the squares of the corresponding nine indices in~\cite{Savadjiev_NI2010}. 
Thus, the curving and dispersion indices in~\cite{Savadjiev_NI2010} can also be obtained by choosing the corresponding $\BM{W}$.

\section{Method: Director Field Analysis}
\label{sec:method}

Section~\ref{sec:method_tensor} provides a unified framework to explore geometric structure information (e.g., boundary, curving, dispersion, etc.) from a tensor field, by considering a different weighting matrix on the spatial gradients. 
However, it is challenging to generalize this framework to ODFs in HARDI, where ODFs are normally general spherical functions with antipodal symmetry. 

In this section, we propose a novel mathematical framework, called Director Field Analysis (DFA). 
Section~\ref{sec:dir} defines director related concepts to deal with vectors with sign ambiguity. 
Section~\ref{sec:math_dir} provides a set of mathematical tools for DFA. 
Section~\ref{sec:order} proposes OO and OD for tensors and ODFs in voxels and in a spatial neighborhood, and gives closed-form results in some specific cases. 
Section~\ref{sec:frame} extracts the principal direction and its related local orthogonal frame in voxels exhibiting anisotropic diffusion. 
Section~\ref{sec:distortion} defines four orientational distortion indices and demonstrates the implementation of the calculation by using the local orthogonal frame in Section~\ref{sec:frame}. 
Section~\ref{sec:dir} and~\ref{sec:math_dir} are the theory part of DFA. 
Section~\ref{sec:order}, \ref{sec:frame}, and~\ref{sec:distortion} are the application part of DFA in diffusion MRI. 
Fig.~\ref{fig:DFA} demonstrates DFA to a spherical function field obtained from DTI or HARDI. 

\subsection{Director and Director Field}
\label{sec:dir}

\begin{figure*}[t!]
    \begin{tabular}{c@{\hskip 0.1in} c@{\hskip 0.1in} c }
      \includegraphics[scale=2,draft=false]{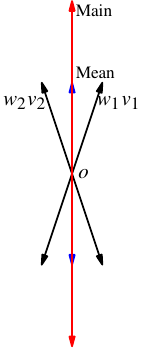} &
      \includegraphics[scale=.26,draft=false]{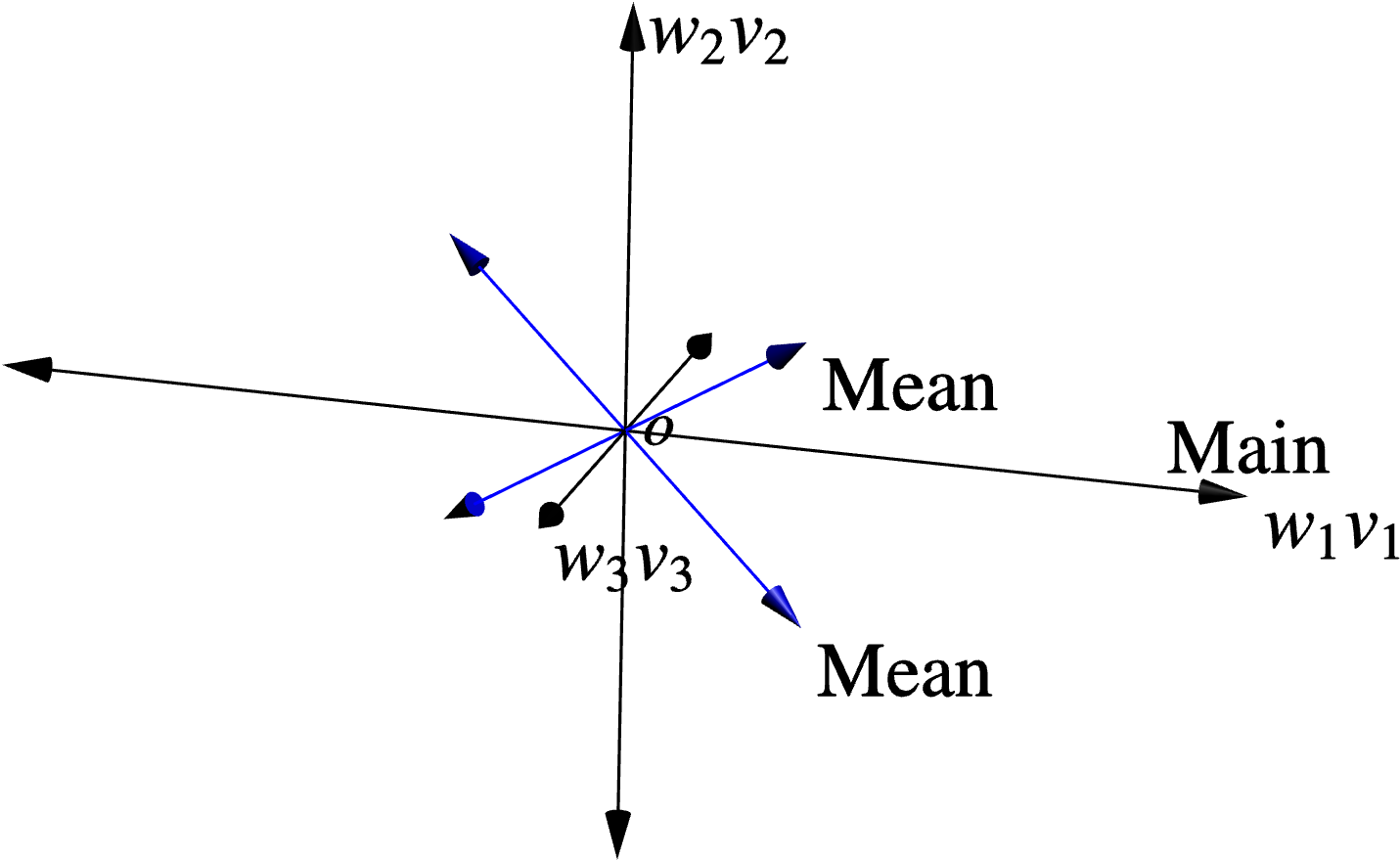} &
\includegraphics[scale=2,draft=false]{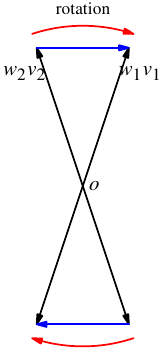} \\
      (a) & (b)  & (c)
\end{tabular}
  \caption{\label{fig:directors}The mean, main, and difference of directors. 
  Directors $(\Vv_i, w_i)$, $i=1,2,3$, are visualized as vectors $w_i\Vv_i$ and $-w_i\Vv_i$, and the length of $w_i\Vv_i$ is the positive weight $w_i$. 
  (a) the mean (in blue) and main (in red) directors of two directors (in black), where $w_1=w_2$, and $\Vv_1^T\Vv_2>0$. See Proposition~\ref{prop:twodir}. 
  (b) the mean (in blue) and main directors of three directors (in black) which are orthogonal to each other, where $w_1>w_2>w_3$, $\Vv_i^T\Vv_j=\delta_i^j$. 
  The mean director is not unique, because $\frac{1}{3}\sum_i s_iw_i\Vv_i$ with an arbitrary sign assignment $\{s_i\}$ can be the mean director. 
  The main director is $w_1\Vv_1$. 
  (c) the difference of two directors. 
  The two blue vectors denote the director representation of the difference, i.e., $\texttt{Diff}_\text{d}(w_1\Vv_1, w_2\Vv_2)=w_1\Vv_1-w_2\Vv_2$. 
  The two red arcs denote the rotation matrix representation of the difference, i.e., $\texttt{Diff}_\text{r}(w_1\Vv_1, w_2\Vv_2)= \BM{R}$, where $\BM{R}$ is a scaled rotation matrix such that $w_1\Vv_1=\BM{R}w_2\Vv_2$. 
  The rotation matrix representation has no sign ambiguity, while the director representation has the sign ambiguity.} 
\end{figure*}

We define a \emph{director} as a unit norm vector $\Vv$ that is equivalent to $-\Vv$.  
The \emph{director} term is borrowed from studies of liquid crystals~\footnote{\label{nt:lq}\href{https://en.wikipedia.org/wiki/Liquid_crystal}{https://en.wikipedia.org/wiki/Liquid\_crystal}}. 
We also define a \emph{director with weight}, or \emph{weighted directors}, as a vector associated with a weight $(\Vv, w)$, which is equivalent to $(-\Vv, w)$, where $\|\Vv\|=1$, $w\in \mathbb{R}^1$. 
If $w\geq 0$, then a weighted director $(\Vv,w)$ can be represented as $w\Vv$. See Fig.~\ref{fig:directors}. 
A director $\Vv$ can be uniquely represented as a dyadic tensor, $\Vv\Vv^T$, which avoids the sign ambiguity, 
and a director with weight $(\Vv,w)$ can be uniquely represented as a dyadic tensor, $w\Vv \Vv^T$. 
We define a \emph{director field} as $\{(\uu_i(\Vx), w_i(\Vx))\}$, where there are some weighted directors in each voxel $\Vx$.

Directors occur very often in diffusion MRI studies. 
Eigenvectors of diffusion tensors, local maxima of ODFs, and local fiber directions are all directors. 
Based on eigen-decomposition, a tensor $\BM{D}=\sum_{i=1}^3 \lambda_i \Vv_i\Vv_i^T$ is the sum of three dyadic tensors that represent three weighted directors. 
A spherical function $f(\uu)$ which satisfies antipodal symmetry, i.e., $f(\uu)=f(-\uu)$,  can be seen as infinite weighted directors $\{(\uu_i, f(\uu_i))\}$. 
Thus, a spherical function field, $f(\uu,\Vx)$, is a director field by definition.  

An ODF in a voxel exhibiting anisotropic diffusion is anisotropic, and the orientations where the ODF takes its local peak (i.e., local maximal values) are normally considered to be local fiber directions in that voxel. 
A normal peak detection algorithm for ODFs performs a grid search in a spherical mesh, and then refines the solution by using a gradient ascent on the continuous sphere~\citep{TournierNI2004}. 
Note that peak detection is only performed for voxels exhibiting anisotropic diffusion (e.g., where ODFs have Generalized FA (GFA)~\citep{Tuch2004} values larger than $0.3$).  
Moreover, in order to avoid including small peaks produced by noise, only peaks whose values are larger than a threshold percentage (e.g., $0.5$) of the largest ODF value are counted. 
After peak detection, for each voxel $\Vx$, we obtain a discrete spherical function $g(\uu,\Vx)=\sum_{i}f(\uu_i,\Vx)\delta(\uu-\uu_i)$ from the continuous spherical function $f(\uu,\Vx)$, where $\{\uu_i\}$ are local peaks. 
This discrete spherical function field is also a \emph{director field}, or called a \emph{peak field}, which emphasizes local peaks and suppresses weights for other directors. 
A peak field can also be extracted from a tensor field. 
In each voxel for a tensor field, there is 0 or 1 peak, 
and the principal eigenvector of the tensor is considered as a peak, if the tensor has a large FA value (e.g., larger than $0.3$).

\subsection{Mathematical Tools for Directors}
\label{sec:math_dir}

We provide a set of mathematical tools for analyzing directors and director fields. 
These tools are useful not only for this paper, but also for other applications which deal with continuous or discrete director data.

\subsubsection{Mean Director of a Set of Directors}

For a given $N$ weighted directors $\{(\Vv_i, w_i)\}_{i=1}^N$, if we convert a director to a vector by assigning a sign, we have total $2^N$ possible sign assignments. 
Thus, we have $2^N$ possible Euclidean mean vectors for the $N$ vectors. 
We define a \emph{mean weighted director} of a set of weighted directors as the Euclidean mean vector with the sign assignment that takes the maximal norm among the $2^N$ mean vectors. 

\begin{definition}{Mean director of weighted directors.}\label{def:mean_dir}
  A mean weighted director of a set of weighted directors $\{(\Vv_i, w_i) \}_{i=1}^N$ is defined as $\texttt{Mean}(\{(\Vv_i, w_i) \}_{i=1}^N)=\frac{1}{N}\sum_{i=1}^N w_i s_i \Vv_i$, 
  where the signs $\{s_i\}= \argmax_{s_i=\{1,-1\}} \|\sum_{i=1}^N w_i s_i \Vv_i\|^2$, and $\texttt{Mean}(\cdot)$ is the mean director operator. 
\end{definition}

It is obvious that $\{(\Vv_i, w_i)\}_{i=1}^N$ and $\{(\Vv_i, |w_i|)\}_{i=1}^N$ have the same mean director. 
Thus, without loss of generality, we assume non-negative weights for calculating the mean director. 
If the angle between any two vectors $\Vv_i$ and $\Vv_j$ is no more than $90^\circ$, then the sign assignment for the mean director can be proved to be $s_i=1$, $\forall i$.  
See Proposition~\ref{prop:mean_cone_dir} whose proof is based on the proof of the mean director of two directors, which is trivial. 
The mean director may be not unique when some directors are orthogonal. 

\begin{proposition}{Mean director of weighted directors in a $90^\circ$ cone.}\label{prop:mean_cone_dir}
  For a set of weighted directors $\{(\Vv_i, w_i)\}_{i=1}^N$ with non-negative weights, 
  if all directors are in a $90^\circ$ cone, i.e., $\Vv_i^T\Vv_j \geq 0 $, $\forall i, j$, 
  then the mean weighted director is $\frac{1}{N}\sum_{i=1}^N w_i \Vv_i$. 
\end{proposition}

\subsubsection{Main Director of a Set of Directors} 

We define the \emph{main director} of a set of weighted directors as the main axis in Principal Component Analysis (PCA) by using eigen-decomposition. 
This concept is from the average director of molecule orientations in liquid crystals. 

\begin{definition}{Main director of weighted directors.}\label{def:main_dir}
  A \emph{main weighted director} of a set of weighted directors $\{(\Vv_i, w_i) \}_{i=1}^N$ is defined as $ \texttt{Main}(\{(\Vv_i, w_i) \}_{i=1}^N) = (\Vv_0, \lambda_0)$, 
  where $\lambda_0$ is the eigenvalue of the tensor $\sum_{i=1}^N w_i \Vv_i\Vv_i^T$ which has the largest absolute value among all eigenvalues, and $\Vv_0$ is its corresponding eigenvector, and $\texttt{Main}(\cdot)$ is the main director operator. 
\end{definition}

Note 
1) The dyadic tensor of the largest eigenvalue and eigenvector $\lambda_0 \Vv_0\Vv_0^T $ is the best rank-1 approximation of $\sum_{i=1}^N w_i \Vv_i\Vv_i^T$ in terms of the L2 norm. 
2) The main director may not be unique, considering there may be more than one eigenvalues which are equal, and are all the largest eigenvalue. 
3) Unlike the mean director which is independent of the signs of the weights, the main director is dependent on the weight signs. 
4) Although, in general, the mean and the main directors are not the same, in some cases, they may give the same direction. 
See Proposition~\ref{prop:twodir} and Fig.~\ref{fig:directors} (a) for the two director case. 
See Fig.~\ref{fig:directors} (a) and (b) for an illustration of the mean and main directors.  

\begin{proposition}{Two weighted directors with the same weight.}\label{prop:twodir}
  For two weighted directors with the same weight, denoted as $(\Vv_1, w)$ and $(\Vv_2, w)$, 
  the main director is $(\frac{\Vv_1+\Vv_2}{\|\Vv_1+\Vv_2\|}, w(1+ \Vv_1^T\Vv_2))$ if $w \Vv_1^T\Vv_2 \geq 0$, 
  and is $(\frac{\Vv_1-\Vv_2}{\|\Vv_1-\Vv_2\|}, w(1 - \Vv_1^T\Vv_2))$ if $w \Vv_1^T\Vv_2 \leq 0 $. 
  The mean director is $\frac{|w|}{2}(\Vv_1+\Vv_2)$, if $\Vv_1^T\Vv_2 \geq 0$, 
  and is $\frac{|w|}{2}(\Vv_1-\Vv_2)$, if $\Vv_1^T\Vv_2 \leq 0$. 
\end{proposition}

The mean director and main director describe different meaningful information about directors. 
Take a diffusion tensor $\BM{D}=\sum_{i=1}^3 \lambda_i \Vv_i \Vv_i^T$, $\lambda_1>\lambda_2>\lambda_3>0$, as an example. 
There are three weighted directors $\{(\Vv_i, \lambda_i)\}_{i=1}^3$, or represented as $\{\lambda_i\Vv_i\}$ because of non-negative weights. 
The mean director is $\frac{1}{3}\sum_{i=1}^3 s_i \lambda_i \Vv_i$ with any sign assignment of $\{s_i\}$, 
while the main director is $\lambda_1\Vv_1$. 
See Fig.~\ref{fig:directors} (b). 
Additionally, small changes in $\lambda_2$, $\lambda_3$, $\Vv_2$, and $\Vv_3$ may change the mean director, but not the main director, if $\lambda_1$ and $\Vv_1$ are still the largest eigenvalue and eigenvector. 
This example clearly shows that the mean director concept is a generalization of the mean vector concept in vector space, while the main director emphasizes the main axis in PCA. 
Please note that in a general case, the change of any director $(\Vv_i, w_i)$ may cause the change of the main director (i.e., the largest eigenvector and eigenvalue of the tensor $\sum_{i}w_i \Vv_i \Vv_i^T$) and also the mean director.

\subsubsection{Two Representations of the Difference between Two Directors}
\label{sec:diff_dir}

We aim to generalize the tensor field analysis in Section~\ref{sec:method_tensor} to director fields (i.e., ODF fields), 
and explore geometric structure information by using spatial derivatives which rely on the concept of difference between two directors.

We propose two ways, i.e., the director representation denoted as $\texttt{Diff}_\text{d}$ and the rotation matrix representation denoted as $\texttt{Diff}_\text{r}$, 
to represent the difference between two weighted directors with non-negative weights, $w_1\Vv_1$ and $w_2\Vv_2$. 
These two directors can be converted to the vectors $w_1s_1\Vv_1$ and $w_2s_2\Vv_2$ by assigning the sign $s_1=1$ (or $s_1=-1$), and $s_2$ such that $w_1w_2s_1s_2\Vv_1^T\Vv_2 \geq 0$. 
Thus, there are two different cases because of the sign ambiguity.
We can represent the difference as a director, i.e., $\texttt{Diff}_\text{d}(w_1\Vv_1, w_2\Vv_2) = w_1s_1\Vv_1 - w_2s_2\Vv_2$. 
We can also represent the difference as a scaled rotation matrix $\BM{R}$ which rotates $w_2s_2\Vv_2$ to $w_1s_1\Vv_1$, i.e., $w_1s_1\Vv_1 = \BM{R}w_2s_2\Vv_2 $, $\texttt{Diff}_\text{r}(w_1\Vv_1, w_2\Vv_2) = \BM{R}$. 
The rotation matrix can be calculated from the rotation axis $s_2\Vv_2\times s_1\Vv_1$ (i.e., the cross product of $s_2\Vv_2$ and $s_1\Vv_1$) and the rotation angle (i.e., the angle between $s_2\Vv_2$ and $s_1\Vv_1$)~\footnote{\url{https://en.wikipedia.org/wiki/Rotation_matrix\#Rotation_matrix_from_axis_and_angle}}, 
and the scale can be calculated from the weights $w_1$ and $w_2$. 
Note that this rotation matrix is the same for the above two cases, without sign ambiguity.  
See Fig.~\ref{fig:directors} (c) as an illustration. 
The director representation of the difference has the sign ambiguity, but it gives a vector without a sign which can be projected onto axes.
The rotation matrix representation is unique without sign ambiguity, but cannot be projected onto axes.

\subsubsection{Spatial Gradient and Directional Derivative of a Director Field} 
\label{sec:derivative_dir}

Considering a director field where there is only one director with non-negative weight $w\Vv(\Vx)$ (simplified notation for $w(\Vx)\Vv(\Vx)$) at each position $\Vx \in \mathbb{R}^3$, 
a directional derivative along $\uu$ at $\Vx$ is defined as
\begin{equation}\label{eq:derivative_dir}
  \frac{\partial w \Vv}{ \partial \uu} = \lim_{k\to 0} \frac{\texttt{Diff}( w\Vv(\Vx+ k\uu), w\Vv(\Vx- k\uu))}{ 2 k}. 
\end{equation}
Thus, there are also director and rotation matrix representations of the directional derivative because of the two representations of $\texttt{Diff}$, i.e., $\texttt{Diff}_\text{d}$ and $\texttt{Diff}_\text{r}$. 

For the director field where directors $w\Vv(\Vx)$ are only obtained in a integer lattice, 
the central difference can be used to approximate the spatial gradient $[\frac{\partial w\Vv }{ \partial x_i}]$, where
\begin{small}
\begin{equation}\label{eq:central_diff}
  \frac{\partial w\Vv }{ \partial x_i} \approx \texttt{Diff}(w\Vv(\Vx+\VV{o}_i) , \texttt{Mean}(\{w\Vv(\Vx-\VV{o}_i), w\Vv(\Vx+\VV{o}_i)\}) ), 
\end{equation}
\end{small}%
$\VV{o}_1=[1,0,0]^T$, $\VV{o}_2=[0,1,0]^T$, $\VV{o}_3=[0,0,1]^T$ are the unit norm vectors along spatial axes, 
and $\texttt{Mean}$ is the mean director operator in Definition~\ref{def:mean_dir}. 

We normally use the rotation matrix representation for the spatial gradient $\frac{\partial w\Vv}{ \partial x_i}$, considering this representation is unique. 
Let $\{\BM{R}_i(\Vx)\}$ be the rotation matrices for the spatial gradient at $\Vx$ along axes $\{\Vx_i\}$, i.e., $\texttt{Diff}_\text{r}$ is used in~\EEqref{eq:central_diff}. 
Then, analogously to the spatial gradient of a vector field, the director $w\Vv(\Vx+ k\uu)$ at position $\Vx+k\uu$ with small $k$ can be approximated as the sum of rotated directors, i.e., 
\begin{footnotesize}
\begin{equation}\label{eq:weighted_ratation_dir}
  \sum_{i=1}^3 k\VV{p}_i(\Vx), \quad  \text{where} \ \  \VV{p}_i(\Vx) =   \left\{ \begin{aligned}
    & u_i\BM{R}_i(\Vx) w(\Vx)\Vv(\Vx), \ \text{if}\ \ u_i \geq 0\\
    & -u_i\BM{R}_i^T(\Vx) w(\Vx)\Vv(\Vx), \ \text{if}\ \  u_i<0.
       \end{aligned}
 \right.
\end{equation}
\end{footnotesize}%
Note that every director in the above sum is a rotated $w(\Vx)\Vv(\Vx)$ in a small local rotation, 
thus we assume all directors are in a $90^\circ$ cone to obtain a simple sum of vector representation. 
See Proposition~\ref{prop:mean_cone_dir}. 
If~\EEqref{eq:central_diff} is used to approximate $\BM{R}_i$, then the angle between $w\Vv(\Vx)$ and $\BM{R}_i w\Vv(\Vx)$ is no more than $45^\circ$, 
thus all three rotated directors are indeed in a $90^\circ$ cone.

\subsection{Orientational Order and Dispersion}
\label{sec:order}

Before working on a field of ODFs, an ODF in a voxel can provide some geometric information at the voxel level, 
including GFA~\citep{Tuch2004}, and orientation dispersion~\citep{zhang_NODDI_NI2012}.

\subsubsection{Orientational Order Transform and Orientational Tensor}

The NODDI model is increasingly used to study neurite orientation dispersion~\citep{zhang_NODDI_NI2012}.  
NODDI uses the Watson distribution in~\EEqref{eq:watson} to model the ODF with a single orientation, 
where $M$ is the confluent hypergeometric function, 
$\Vn_0 \in \mathbb{S}^2$ is a given axis
\begin{footnotesize}
\begin{equation}\label{eq:watson}
  f(\uu \mid (\Vn_0, \kappa) ) = \frac{1}{4 \pi \ M(1/2, 3/2, \kappa) } \exp(\kappa (\uu^T \Vn_0 )^2), \quad \uu \in \mathbb{S}^2. 
\end{equation}%
\end{footnotesize}%
Note that the original formula of the Watson distribution in~\cite{zhang_NODDI_NI2012} has no unit integral in $\mathbb{S}^2$, because it missed $4\pi$. 
An orientation dispersion index (OD) was defined as~\EEqref{eq:OD_w}, where we denote it as $\text{OD}_{\text{w}}$ because it only applies to the Watson distribution. 
\begin{equation}\label{eq:OD_w}
  \text{OD}_{\text{w}}=\frac{2}{\pi}\arctan(\frac{1}{\kappa})
\end{equation}%
Note that in order to obtain good contrast in the dispersion index map, 
in the NODDI toolbox provided by the authors, a scaled $\kappa$ ($10\kappa$ in the codes) is used in~\EEqref{eq:OD_w} to calculate $\text{OD}_{\text{w}}$, instead of the estimated $\kappa$ from the NODDI model. 
$\text{OD}_{\text{w}}$ can not be used for ODFs that have general shapes, have more than one peak, or are not antipodally symmetric. 
Some other works also proposed dispersion indices based on different models of ODFs, e.g., Bingham distributions~\cite{tariq:NI2016} and mv-$\Gamma$ distributions in DIAMOND~\cite{scherrer:MRM2015}. 
These dispersion indices cannot work for general ODFs. 
Inspired by liquid crystals, we would like to define the degree of dispersion for general ODFs, independent of microscopic diffusion signal models.

For a general spherical function $f(\uu)$, $\uu\in\mathbb{S}^2$, 
we define the \emph{orientational tensor} as 
\begin{equation}\label{eq:OO_qtensor}
  \BM{Q}(f) = \int_{\mathbb{S}^2} \uu\uu^T f(\uu) \rmd \uu, 
\end{equation}
which is related to the Q-tensor in liquid crystal modeling~\citep{andrienko_2006}~\textsuperscript{\ref{nt:lq}}. 
$\BM{Q}(f)$ is a $3\times 3$ symmetric matrix dependent on $f(\uu)$. 
If $f(\uu)$ is non-negative, then $\BM{Q}(f)$ is positive semidefinite. 
We propose the \emph{orientational order index (OO)} from  the theory of liquid crystals~\citep{andrienko_2006} 
to describe the orientation or dispersion of a general spherical function along a given axis $\VV{n}$: 
\begin{footnotesize}
\begin{align}
  \text{OO}(\VV{n}) &= \int_{\uu\in\mathbb{S}^2} P_2(\uu^T\VV{n}) f(\uu) \rmd \uu \nonumber \\
       &=\int_{\uu\in\mathbb{S}^2} \frac{3(\uu^T \VV{n})^2 -1}{2} f(\uu) \rmd \uu \label{eq:OO}
\end{align}
\end{footnotesize}%
where $P_2$ is the 2nd-order Legendre polynomial. 
By definition, \EEqref{eq:OO} is an \emph{integral transform} in $\mathbb{S}^2$ which converts the spherical function $f(\uu)$ to another spherical function $\text{OO}(\VV{n})$, and the kernel is $P_2(\uu^T\VV{n})$, 
similar to the Funk-Radon transform used in Q-Ball imaging~\citep{Tuch2004}, where the kernel is $\delta(\uu^T\VV{n})$. 
We call~\EEqref{eq:OO} the \emph{Orientational Order Transform (OOT)}, i.e., $\text{OOT}(f)=\text{OO}(\Vn)$.  
Note that we have 
\begin{equation}
\text{OO}(\Vn)=  \frac{3}{2}\Vn^T \BM{Q}(f) \Vn - \frac{1}{2} \int_{\mathbb{S}^2} f(\uu) \rmd \uu. 
\end{equation}
By definition, $\text{OO}(\Vn)$ is antipodally symmetric and has a global maximum and a global minimum on the unit sphere, which correspond to the largest and smallest eigenvectors of $\BM{Q}(f)$, respectively. 
Based on Definition~\ref{def:main_dir}, the main director of infinite weighted directors $\{(\uu_i, f(\uu_i))\}$ is the maximum point of $\text{OO}(\Vn)$. 

\begin{figure}[t!]
  \centering
\includegraphics[scale=1.4,draft=false]{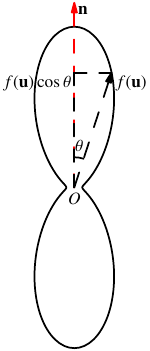}
  \caption{\label{fig:OO}A cross-section view of a spherical function $f(\uu)$ along axis $\Vn$. The projection of $f(\uu)\uu$ onto $\Vn$ is $(f(\uu)\cos\theta)\Vn$.}
\end{figure}

Although $\text{OO}(\VV{n})$ is a spherical function, it is a scalar index when $\VV{n}$ is chosen as a physically meaningful axis, e.g., $f(\uu)$ takes its maximal value at $\VV{n}$. 
Let $\theta$ be the angle between $\uu$ and axis $\VV{n}$, then $P_2(\uu^T\VV{n})=\frac{3\cos^2\theta-1}{2}$. 
Thus, if $f(\uu)$ is a Probability Density Function (PDF) on the unit sphere, 
then $\text{OO}(\VV{n})$ is $\langle \frac{3\cos^2\theta -1}{2} \rangle $, where $\langle \cdot \rangle$ signifies the expectation operation. 
As shown in Fig.~\ref{fig:OO}, $\langle \cos^2\theta \rangle $ is the expectation of the squared projected length of $\uu$ onto the axis $\VV{n}$. 
If $f(\uu)$ is more concentrated along $\Vn$, then $\langle \cos^2\theta \rangle $ is larger, so is $\text{OO}(\Vn)$. 
Based on the definition, when $f(\uu)$ is a PDF, then we have $\text{OO}(\VV{n})\in[-0.5, 1]$. 
If $f(\uu)=\delta(\uu^T\VV{n}_0-1)$, i.e., the delta function along a given $\VV{n}_0$ axis, then $\text{OO}(\VV{n}_0)=1$. 
If $f(\uu)=0$, $\forall \uu \in \mathbb{S}^2$ such that $\uu^T\VV{n}_0\neq 0 $, i.e., $f(\uu)$ is non-zero only in the plane orthogonal to $\VV{n}_0$, then $\text{OO}(\VV{n}_0)=-0.5$. 
If $f(\uu)$ is the isotropic PDF, i.e., $f(\uu)=\frac{1}{4\pi}$, then $\text{OO}(\VV{n})=0$. 
In practice, if we choose the axis $\VV{n}$ such that $f(\uu)$ takes its local or global maximal value, then $\text{OO}(\VV{n})$ is normally non-negative.

We define the \emph{orientational dispersion} along axis $\VV{n}$ as 
\begin{equation}
  \text{OD}(\VV{n})=1-\text{OO}(\VV{n}).
\end{equation}
Then $\text{OD}(\VV{n})\in [0,1.5]$.

Note that the proposed OO is different from the order parameter in~\cite{lasivc:FrontiersInPhysics2014,szczepankiewicz:NI2015} which was also inspired by liquid crystals~\citep{andrienko_2006}. 
In \cite{lasivc:FrontiersInPhysics2014,szczepankiewicz:NI2015}, the order parameter is calculated by estimating the variance of microscopic diffusion parameters from the contrast between signals measured by directional and isotropic diffusion encoding. 
However, it cannot be used for general DTI and HARDI data.  
The proposed OO in this paper is defined for general spherical functions (i.e., ODFs) along a given axis, independent of microscopic diffusion models and reconstruction of the ODFs.

\subsubsection{Axisymmetric Spherical Functions}

When $f(\uu)$ is axisymmetric, and its axis is given by $\VV{n}_0$, i.e., $f(\uu)=f'(\uu^T\VV{n}_0)$, where $f'(x)$ is the corresponding scalar function defined in $[-1,1]$, 
then OOT has a closed form: 
\begin{footnotesize}
\begin{align}
  \text{OO}(\VV{n})  &  = \int_0^\pi \left(  \int_0^{2\pi} P_2( \cos\theta \cos t + \sin\theta\cos t \sin \phi)\rmd t \right)f'(\cos\theta) \rmd \theta \nonumber \\
     &= \frac{(1+3\cos(2\phi))\pi}{2} a_2  = \frac{1+3\cos(2\phi)}{4} \text{OO}(\Vn_0) \label{eq:OOT_sym}
\end{align}
\end{footnotesize}
where $\phi=\arccos(|\Vn^T\Vn_0|)$ is the angle between $\Vn$ and the axis $\Vn_0$, 
and $a_2 = \int_{-1}^1 P_2(x) f'(x) \rmd x$ is the 2nd-order Legendre coefficient of $f'(x)$. 
Note that if $a_2>0$, when $\VV{n}=\VV{n}_0$, $\phi=0$, then $\text{OO}=2\pi a_2$ is the global maximum of $\text{OO}(\VV{n})$. 
When $\VV{n}^T\VV{n}_0=0$, $\phi=\pi/2$, then $\text{OO}=-\pi a_2$ is the global minimum of $\text{OO}(\VV{n})$. 
In the following development, without any ambiguity, we will use OO to denote $\text{OO}(\Vn_0)$, and OD to denote $\text{OD}(\Vn_0)$, for axisymmetric spherical functions. 

\subsubsection{Watson Distributions} 

The Watson distribution defined in~\EEqref{eq:watson} is axisymmetric with the axis $\VV{n}_0$.  
Thus, based on the above analysis of axisymmetric spherical functions, 
we have $\text{OO}(\Vn)= \frac{1+3\cos(2\phi)}{4} \text{OO}$, and
\begin{equation}\label{eq:OO_watson}
  \text{OO} = \frac{3 e^\kappa}{2\sqrt{\kappa\pi}\ \texttt{Erfi}(\sqrt{\kappa}) }  - \frac{3+2\kappa}{4\kappa}
\end{equation}
where $\texttt{Erfi}(x)=\frac{2}{\sqrt{\pi}}\int_0^x \exp(t^2)\rmd t$ is the imaginary error function. 
Then $\text{OD}=1-\text{OO}$.  
The left part of Fig.~\ref{fig:OO_tensor_watson} shows the above $\text{OD}$ and $\text{OD}_{\text{w}}$ as functions of $\kappa$, 
where the axis $\VV{n}$ is set as the Watson distribution's axis $\Vn_0$. 
Both dispersion indices decrease as $\kappa$ increases. 
Based on the derivation of $\kappa$, $\text{OD}_{\text{w}}$ is more sensitive to changes of $\kappa$ when $\kappa$ is small ($<2$), 
while it is less sensitive when $\kappa$ is large ($>2$).  
Compared with $\text{OD}_{\text{w}}$, the change of $\text{OD}$ is smoother for the change of $\kappa$ over the entire range of $\kappa$. 

\subsubsection{Tensors} 

\begin{figure*}[t!]
  \centering
\includegraphics[width=0.48\linewidth]{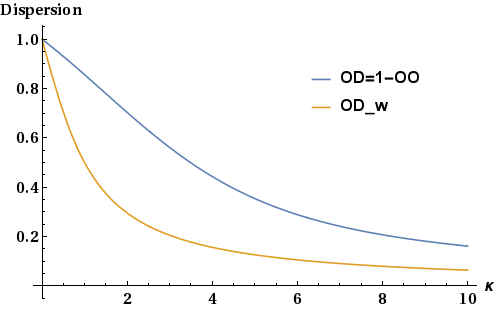}
\includegraphics[width=0.48\linewidth]{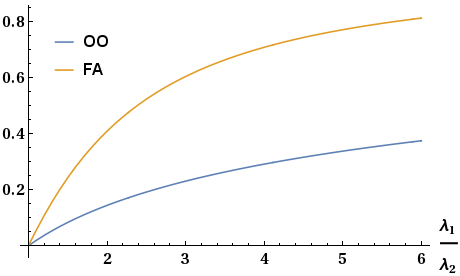}
\caption{\label{fig:OO_tensor_watson} 
  Left: dispersion indices ($\text{OD}$ and $\text{OD}_{\text{w}}$) of a Watson distribution as functions of $\kappa$. 
  Right: OO and FA of prolate tensors ($\lambda_2 = \lambda_3$) as functions of $\frac{\lambda_1}{\lambda_2}$.}
\end{figure*}

For the tensor model in DTI, denoted as $\BM{D}$, OOT is defined for its ODF, i.e.,   
\begin{equation}\label{eq:odf_tensor}
  \Phi (\uu \mid \BM{D}) = \frac{1}{4\pi |\BM{D}|^{\frac{1}{2}} } \frac{1}{(\uu^T \BM{D}^{-1}\uu)^{\frac{3}{2}}}, 
\end{equation}%
which is a PDF on the unit sphere. 
When the three eigenvalues of $\BM{D}$ satisfy $\lambda_1>\lambda_2=\lambda_3>0$, 
$\Phi (\uu \mid \BM{D})$ is an axisymmetric function with the axis $\Vv_1$ that is the principal eigenvector of $\BM{D}$. 
OOT has a closed-form expression in~\EEqref{eq:OOT_sym}, and
\begin{footnotesize}
\begin{equation}\label{eq:OO_tensor}
  \text{OO} = \frac{ \sqrt{\lambda_1-\lambda_2}(2\lambda_1+\lambda_2) - 3\lambda_1\sqrt{\lambda_2}\arctan\left(\sqrt{\frac{\lambda_1-\lambda_2}{\lambda_2}}\right)   }{2(\lambda_1 - \lambda_2)^{\frac{3}{2}}}.
\end{equation}%
\end{footnotesize}%
The right panel of Fig.~\ref{fig:OO_tensor_watson} shows $\text{OO}$ and FA as functions of $\lambda_1/\lambda_2$, where we set $\VV{n}=\Vv_1$. 
Both $\text{OO}$ and FA increases as $\lambda_1/\lambda_2$ increases. 
Thus, $\text{OO}$ can be seen as a type of anisotropy index for tensors. 
For general tensors with $\lambda_1>\lambda_2>\lambda_3>0$, no such closed form solution like~\EEqref{eq:OO_tensor} and~\EEqref{eq:OOT_sym} exists, 
but we can calculate OO using the spherical harmonic representation of the ODF. 

\subsubsection{Spherical Harmonic Representation} 

For a general spherical function $f(\uu)$, OO and OD can be analytically calculated from the spherical harmonic coefficients of the rotated function. 
Considering $f(\uu)$ is a real function on the unit sphere, it can always be linearly represented by the real Spherical Harmonic (SH) basis $\{Y_l^m(\uu)\}$, i.e., 
\begin{equation}\label{eq:SH_f}
  f(\uu) = \sum_{l,m} c_{l,m} Y_l^m(\uu)
\end{equation}
\begin{equation}\label{eq:SH}
  Y_l^m(\theta,\phi) =         
\left\{  \begin{array}{lcl}
  \sqrt{2}\mbox{Re}(y_l^{|m|}(\theta,\phi))  &  \mbox{if} &  -l\leq m<0  \\  
y_l^m(\theta,\phi)    &  \mbox{if}  &  m=0  \\
\sqrt{2}\mbox{Im}(y_l^m(\theta,\phi))  &  \mbox{if}  &  l\geq m>0
\end{array}\right.   
\end{equation}%
where $y_l^m(\theta,\phi) = \sqrt{\frac{2l+1}{4\pi}\frac{(l-m)!}{(l+m)!}} e^{im\phi} P_l^m(\cos\theta)$ is the complex SH basis, $P_l^m(\cdot)$ is the associated Legendre polynomial. 
For any rotation matrix, the SH coefficients of the rotated function can be calculated with very high accuracy 
based on the Wigner D-matrix~\footnote{\href{https://en.wikipedia.org/wiki/Spherical_harmonics}{https://en.wikipedia.org/wiki/Spherical\_harmonics}}, 
or based on fitting rotated function samples~\citep{lessig_2012}. 
Let $\BM{R}$ be the rotation matrix which rotates the axis $\VV{n}$ to $\VV{z}$-axis, 
and $\{a_{l,m}\}$ be the real SH coefficients of the rotated function $(Rf)(\uu)=f(\BM{R}^{-1}\uu)$, 
considering the orthogonality of the real SH basis and $Y_2^0(\theta,\phi)=\sqrt{\frac{5}{4\pi}}P_2(\cos\theta)$, we have 
\begin{footnotesize}
\begin{align}
  \text{OO}(\VV{n}) &= \int_{\uu\in\mathbb{S}^2} P_2(\uu^T\VV{n}) f(\uu) \rmd \uu \nonumber \\
   &=\int_{\uu\in\mathbb{S}^2} P_2(\cos \theta) \sum_{l,m} a_{l,m} Y_l^m(\theta, \phi) \rmd \uu = \sqrt{\frac{4\pi}{5}} a_{2,0}. \label{eq:OO_sh}
\end{align}
\end{footnotesize}%
Note that $\text{OO}(\VV{n})$ is only determined by the rotated SH coefficient $a_{2,0}$ that is only related to $\{c_{2,m}\}_{-2\leq m \leq 2}$ and the axis $\VV{n}$, based on the rotation property of the SH basis. 
Thus, $\text{OO}(\VV{n})$ is only related to the SH coefficients of $f(\uu)$ with $l=2$, and also the axis $\Vn$. 

\subsubsection{Relationship Between OO, OD, and GFA}
\label{sec:gfa_OO}

For an ODF in an SH representation in~\EEqref{eq:SH_f}, its GFA~\citep{Tuch2004} is
\begin{equation}\label{eq:GFA}
  \text{GFA}  = \sqrt{ 1- \frac{ c_{0,0}^2}{\sum_{lm} c_{l,m}^2}}. 
\end{equation}

Note that the rotation of a spherical function does not change the shape of the function and the norm of SH coefficients, 
thus we have $\sum_{m}a_{2,m}^2=\sum_{m}c_{2,m}^2$. 
Based on~\EEqref{eq:OO_sh}, we have 
\begin{align}
  \text{OO}(\Vn) &= \sqrt{\frac{4\pi}{5}} a_{2,0} \leq\sqrt{\frac{4\pi}{5}} \sqrt{\sum_{m=-2}^2 c_{2,m}^2} \nonumber \\ 
   & \leq \sqrt{\frac{4\pi}{5}} \sqrt{\sum_{l\geq 2}\sum_{m=-l}^l c_{l,m}^2}.\label{eq:OO_ineq_1}
\end{align}
Combining~\EEqref{eq:GFA} and~\EEqref{eq:OO_ineq_1}, we have  
\begin{equation}
  \text{OO}(\Vn)  \leq \sqrt{\frac{4\pi c_{0,0}^2}{5}} \sqrt{ \frac{1}{1-\text{GFA}^2} - 1}.
\end{equation}
The above inequality gives an upper bound of $\text{OO}(\Vn)$ as a function of $\text{GFA}$ which is independent of $\Vn$. 
Note that the above upper bound is tight, and the equality holds when $c_{l,m}=0$, for $l>2$, and $a_{2,m}=0$ for $m\neq 0$ after rotation.
If the ODF has unit integral, i.e., $\int_{\mathbb{S}^2} f(\uu) \rmd \uu =1$~\footnote{Note that ODFs estimated by some methods (e.g., constrained spherical deconvolution~\citep{tournier_NI2007}, Q-Ball Imaging~\citep{Tuch2004,Descoteaux2007}), 
do not have the unit integral, if there is no normalization after estimation.}, then $c_{0,0}= \frac{1}{\sqrt{4\pi}}$, and we have 
\begin{equation}\label{eq:OO_ineq}
  \text{OO}(\Vn)  \leq \sqrt{\frac{1}{5}} \sqrt{ \frac{1}{1-\text{GFA}^2} - 1 }
\end{equation}
Thus, for an ODF with low GFA, OO is also low, and OD is high, no matter how we choose the axis $\Vn$. 
Note that~\EEqref{eq:OO_ineq} does not imply that an ODF with high GFA tends to have high OO, because it is an upper bound of $\text{OO}(\Vn)$, not a lower bound. 

\subsubsection{Mixture Model}

\begin{figure*}[t!]
  \centering
\includegraphics[width=0.45\linewidth]{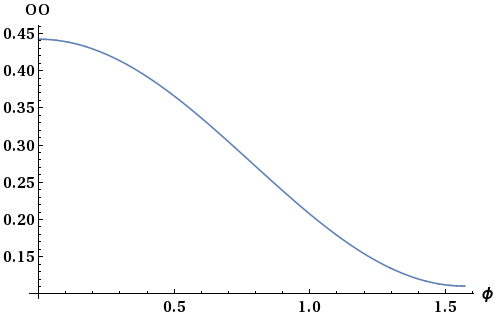}
\includegraphics[width=0.45\linewidth]{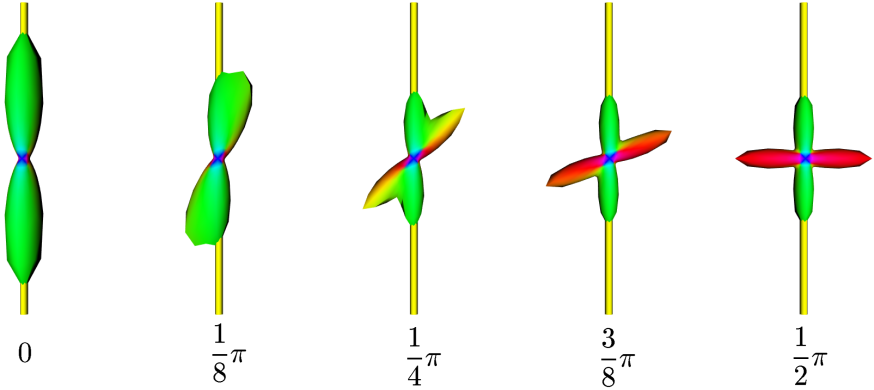}
\caption{\label{fig:OO_tensor2} 
  OO for the mixture tensor model.
  Left: OO as a function of the angle between two tensor components. 
  Right: ODF glyphs of the two-tensor model for different crossing angles, 
  where the yellow tube shows the $y$-axis which is used to calculate OO. }
\end{figure*}

OOT in~\EEqref{eq:OO} is a linear transform.  
Thus, if  $f(\uu)=\sum_i w_i f_i(\uu)$ is the PDF of a mixture of models, where $f_i(\uu)$ is the PDF for the $i$-th model, and $w_i$ is the weight,  
then $\text{OO}(\VV{n})= \sum_i w_i \text{OO}_i(\VV{n})$ is also a mixture of OO functions. 
Fig.~\ref{fig:OO_tensor2} illustrates OO for a two-tensor model with a crossing angle $\phi$, 
where two tensors share the same eigenvalues $[1.7,0.2,0.2]\times 10^{-3} mm^2/s$, the weights are $0.5$ and $0.5$, 
and one tensor component is along the $y$-axis and the other one rotates from the $y$-axis to the $x$-axis. 
Based on~\EEqref{eq:OOT_sym} and~\EEqref{eq:OO_tensor}, OO for the mixture model can be analytically calculated.  

\subsubsection{OO and OD for a General ODF Along the Principal Peak}
\label{sec:OO_OD_odf}

In the above context, we focus on $\text{OO}(\Vn)$ and $\text{OO}(\Vn)$ as spherical functions. 
A physically meaningful axis $\Vn_0$ is needed to obtain scalar indices of OO and OD from $\text{OO}(\Vn)$ and $\text{OO}(\Vn)$. 
For an axisymmetric function $f(\uu)$, its axis can be used as described above. 
For a general function (e.g., an ODF), we can set the axis as the local maxima of $f(\uu)$ (e.g., detected peaks of the ODF), 
because the peaks of ODFs are considered as local fiber directions in dMRI. 
A general ODF may have more than one peak. 
The \emph{principal peak} of the anisotropic ODF $f(\uu)$, where the ODF takes its global maximum $\uu_1$, i.e., $f(\uu_1,\Vx)>f(\uu,\Vx)$, $\forall \uu\in \mathbb{S}^2$, 
is used to calculate OO and OD for the ODF. 
Note that peaks are detected from ODFs with all orders of SH coefficients, not only SH coefficients with $l=2$. 
Thus, the scalar indices of OO and OD are actually dependent on SH coefficients of ODFs with all orders. 
See Algorithm~\ref{alg:OO_OD} for the pipeline to calculate OO and OD maps from a given ODF field with SH representation, 
where peaks are detected for voxels whose GFA values are larger than a given threshold (e.g., $0.3$). 
It is also possible to calculate OO and OD for all voxels by setting the GFA threshold as $0$.
As shown in Section~\ref{sec:gfa_OO}, for the voxels with $\text{GFA}<0.3$, we have $\text{OO}<0.14$ and $\text{OD}>0.86$. 

\begin{algorithm}[!h]
\caption{\label{alg:OO_OD}\textbf{Calculation of OO and OD for ODFs with SH representation along principal peaks:}}
\SetAlgoLined
  \KwIn{ODF field $f(\uu,\Vx)=\sum_{lm}c_{l,m}(\Vx)Y_l^m(\uu)$ in SH representation.}
\KwOut{OO map, OD map.}
  Peak detection using gradient ascent for ODFs in voxels with the anisotropy higher than a given threshold (e.g., $\text{GFA}>0.3$). See Section~\ref{sec:dir}\;
  \For{each voxel $\Vx$ with detected peaks $\{\uu_i(\Vx)\}$}{
  1) Find the principal peak $\uu_1$ with the largest ODF value, i.e., $f(\uu_1,\Vx)>f(\uu_i,\Vx)$, $\forall i$ \;
  2) Calculate rotation matrix $\BM{R}$, which rotates $\uu_1$ to the $\VV{z}$-axis \;
  3) Calculate the rotated SH coefficient $a_{2,0}$ from $\{ c_{2,m}\}_{-2\leq m \leq 2}$ under the rotation $\BM{R}$ \;
  4) $\text{OO}=\sqrt{\frac{4\pi}{5}} a_{2,0}$ as shown in~\EEqref{eq:OO_sh}, and $\text{OD}=1-\text{OO}$ \;
  }
\end{algorithm}

\subsubsection{OO, OD and the Orientational Tensor in a Spatial Region}

The above OO, OD, and the orientational tensor are defined for a single voxel. 
They can also be defined for voxels in a spatial region of voxels. 
A linear weighting generalization of OO can be defined as  
\begin{equation}\label{eq:OOT_region}
  \text{OO}(\VV{n})  = \int_{\Vx\in \Omega} \int_{\uu\in\mathbb{S}^2} P_2(\uu^T\VV{n}) w(\Vx) f(\uu, \Vx) \rmd \uu \rmd \Vx.
\end{equation}
The orientational tensor in a spatial region is
\begin{equation}\label{eq:OO_qtensor_region}
  \BM{Q}(f) = \int_{\Vx\in \Omega} \int_{\uu\in\mathbb{S}^2} w(\Vx) f(\uu, \Vx) \uu\uu^T \rmd \uu \rmd \Vx.
\end{equation}
Because of the linearity of the integration, 
\EEqref{eq:OOT_region} is actually OOT in~\EEqref{eq:OO} performed on the region smoothed spherical function $\int_{\Vx\in \Omega} w(\Vx) f(\uu, \Vx) \rmd \Vx$, 
and \EEqref{eq:OO_qtensor_region} is the orientational tensor for the region smoothed function.
The largest eigenvector of $\BM{Q}(f)$ in~\EEqref{eq:OO_qtensor_region} indicates the main orientation of all ODFs $f(\uu,\Vx)$ in the region $\Omega$.

\subsection{Local Orthogonal Frame}
\label{sec:frame}

\begin{figure*}[t!]
  \centering
    \begin{tabular}{c@{\hskip 0.1in} c@{\hskip 0.1in} c }
  \includegraphics[width=0.31\textwidth]{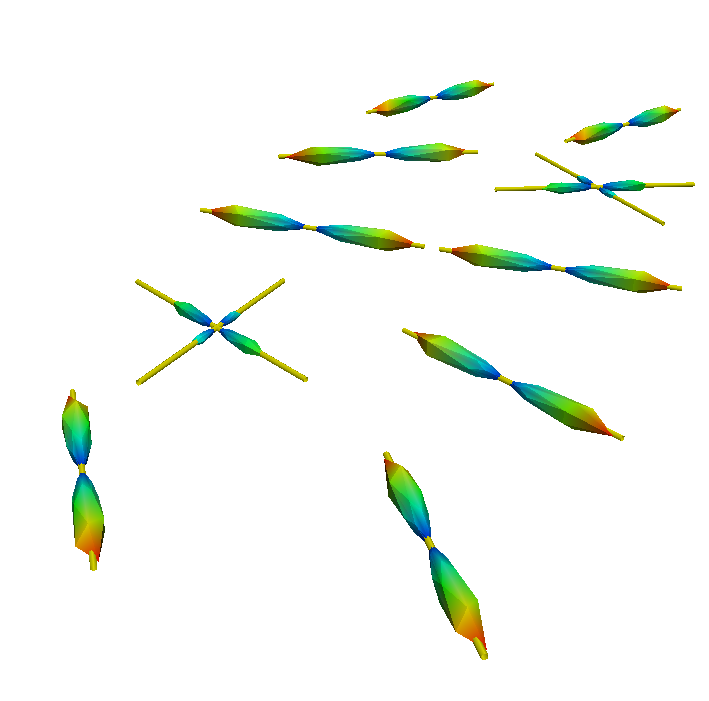} &
      \includegraphics[width=0.31\textwidth]{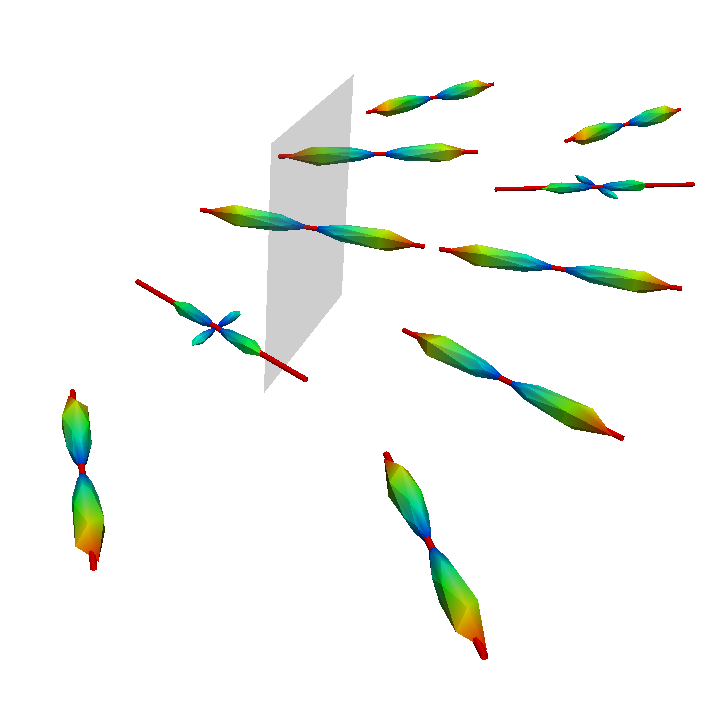} &
\includegraphics[width=0.31\textwidth]{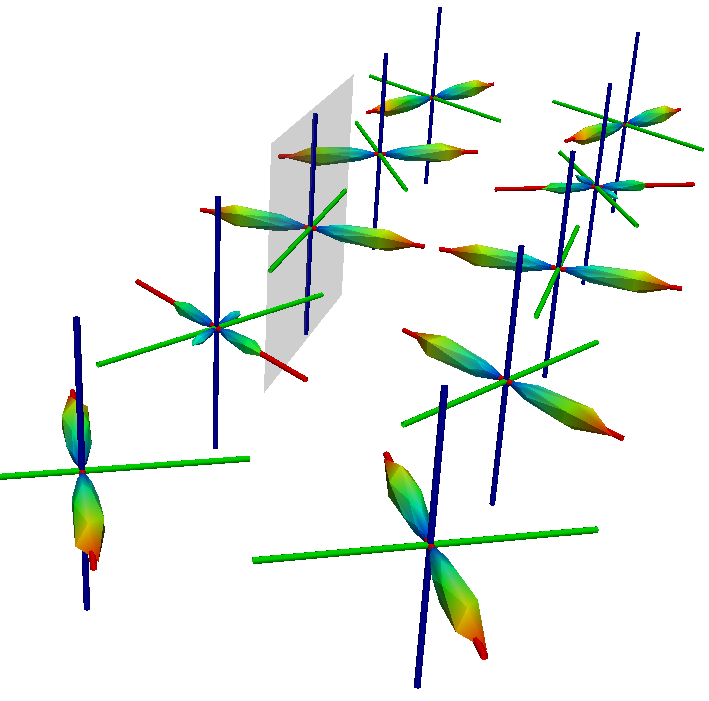} \\
      (a) & (b)  & (c)
\end{tabular}
  \caption{\label{fig:demo_frame}Sketch to determine local orthogonal frames from an ODF field, where an ODF may have 0, 1, or more than 1 peaks. 
  (a) an ODF field with peaks, where yellow tubes denote peaks. 
  (b) the orthogonal plane for the principal peak, where red tubes denote principal peaks. 
  (c) local orthogonal frames, where three tubes in red, green, and blue colors denote three directors in local orthogonal frames.}
\end{figure*}

As described in Section~\ref{sec:dir}, after peak detection on a spherical function field or a tensor field, the obtained peak field is also a director field. 
We propose extracting a local orthogonal frame in each voxel exhibiting anisotropic diffusion from the detected peak field. 
The orthogonal frame has three orthogonal orientations. 
Denote the peaks at voxel $\Vx$ as $\{\uu_i(\Vx)\}$. 
The first orientation is the \emph{principal peak} where the ODF takes its global maximum $\uu_1(\Vx)$, i.e., $f(\uu_1,\Vx)>f(\uu,\Vx)$, $\forall \uu \in \mathbb{S}^2$. 
We call it the \emph{principal director} of the voxel $\Vx$.
The other two orientations are in the orthogonal plane of the principal direction. 
Considering $f(\uu)$ is normally antipodally symmetric in diffusion MRI, all these orientations are equivalent to their antipodal ones. 
Thus, we project all peaks in a spatial local neighborhood onto the orthogonal plane, 
and define a weighted sum of dyadic tensors in voxel $\Vx$: 
\begin{equation}\label{eq:dyadic_tensor}
  \BM{Q}_\Vx = \sum_{\Vy\in\Omega_\Vx}\sum_{i} w(\Vy,\Vx) f(\uu_i(\Vy),\Vy)\uu_{i,\perp}(\Vy)\uu_{i,\perp}^T(\Vy) 
\end{equation}
\begin{equation}\label{eq:dyadic_tensor_2}
  \uu_{i,\perp}(\Vy)=\uu_i(\Vy)-(\uu_i^T(\Vy)\uu_1(\Vx)) \uu_1(\Vx)
\end{equation}
where $\Omega_\Vx$ is a local neighborhood of voxel $\Vx$, $w(\Vy,\Vx)$ is the spatial weight which is normally set to be proportional to $\exp(-\frac{\|\Vy-\Vx\|^2}{2\sigma^2})$, 
$\delta$ which is normally set as 1 voxel controls spatial weight concentration,  
$\uu_{i,\perp}(\Vy)$ is the projected orientation $\uu_i(\Vy)$ onto the orthogonal plane of $\uu_1(\Vx)$. 
The above $3\times 3$ matrix $\BM{Q}_\Vx$ is actually the orientational tensor of all projected peaks in region $\Omega_\Vx$ based on~\EEqref{eq:OO_qtensor_region}, 
where the continuous integral is replaced by a discrete summation over all projected peaks in region $\Omega_\Vx$. 
Note that although we can define $\BM{Q}_\Vx$ using continuous ODF $f(\uu,\Vx)$ with projected directors in a continuous integration like~\EEqref{eq:OO_qtensor_region}, 
we choose a discrete summation over peaks, which actually focuses only on peaks and sets zero weights for orientations that are not peaks in the continuous integration. 
$\BM{Q}_\Vx$ in~\EEqref{eq:dyadic_tensor} has at most two non-zero eigenvalues, because it is defined by using $\{\uu_{i,\perp}(\Vy)\}$ in the orthogonal plane. 
The eigenvector for the largest absolute eigenvalue of $\BM{Q}_\Vx$ is set as the second orientation of the orthogonal frame, 
which is the main director of directors $\{(\uu_{i,\perp},\ w(\Vy,\Vx)f(\uu_i,\Vy) )\}$, and indicates the main orientation of the local spatial change of $\uu_1(\Vx)$ in the orientational plane. 
Note that we define $\BM{Q}_\Vx$ using the isotropic spatial weight $w(\Vy,\Vx)$ to capture the general spatial change of the principal director $\uu_1(\Vx)$ in the orthogonal plane. 
If one has a good motivation and specific spatial prior knowledge (e.g., to capture local change only in a specific region like hippocampus), 
an anisotropic spatial weight $w(\Vy,\Vx)$ with consideration of spatial prior knowledge may be useful. 
The third orientation in the orthogonal frame is set as the cross product of the first two orientations. 
These three orientations are three orthogonal directors due to sign ambiguity. 
Please see the sketch map in Fig.~\ref{fig:demo_frame} to determine local orthogonal frames from a given ODF field. 
If these two eigenvalues of $\BM{Q}_\Vx$ are equal or their difference is very small, then we set the second and third orientations in the orthogonal frame to zero, 
which means any two orthogonal vectors in the orthogonal plane can be the second and third axes in the orthogonal frame.

\subsection{Local Distortion Indices: Splay, Bend, and Twist}
\label{sec:distortion}

\begin{figure*}[t!]
  \centering
    \begin{tabular}{c@{\hskip 0.5in} c@{\hskip 0.5in} c }
  \includegraphics[scale=.15]{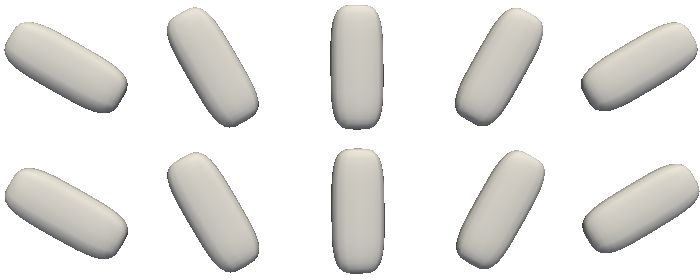} &
  \includegraphics[scale=.15]{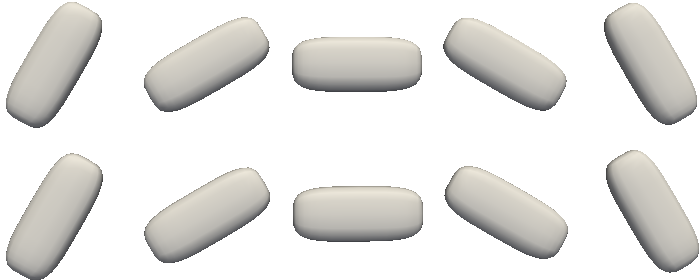} &
\includegraphics[scale=.15]{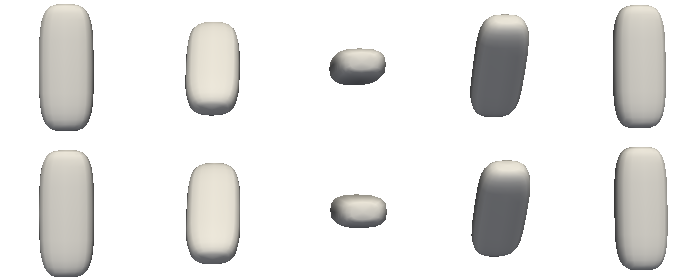} \\
  splay & bend & twist 
\end{tabular}
\caption{\label{fig:distortion}Demonstration of three types of distortions, i.e., splay, bend, and twist.}
\end{figure*}

\textbf{Three types of orientational distributions in liquid crystals}. 
Based on the liquid crystal analogy, there are three fundamental types of distortions~\textsuperscript{\ref{nt:lq}} for the director field as demonstrated in Fig.~\ref{fig:distortion}. 
1) \emph{splay}: bending occurs perpendicular to the director;
2) \emph{bend}:  bending is parallel to the director and molecular axis;
3) \emph{twist}: neighboring directors are rotated with respect to one another, rather than aligned. 
These three fundamental distortions can be used to describe a myriad of complex geometric patterns that liquid crystals can assume.
We would like to quantify these fundamental distortion patterns in dMRI by exploring the local spatial changes of principal directors. 

\textbf{Spatial derivatives of the local orthogonal frame}. 
With the local orthogonal frame $\{\uu_1(\Vx),\uu_2(\Vx),\uu_3(\Vx)\}$ at each voxel $\Vx$ obtained above, 
we can define the spatial directional derivatives of $\uu_i(\Vx)$ along a direction $\Vv$ as 
\begin{equation}\label{eq:derivative_dir_v}
  \frac{\partial \uu_i}{\partial \Vv}= \lim_{k\to 0} \frac{ \texttt{Diff}_\text{d}(\uu_i(\Vx+ k\Vv), \uu_i(\Vx- k\Vv))}{ 2 k}. 
\end{equation}%
$\texttt{Diff}_\text{d}$ is the director representation of the difference of two directors as described in Section~\ref{sec:diff_dir}.
Note that we use the director representation for the spatial derivative, instead of a rotation matrix representation, because we would like to project the director onto different axes. 
See Section~\ref{sec:diff_dir} and~\ref{sec:derivative_dir}.

\textbf{Spatial derivatives of vectors, and Maurer-Cartan connection forms in the moving frame method}. 
If we assume $\{\uu_i\}$ are all well-aligned unit vectors (i.e., no sign ambiguity), 
then we have 
\begin{equation}\label{eq:frame_derivative}
  \frac{\partial \uu_i}{\partial \Vv} = \nabla_\Vx \BM{\uu}_i :  \Vv 
\end{equation}
with elements $\left[ \sum_k \frac{\partial u_{il}}{\partial x_k}  v_k \right]$, 
where $u_{il}$ is the $l$-th element of $\uu_i$, and $\nabla_\Vx \BM{\uu}_i=[ \frac{\partial u_{il}}{\partial x_k}]$ is the spatial gradient matrix of $\uu_i(\Vx)$. 
Similarly with~\EEqref{eq:tensor2scalar}, we can extract some features by devising $\Vv$ and a weighting vector $\Vw$:
\begin{equation}\label{eq:frame_weight}
  \Vw^T\frac{\partial \uu_i}{\partial \Vv}= \Vw : \nabla_\Vx \BM{\uu}_i :  \Vv = \sum_{lm} w_l v_m \frac{\partial u_{il}}{\partial x_m}.
\end{equation}
\EEqref{eq:frame_weight} can be seen as a generalization of~\EEqref{eq:tensor2scalar} in tensor field analysis.
When we set $\Vw=\uu_j$ and $\Vv=\uu_k$, \EEqref{eq:frame_weight} is the projection of the directional derivatives onto $\uu_j(\Vx)$, denoted as $c_{ijk}$:
\begin{equation}
  c_{ijk} = \uu_j^T\frac{\partial \uu_i}{\partial \uu_k}= \uu_j : \nabla_\Vx \BM{\uu}_i :  \uu_k = \sum_{lm}u_{jl} u_{km} \frac{\partial u_{il}}{\partial x_m}.
\end{equation}%
$\{c_{ijk}(\Vx)\}$ is the Maurer-Cartan connection form in the moving frame method~\footnote{\url{https://en.wikipedia.org/wiki/Maurer-Cartan_form}}. 
$c_{ijk}(\Vx)$ denotes the spatial change rate of frame vector $\uu_i$ towards $\uu_j$ when moving the frame along $\uu_k$ at voxel $\Vx$~\citep{piuze_PAMI2015}.

\textbf{Orientational distortion indices}. 
In this paper, instead of directly using the connections $\{c_{ijk}(\Vx)\}$, 
we propose three scalar indices to describe the relative prevalence of each of the three types of local distortions of white matter, inspired by liquid crystals~\citep{andrienko_2006}. 
We define three indices and a total distortion index as
\begin{footnotesize}
\begin{align}
  \text{Splay index:} \ \ & s = \sqrt{c_{122}^2 + c_{133}^2} = \sqrt{ (\uu_2^T\frac{\partial \uu_1}{\partial \uu_2} )^2 +  (\uu_3^T\frac{\partial \uu_1}{\partial \uu_3} )^2} \label{eq:splay}\\
  \text{Bend index:} \ \ & b = \sqrt{c_{121}^2 + c_{131}^2} = \sqrt{ (\uu_2^T\frac{\partial \uu_1}{\partial \uu_1} )^2 +  (\uu_3^T\frac{\partial \uu_1}{\partial \uu_1} )^2}  \label{eq:bend}\\
  \text{Twist index:} \ \ & t = \sqrt{c_{123}^2 + c_{132}^2} = \sqrt{ (\uu_2^T\frac{\partial \uu_1}{\partial \uu_3} )^2 +  (\uu_3^T\frac{\partial \uu_1}{\partial \uu_2} )^2} \label{eq:twist}
\end{align}
\begin{equation}\label{eq:distortion}
  \text{Total distortion index:} \ \ d = \sqrt{s^2+b^2+t^2}. 
\end{equation}
\end{footnotesize}%

\begin{algorithm}[t!]
\caption{\label{alg:derivative}\textbf{Calculation of spatial directional derivatives of the principal director:}}
\SetAlgoLined
  \KwIn{A local orthogonal frame field $\{\uu_1(\Vx), \uu_2(\Vx), \uu_3(\Vx)\}$.}
\KwOut{Three spatial directional derivatives $\frac{\partial \uu_1}{\partial \uu_i}$, $i=1,2,3$.}
  // Calculate three rotational matrices $\{\BM{R}_i\}$, $i=1,2,3$ \;
  $\VV{o}_1=[1,0,0]^T$, $\VV{o}_2=[0,1,0]^T$, $\VV{o}_3=[0,0,1]^T$ \;
  \For{$i=1,2,3$}{
    $\Vv_1 = \uu_1(\Vx+\VV{o}_i)$, $\Vv_0 = \uu_1(\Vx-\VV{o}_i)$ \;
    \eIf{$\Vv_1^T\Vv_0\geq 0$}
    {$\Vv_2 = (\Vv_1+\Vv_0)/2$}
    {$\Vv_2 = (\Vv_1-\Vv_0)/2$}
  $\Vv_2= \frac{\Vv_2}{\|\Vv_2\|}$ \ \ \ \ // $\Vv_2$ is the normalized mean director of $\Vv_1$ and $\Vv_0$ \;
  Calculate rotation matrix $\BM{R}_i$ which rotates $\Vv_2$ to $\Vv_1$ \;
}
  // Calculate spatial directional derivatives from rotation matrices \;
  \For{$i=1,2,3$}{
    \For{$j=1,2,3$}{
      $u_{i,j}=\uu_i^T\VV{o}_j$ \;
    \eIf{$u_{i,j}\geq 0$}
      { $\VV{p}_j = u_{i,j} \BM{R}_j \uu_1 $, $\VV{n}_j = u_{i,j} \BM{R}_j^T \uu_1 $ \;}
      { $\VV{p}_j = -u_{i,j} \BM{R}_j^T \uu_1 $, $\VV{n}_j = -u_{i,j} \BM{R}_j \uu_1 $ \; }
    }
  $\VV{p}_0 = \VV{p}_1+\VV{p}_2+\VV{p}_3$, $\VV{p}_0=\frac{\VV{p}_0}{\|\VV{p}_0\|}$ \;
  $\VV{n}_0 = \VV{n}_1+\VV{n}_2+\VV{n}_3$, $\VV{n}_0=\frac{\VV{n}_0}{\|\VV{n}_0\|}$ \;
  \eIf{$\|\VV{p}_0-\VV{n}_0\| \leq \|\VV{p}_0+\VV{n}_0\|$ }
  { $\frac{\partial \uu_1}{\partial \uu_i} = \VV{p}_0-\VV{n}_0$}
  { $\frac{\partial \uu_1}{\partial \uu_i} = \VV{p}_0+\VV{n}_0$}
}
\end{algorithm}

\textbf{Numerical calculation of spatial derivatives of directors and orientational distortion indices}. 
Note that the above definitions of four indices and the formulae from~\EEqref{eq:frame_derivative} to~\EEqref{eq:distortion} are for a general vector frame in a vector field without sign ambiguity. 
We would like to calculate the above four indices for the local orthogonal frames in Section~\ref{sec:frame} with sign ambiguity. 
Squared values of $\uu_j^T\frac{\partial \uu_1}{\partial \uu_i}$ in definitions are used to avoid the sign ambiguity of $\{\uu_i\}$ and $\{\frac{\partial \uu_1}{\partial \uu_i}\}$. 
The difficulty of numerically calculating the above four indices is that it is challenging to calculate the three spatial directional derivatives $\{\frac{\partial \uu_1}{\partial \uu_i}\}$, $i=1,2,3$, 
because the local orthogonal frame $\{\uu_i(\Vx)\}$ with three directors is ambiguous with respect to its sign. 
In other words, $\uu_i$ is equivalent to $-\uu_i$, considering the ODF and its peaks are antipodally symmetric. 
We propose calculating the above spatial directional derivatives using a rotation matrix representation and a central difference approximation as described in Section~\ref{sec:derivative_dir}. 
See Algorithm~\ref{alg:derivative} for a detailed implementation. 
The algorithm first calculates three rotation matrices respectively along the $x$, $y$, $z$ axes, which is analogous to the spatial gradient of a vector field. 
Then $\frac{\partial \uu_1}{\partial \uu_i}$ is numerically approximated by the director representation of the difference, i.e., $\texttt{Diff}_\text{d}(\uu_1(\Vx+ \uu_i), \uu_1(\Vx- \uu_i))$, 
where $\uu_1(\Vx+ \uu_i)$ and $\uu_1(\Vx- \uu_i)$ are approximated by the weighted mean of three rotated vectors along three axes, as shown in~\EEqref{eq:weighted_ratation_dir}.  
After $\{\frac{\partial \uu_1}{\partial \uu_i}\}$ are obtained, we can calculate the above four indices in~\EEqref{eq:splay},~\EEqref{eq:bend},~\EEqref{eq:twist}, and~\EEqref{eq:distortion}, from the directional derivatives. 
Note that Algorithm~\ref{alg:derivative} avoids alignment of local frames in~\cite{piuze_PAMI2015} that does not work for general dMRI data.

\section{Experiments}
\label{sec:experiments}

\subsection{Synthetic Data Experiments}

\begin{figure*}[t!]
\resizebox{\textwidth}{!}{%
\begin{tabular}{c | c | c | c | c}
  &  splay & bend &   twist  & total distortion \\
  \hline
  \includegraphics[width=0.19\textwidth]{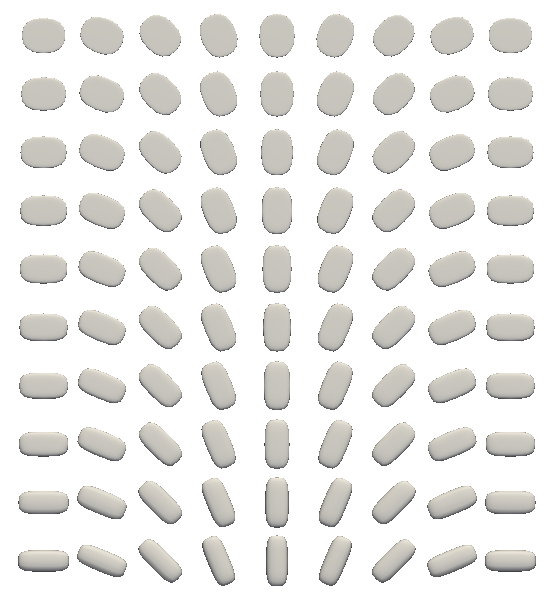}  & \includegraphics[width=0.19\textwidth]{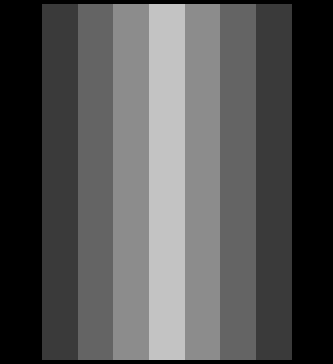}   & \includegraphics[width=0.19\textwidth]{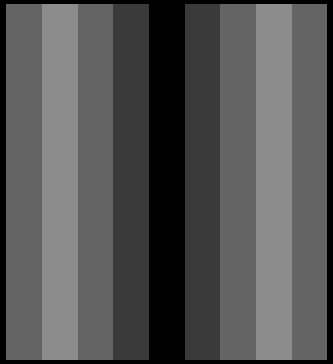} & \includegraphics[width=0.19\textwidth]{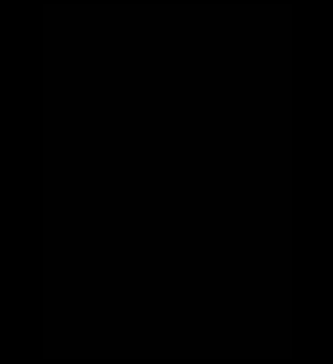} & \includegraphics[width=0.19\textwidth]{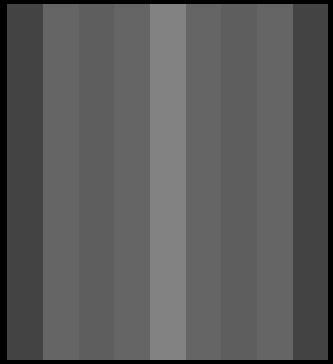}\\
  \hline
 \includegraphics[width=0.19\textwidth]{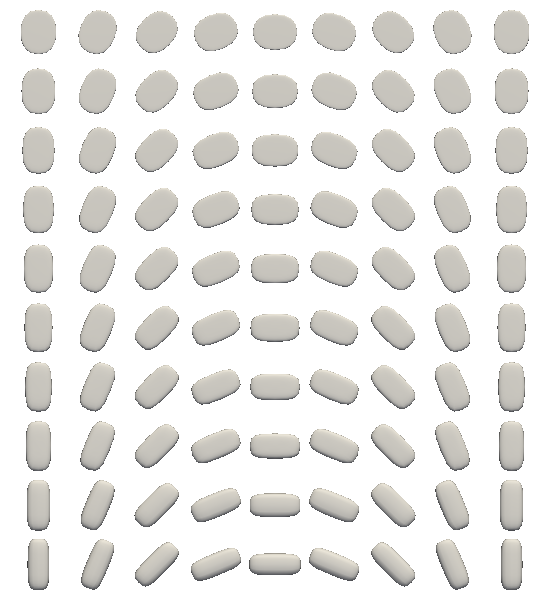}   & \includegraphics[width=0.19\textwidth]{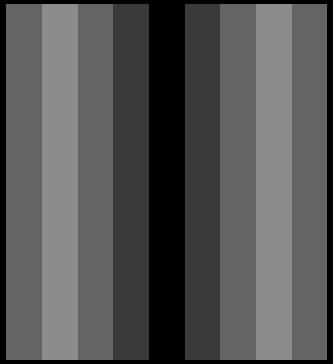}    & \includegraphics[width=0.19\textwidth]{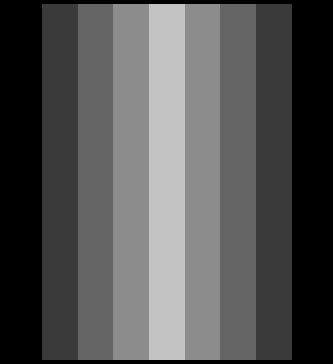}  & \includegraphics[width=0.19\textwidth]{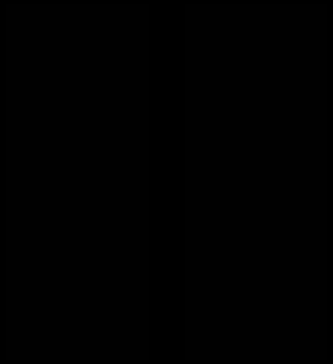} & \includegraphics[width=0.19\textwidth]{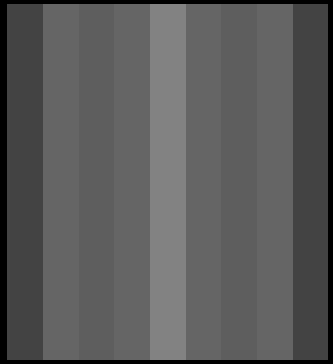}\\
  \hline
 \includegraphics[width=0.18\textwidth]{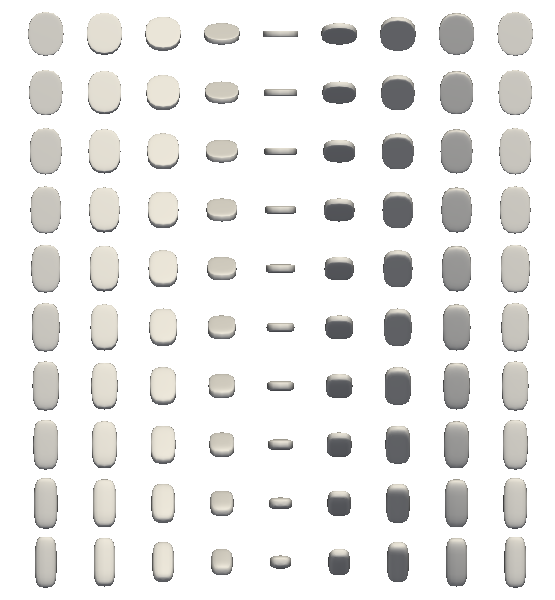}  & \includegraphics[width=0.18\textwidth]{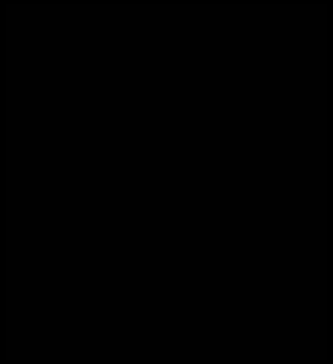}   & \includegraphics[width=0.18\textwidth]{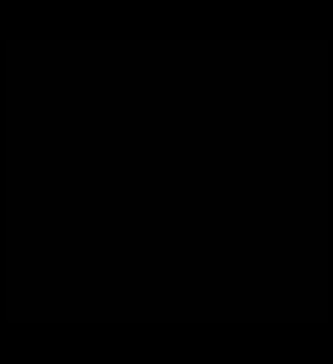} & \includegraphics[width=0.18\textwidth]{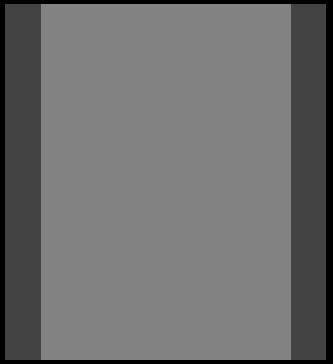} & \includegraphics[width=0.18\textwidth]{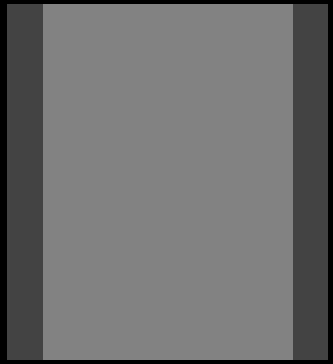}\\
\end{tabular}
  }
  \caption{\label{fig:distortion_exp}Distortion indices calculated from different tensor fields. 
  In each row, the four distortion indices are calculated from the ODF field obtained from the tensor field. 
  The tensors were visualized by using superquadric tensor glyphs~\citep{kindlmann_2004}.}
\end{figure*}

Fig.~\ref{fig:distortion_exp} demonstrates these four orientational distortion indices (i.e, splay, bend, twist, and total distortion) calculated from idealized tensor fields. 
The tensors are visualized by using superquadric tensor glyphs~\citep{kindlmann_2004}. 
The first column of Fig.~\ref{fig:distortion_exp} shows different tensor fields. 
The middle area of the first tensor field is the splaying area, while the middle area of the second tensor field is the bending region. 
These two tensor fields are generated by rotating a tensor from left to right around the $z$-axis perpendicular to the page and decreasing the tensor mode~\citep{Kindlmann_TMI2007}  from the bottom row to the top row. 
The third tensor field shows the twist of tensor orientations, 
which is generated by rotating a tensor around the $x$-axis (i.e., the left-to-right axis), and decreasing the tensor mode from bottom to top.
Fig.~\ref{fig:distortion_exp} shows that 
1) the four indices only depend on the orientations (i.e., local orthogonal frame), not on the tensor or ODF shape; 
2) splay, bend, twist indices provide complementary information about the orientational change, and demonstrate different types of orientational distortions.
Note that the twist index for the third tensor field is actually a constant, and the index value around the boundary is different due to the Neumann boundary condition used in the calculation. 
Although the results in Fig.~\ref{fig:distortion_exp} are for tensor fields,  
the distortion indices are actually determined by the local orthogonal frame field that can be calculated from a general spherical function field as described in Section~\ref{sec:frame}. 

\begin{figure*}[t!]
  \centering
\includegraphics[scale=.3]{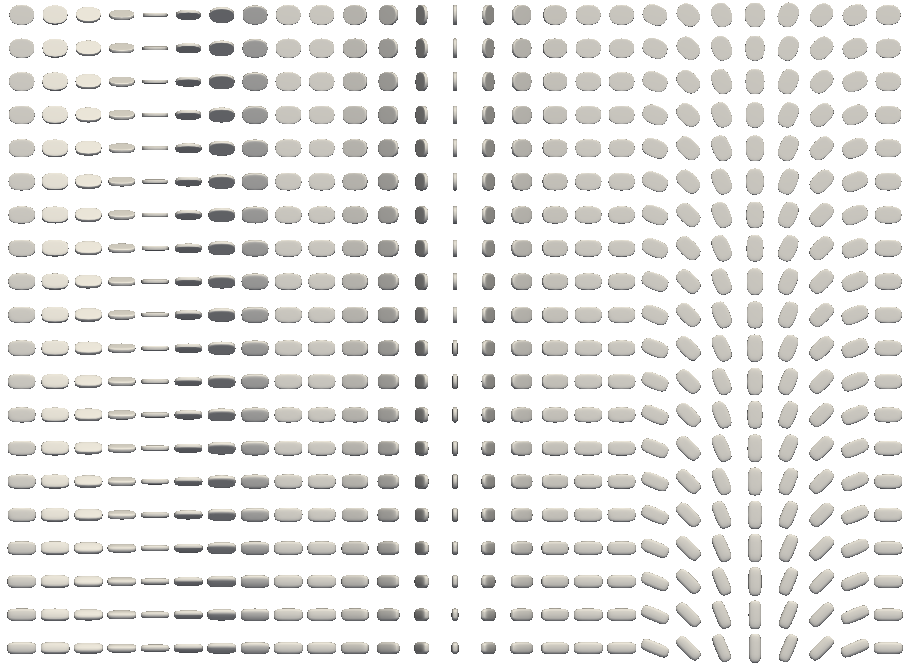}
\vspace{2mm} \\
\begin{tabular}{c  c  c}
  \includegraphics[scale=.4]{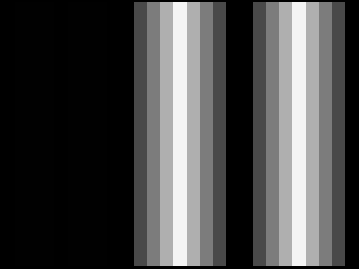}   & \includegraphics[scale=.4]{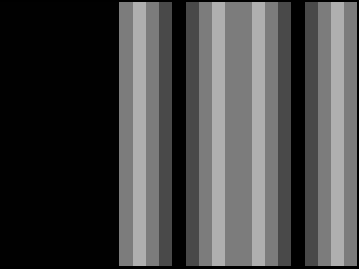} & \includegraphics[scale=.4]{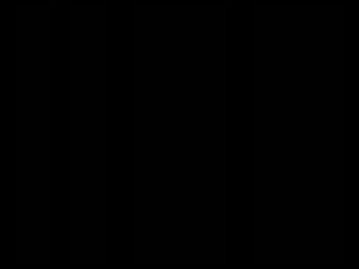}\\
  splay & bend & twist 
\vspace{2mm}
  \\
  \includegraphics[scale=.4]{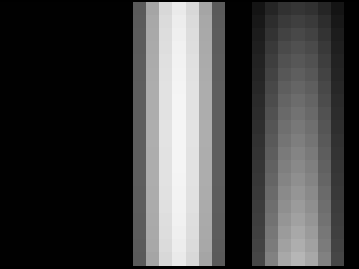}   & \includegraphics[scale=.4]{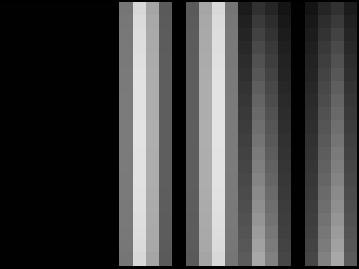} & \includegraphics[scale=.4]{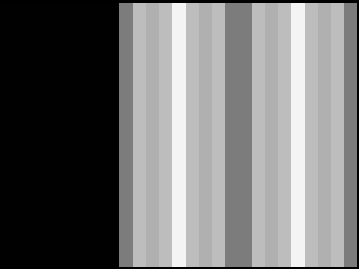}\\
  dispersion & curving &  total distortion
\end{tabular}
\caption{\label{fig:distortion_exp_compare}Dispersion, curving~\citep{Savadjiev_NI2010}, and the proposed four orientational distortion indices calculated from a tensor field.}
\end{figure*}

\begin{figure*}[t!]
  \centering

\begin{tabular}{c  c  c  c}
  \includegraphics[scale=.18]{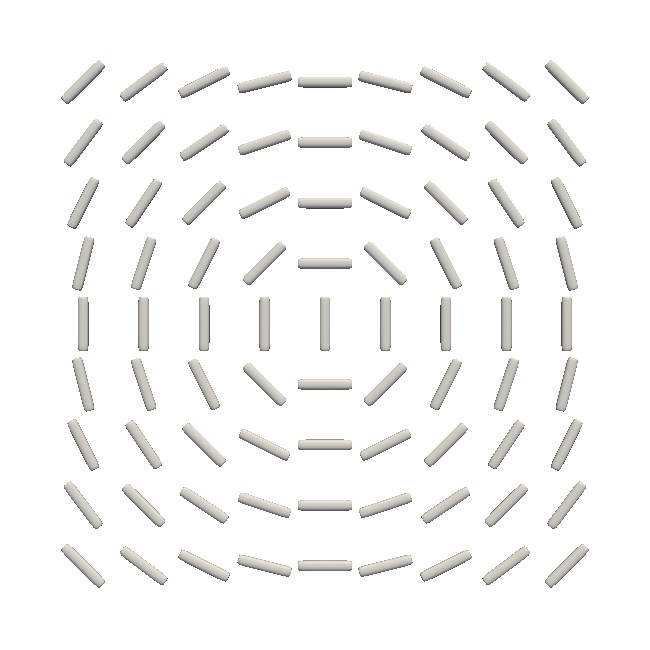} & \includegraphics[scale=.3]{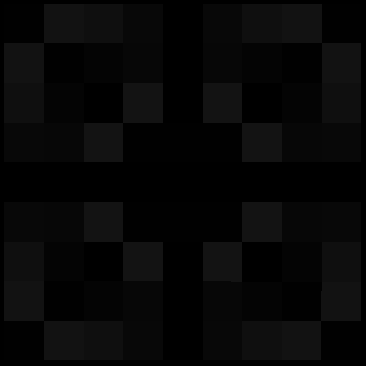} & \includegraphics[scale=.3]{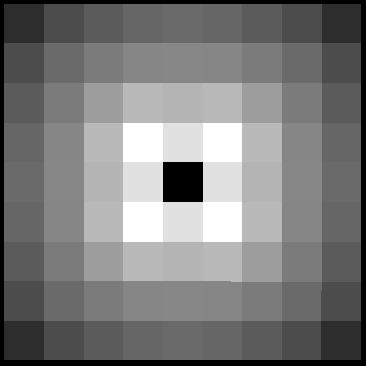} & \includegraphics[scale=.3]{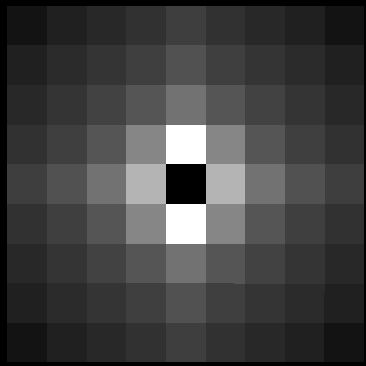}  \\
  tensor field & dispersion & curving  & total distortion
\vspace{2mm} \\
  &  \includegraphics[scale=.3]{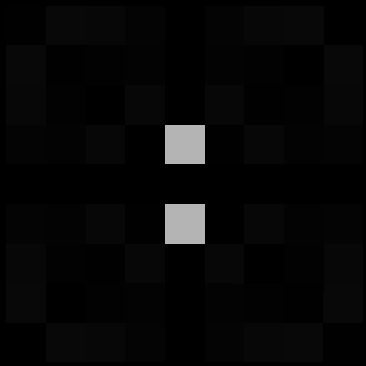}   & \includegraphics[scale=.3]{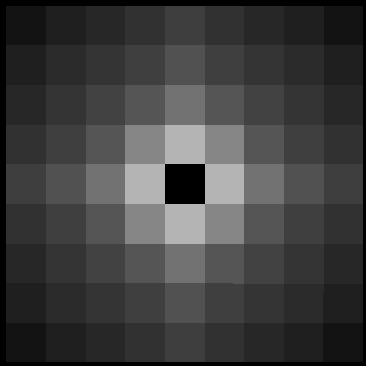} & \includegraphics[scale=.3]{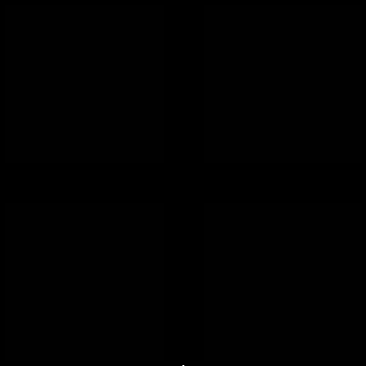} \\
  &  splay & bend & twist \\
\end{tabular}
\begin{tabular}{c  c  c c}
  \includegraphics[scale=.18]{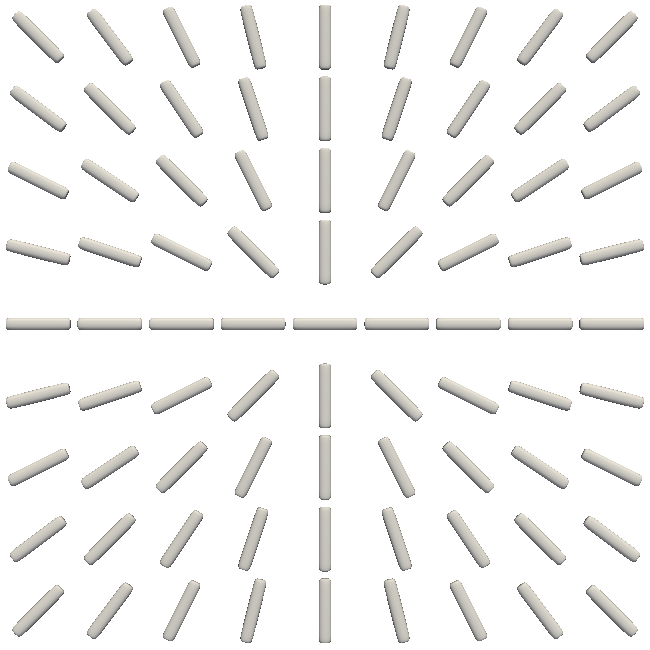}  & \includegraphics[scale=.3]{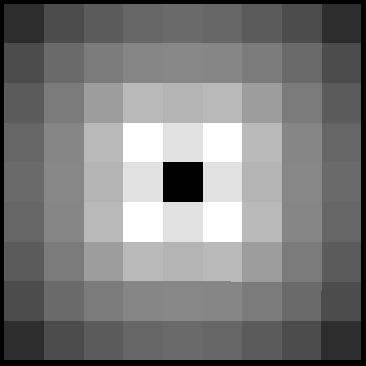} & \includegraphics[scale=.3]{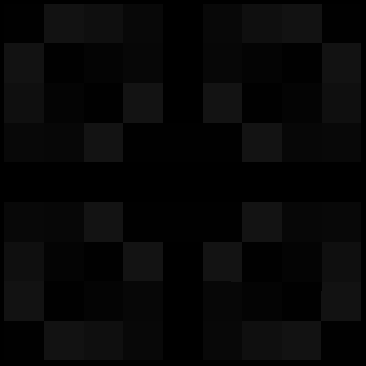} & \includegraphics[scale=.3]{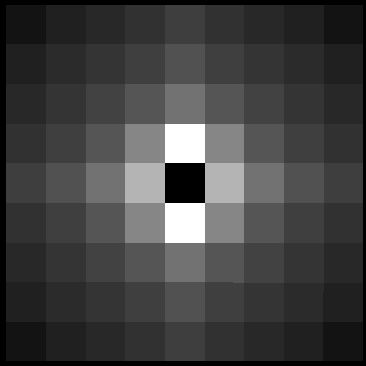}\\
  tensor field & dispersion & curving & total distortion 
\vspace{2mm} \\
  & \includegraphics[scale=.3]{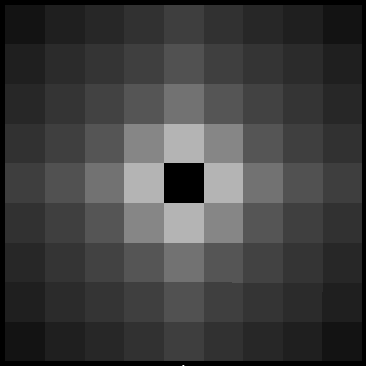}   & \includegraphics[scale=.3]{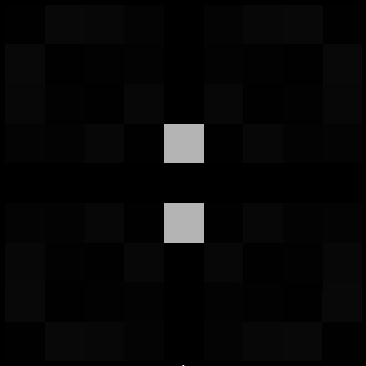} & \includegraphics[scale=.3]{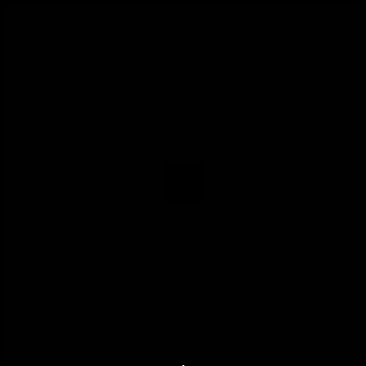} \\
  &  splay & bend & twist 
\end{tabular}
\caption{\label{fig:distortion_exp_circle}Dispersion, curving~\citep{Savadjiev_NI2010}, and the proposed four orientational distortion indices calculated from two tensor fields.}
\end{figure*}

We would like to compare the four orientational distortion indices with the curving and dispersion indices proposed for tensor fields in~\cite{Savadjiev_NI2010}. 
The tensor field in Fig.~\ref{fig:distortion_exp_compare} was used in~\cite{Savadjiev_NI2010}. 
It has three areas where the tensors rotate about its three eigenvectors, respectively. 
From bottom to top, the mode of the tensors changes linearly. 
Fig.~\ref{fig:distortion_exp_circle} shows two other synthetic tensor fields used in~\cite{Savadjiev_NI2010}. 
Fig.~\ref{fig:distortion_exp_compare} and Fig.~\ref{fig:distortion_exp_circle} demonstrate all six scalar indices. 
It can be seen that 
1) the splay index is similar to the dispersion index;
2) the bend index is similar to the curving index;
3) the four orientational distortion indices are independent of tensor shapes, while curving and dispersion indices are dependent on the tensor mode; 
4) when the principal directions are well aligned (e.g., in the left part of the tensor field in Fig.~\ref{fig:distortion_exp_compare}), all distortion indices are close to zero, because they are calculated based on the spatial difference of principal directions; 
5) the definition and calculation of distortion indices are rotationally invariant. 
Note that the singular values of these scalar indices around the central point in the tensor fields in Fig.~\ref{fig:distortion_exp_circle} are attributable to the singularity of the tensor orientation in the central point. 
Although the proposed splay and bend indices have similar contrast compared with the dispersion and curving indices that are only for tensor fields, 
the proposed distortion indices can be defined for both tensor fields and ODF fields. 
Moreover, the proposed splay and bend indices are independent of tensor shapes, while the dispersion and curving indices are related with tensor shapes.

\subsection{Real Data Experiments}

The experimental data are from Human Connectome Project (HCP), Q3 release~\citep{sotiropoulos_HCP_NI13,van:NI2013:HCP}. 
This data set is acquired using three shells, with 90 staggered directions per shell, and at $b=1000,2000$, and $3000\,\text{s}/\text{mm}^2$. 

\begin{figure*}[t!]
\small
\centering

  \begin{subfigure}[b]{0.24\textwidth}
\includegraphics[width=\textwidth]{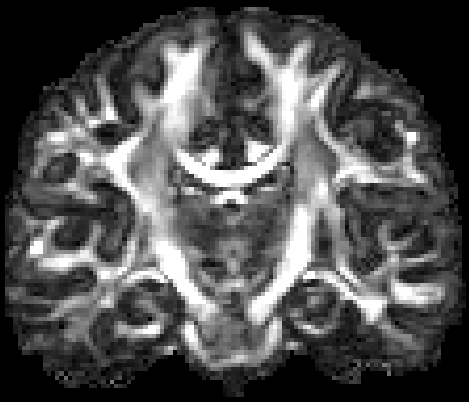}
                    \caption*{$\kappa$ from NODDI}
  \end{subfigure}
  \begin{subfigure}[b]{0.24\textwidth}
\includegraphics[width=\textwidth]{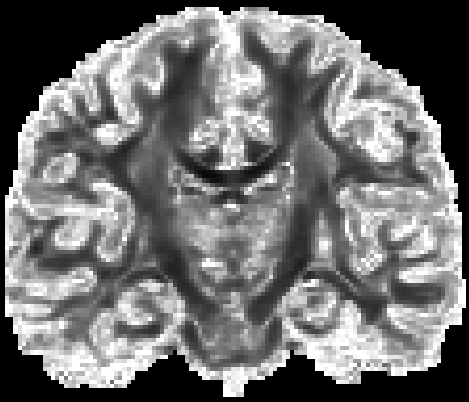}
                    \caption*{$\text{OD}_{\text{w}}$ from NODDI ($10\kappa$)}
  \end{subfigure}
  \begin{subfigure}[b]{0.24\textwidth}
\includegraphics[width=\textwidth]{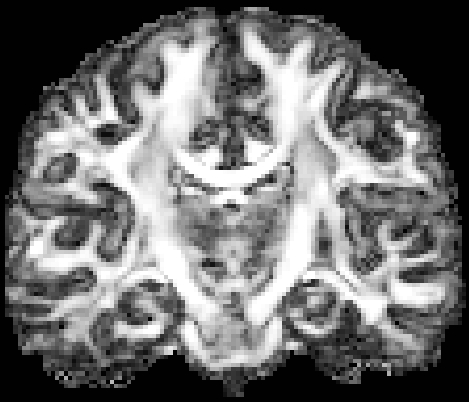}
                    \caption*{$\text{OO}$ from NODDI ($25\kappa$)}
  \end{subfigure}
  \begin{subfigure}[b]{0.24\textwidth}
\includegraphics[width=\textwidth]{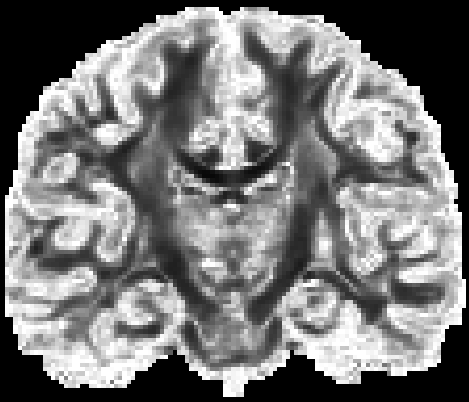}
                    \caption*{$\text{OD}$ from NODDI ($25\kappa$)}
  \end{subfigure}
  \begin{subfigure}[b]{0.24\textwidth}
\includegraphics[width=\textwidth]{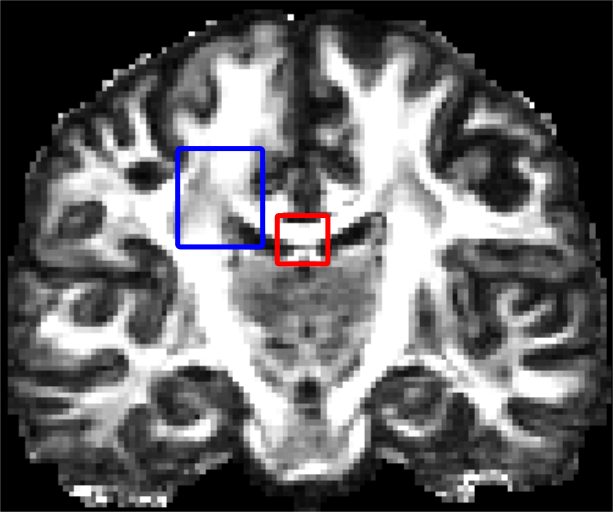}
                    \caption*{FA from tensors}
  \end{subfigure}
  \begin{subfigure}[b]{0.24\textwidth}
\includegraphics[width=\textwidth]{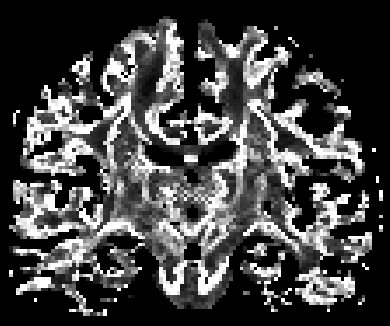}
                    \caption*{total distortion from fODFs}
  \end{subfigure}
  \begin{subfigure}[b]{0.24\textwidth}
\includegraphics[width=\textwidth]{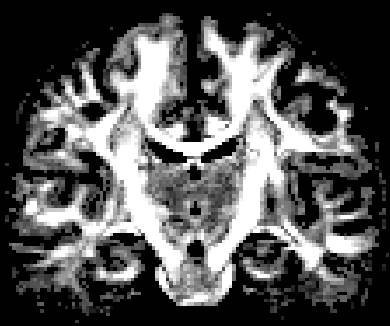}
                    \caption*{OO from fODFs}
  \end{subfigure}
  \begin{subfigure}[b]{0.24\textwidth}
\includegraphics[width=\textwidth]{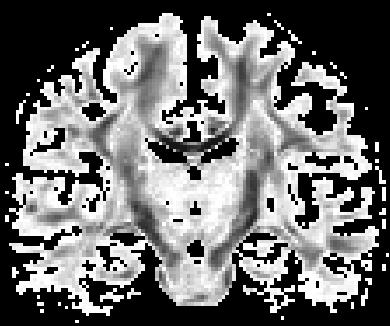}
                    \caption*{OD from fODFs}
  \end{subfigure}

  \caption{\label{fig:HCP_oo}
  First row: NODDI results for multi-shell HCP data, 
  where $\kappa$ is estimated from NODDI model, 
  $\text{OD}_\text{w}$ is calculated from~\EEqref{eq:OD_w} by using $10\kappa$, 
  and $\text{OO}$ and $\text{OD}$ are calculated from~\EEqref{eq:OO_watson} by using $25\kappa$. 
  Second row: DTI and NNSD results for HCP data, where OO, OD and total orientational distortion are calculated from fODFs by NNSD. 
  The close-up views of red and blue regions are in Fig.~\ref{fig:HCP_region1} and~\ref{fig:HCP_region2}.
  }
\end{figure*}

We perform NODDI on the HCP multi-shell data using the released matlab toolbox by the authors~\citep{zhang_NODDI_NI2012}. 
The first row in Fig.~\ref{fig:HCP_oo} shows the parameter maps by the NODDI toolbox, i.e., the $\kappa$ map and $\text{OD}_\text{w}$ map. 
It should be noted that $\text{OD}_\text{w}$ is calculated based on $10\kappa$ and~\EEqref{eq:OD_w} in the author-released toolbox. 
Fig.~\ref{fig:HCP_oo} also shows OO and OD from NODDI, based on the closed form in~\EEqref{eq:OO_watson} with a scaled $25\kappa$. 
The scale on $\kappa$ is used for a better contrast in the obtained dispersion index map. 
The obtained $\kappa$ map has intensities that are less than $0.4$ in most voxels. 
Thus, as shown in Fig.~\ref{fig:OO_tensor_watson}, $10\kappa$ obtains the range $[0,4]$ which is good for $\text{OD}_\text{w}$, 
and $25\kappa$ obtains the range $[0,10]$ which is good for $\text{OD}$. 
The two dispersion index maps calculated in two ways from $\kappa$ visually have similar contrast. 

\begin{figure*}[t!]
\begin{tabular}{c  c  c }
\includegraphics[width=0.32\textwidth]{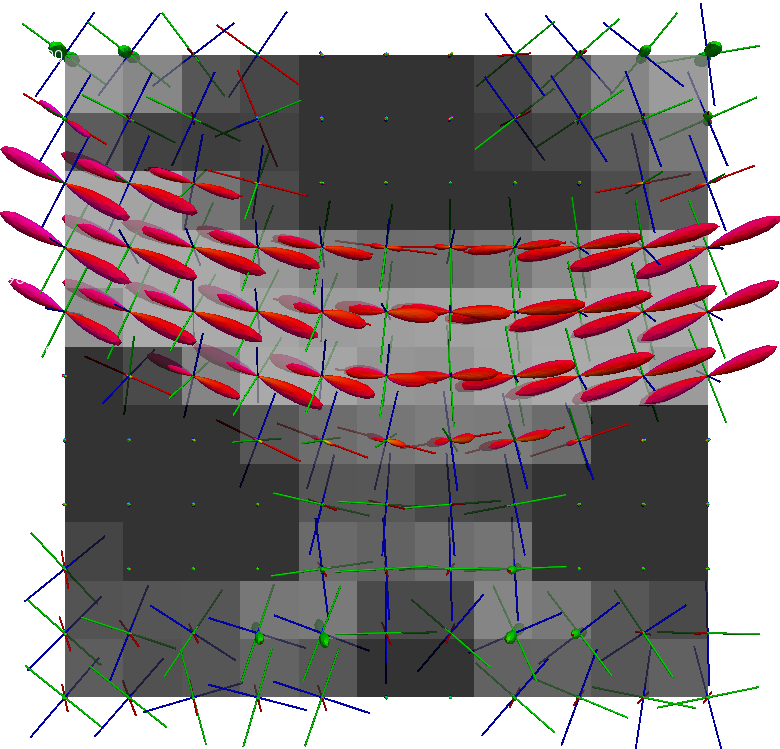}  & \includegraphics[width=0.32\textwidth]{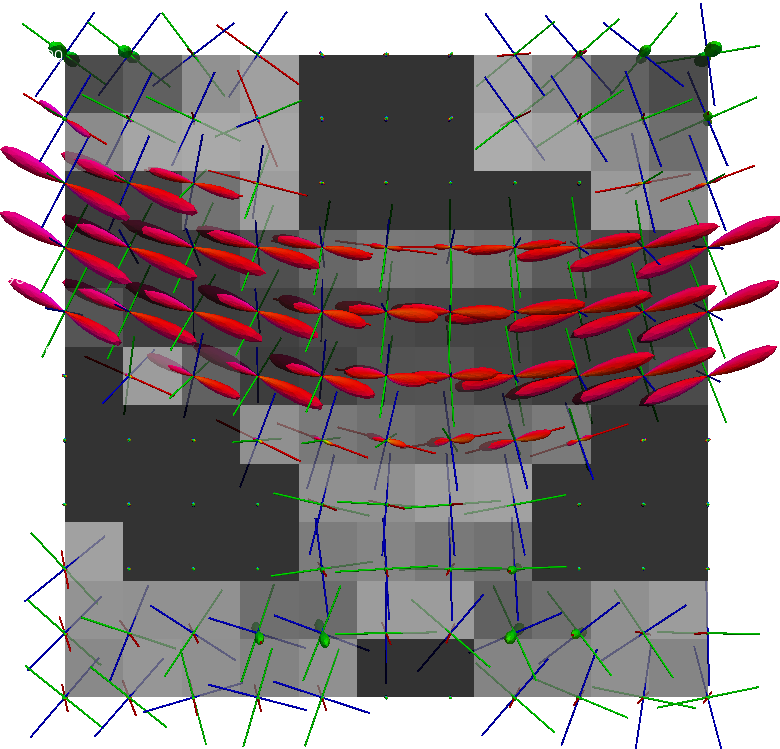}  & \includegraphics[width=0.32\textwidth]{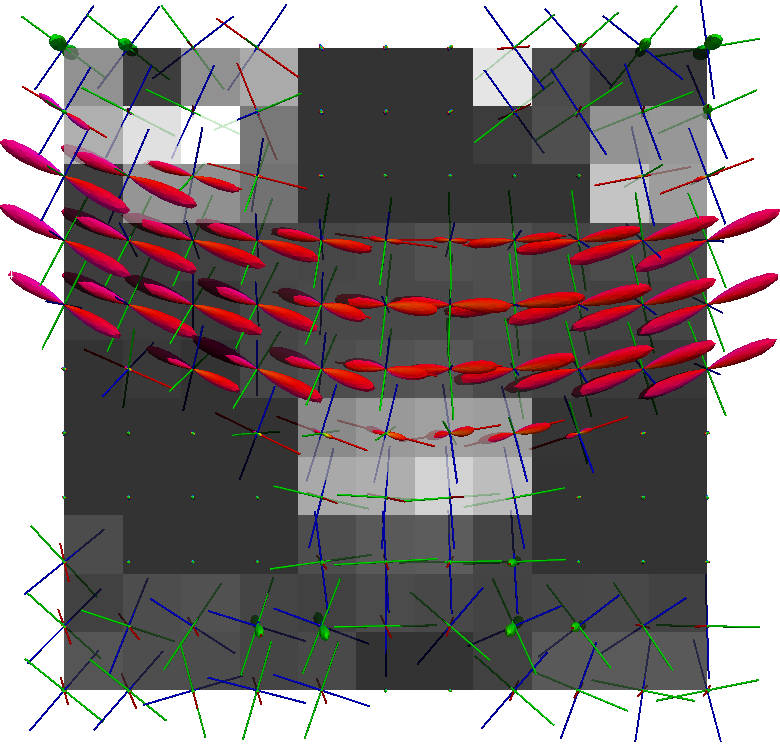} \\
  OO & OD & total distortion\\ 
\includegraphics[width=0.32\textwidth]{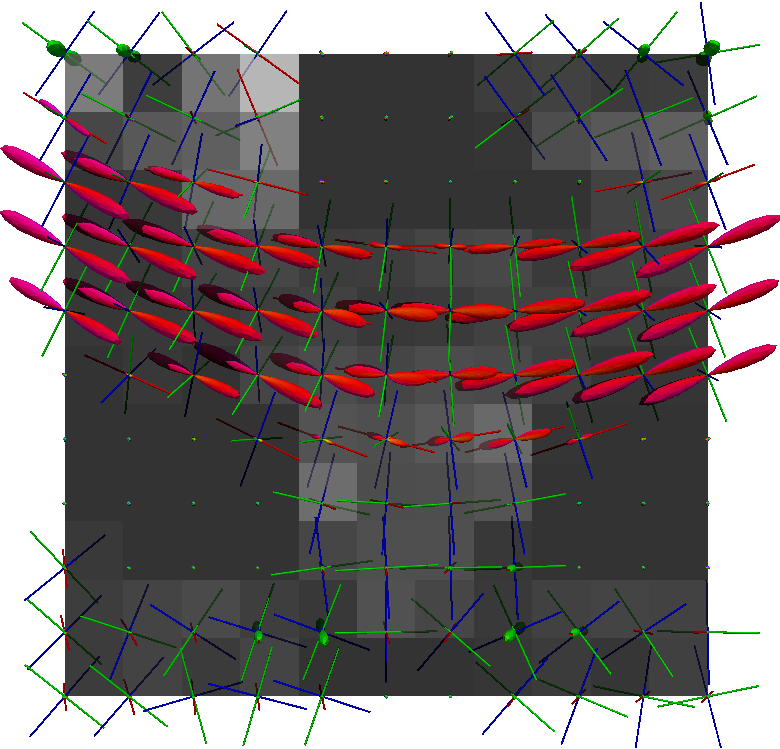}  & \includegraphics[width=0.32\textwidth]{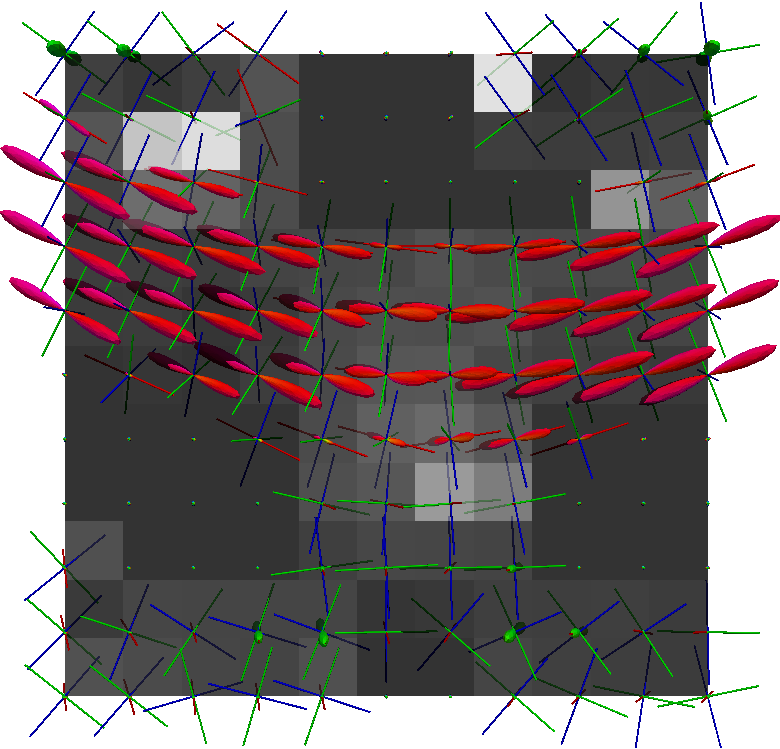} & \includegraphics[width=0.32\textwidth]{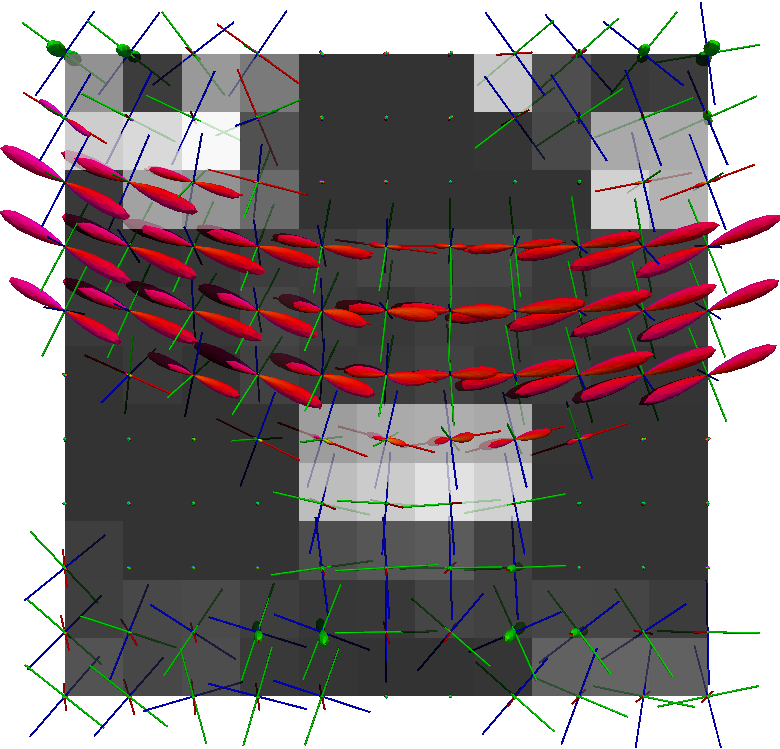}\\
    splay  & bend & twist 
\end{tabular}
  \caption{\label{fig:HCP_region1}fODFs, local orthogonal frames, and six scalar indices for the red region in Fig.~\ref{fig:HCP_oo}. 
  Local orthogonal frames are visualized using tubes in red, green, and blue colors. 
  The scalar indices are shown in the background. This region is also used in the DFA pipeline in Fig.~\ref{fig:DFA}
  }
\end{figure*}

\begin{figure*}[t!]
\begin{tabular}{c  c  c }
\includegraphics[width=0.32\textwidth]{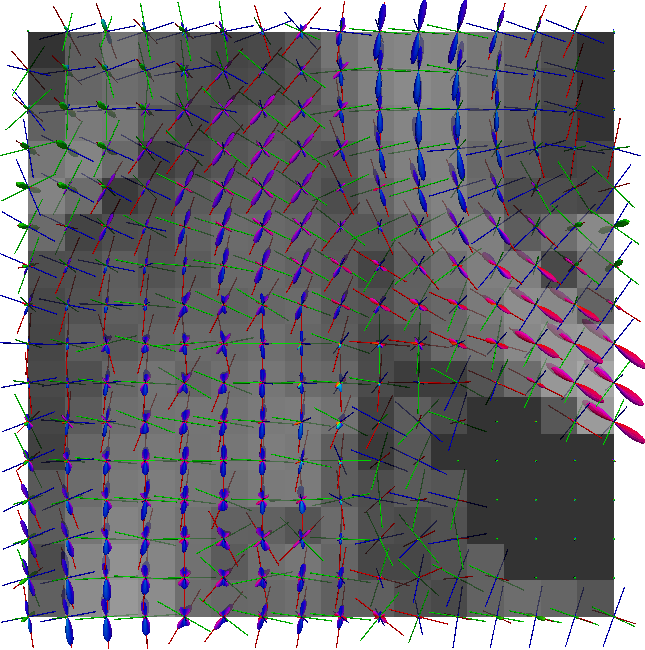}  & \includegraphics[width=0.32\textwidth]{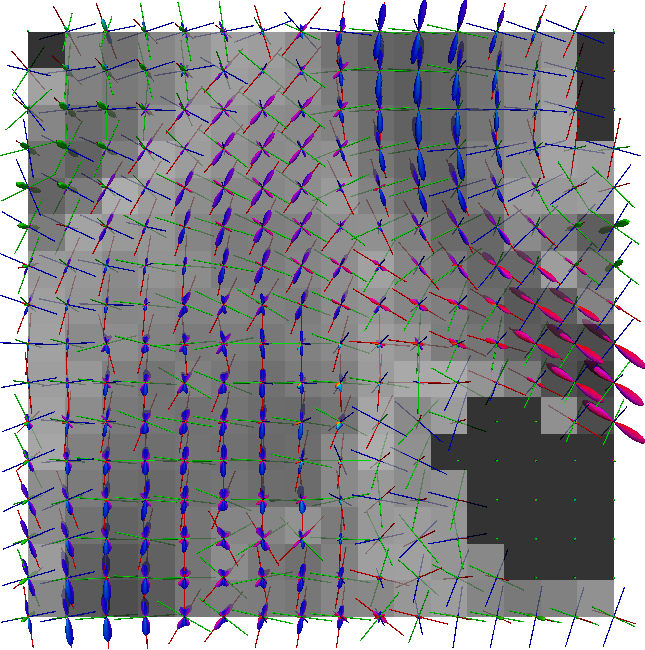} & \includegraphics[width=0.32\textwidth]{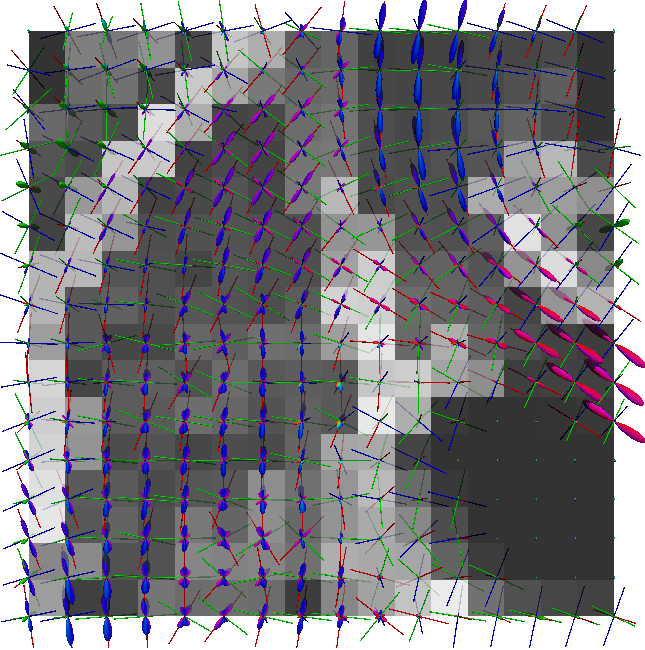}\\
  OO & OD & total distortion\\ 
\includegraphics[width=0.32\textwidth]{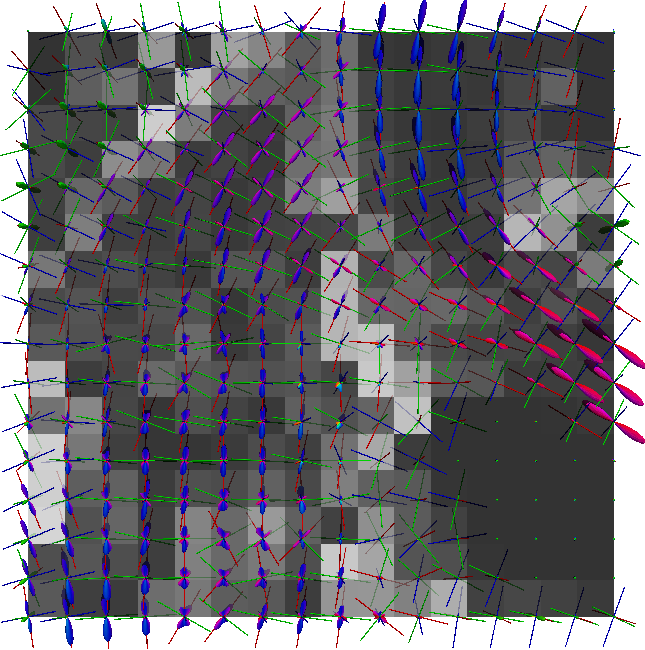}  & \includegraphics[width=0.32\textwidth]{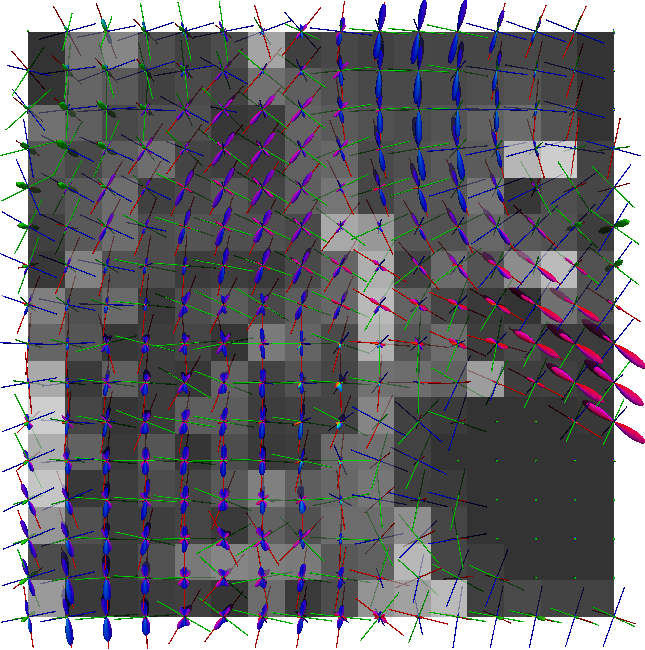} & \includegraphics[width=0.32\textwidth]{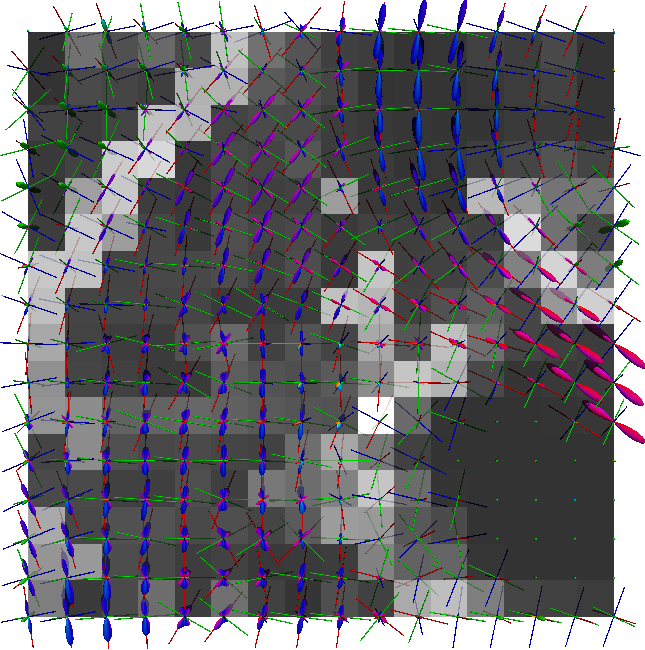}\\
    splay  & bend & twist 
\end{tabular}
  \caption{\label{fig:HCP_region2}fODFs, local orthogonal frames, and six scalar indices for the blue region in Fig.~\ref{fig:HCP_oo}. 
  Local orthogonal frames are visualized using tubes in red, green, and blue colors. 
  The scalar indices are shown in the background. 
  }
\end{figure*}

Non-negative spherical deconvolution (NNSD)~\citep{cheng_NI2014} is performed to estimate non-negative fiber ODFs from three-shell DWI data. 
NNSD works for multi-shell data. 
It is more robust to noise, and the obtained fiber ODFs (fODFs) in isotropic regions are closer to the isotropic spherical PDF, compared with conventional constrained spherical deconvolution~\citep{tournier_NI2007}. 
After obtaining the fODFs by NNSD, the peaks are detected from the estimated fODFs with GFA larger than $0.3$, as described in Section~\ref{sec:frame}. 
OO and OD are calculated from the spherical harmonic representation of fODFs along their principal peaks as shown in Algorithm~\ref{alg:OO_OD}. 
The second row in Fig.~\ref{fig:HCP_oo} demonstrates FA from tensors estimated by DTI, OO and OD from fODFs estimated by NNSD, 
and the total distortion map estimated from the local orthogonal frames of fODFs. 
Fig.~\ref{fig:HCP_region1} and~\ref{fig:HCP_region2} show the close-up views of 
fODFs, local orthogonal frames, and the six proposed indices for the red and blue regions in Fig.~\ref{fig:HCP_oo}, 
where the region shown in~\ref{fig:HCP_region1} is also visualized in the DFA pipeline in Fig.~\ref{fig:DFA}. 
The fODF glyphs are colored by using its sampled directions. 
The three orientations in the local orthogonal frame in each voxel are visualized by using three tubes in red, green, and blue colors respectively. 
There is no local orthogonal frame in some voxels because those voxels have GFA values lower than $0.3$. 
These figures show the following:  
1) OO is high in anisotropic areas with well-aligned directions, while OD is high in isotropic or crossing areas. 
2) The four orientational distortion indices are low in areas with well aligned principal directions, and zero in isotropic voxels without peaks. 
Distortion indices are high in voxels where the principal directions in its local neighborhood change largely. 
3) The central voxels in red region is the crossing area of the Corpus Callosum from left to right and Fornix that goes through the coronal slice. 
The twist index showed high value in this crossing area as expected. 

OO and OD by NODDI are different from OO and OD by NNSD in Fig.~\ref{fig:HCP_oo}. 
We propose OO and OD as general properties (i.e., the degree of aligment and dispersion along peaks) for general ODFs, like GFA for ODFs, independent of diffusion signal models.
OO and OD can be calculated from ODFs estimated by the NODDI model~\citep{zhang_NODDI_NI2012}, the tensor model in DTI~\citep{Basser1994}, and various spherical deconvolution methods~\citep{tournier_NI2007,cheng_NI2014}, etc. 
In this sense, we claim that the proposed OD (and OO) inspired from liquid crystals is more general than the dispersion index in NODDI that only works for Watson distributions. 

\begin{figure*}[t!]
  \centering
\includegraphics[width=0.265\textwidth]{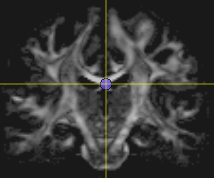} \\
ROI on OO map \\
\begin{tabular}{c  c  c }
\includegraphics[width=0.32\textwidth]{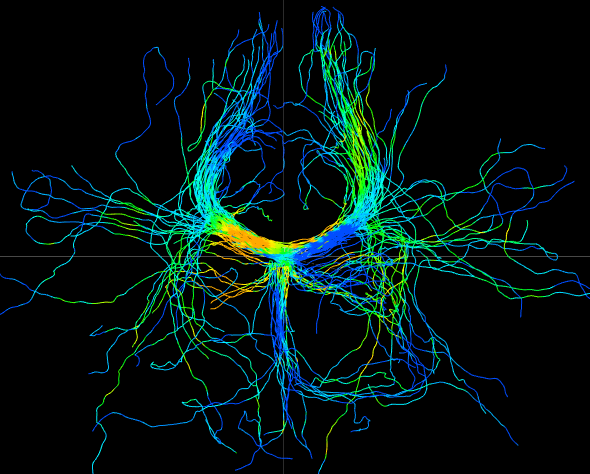}  & \includegraphics[width=0.32\textwidth]{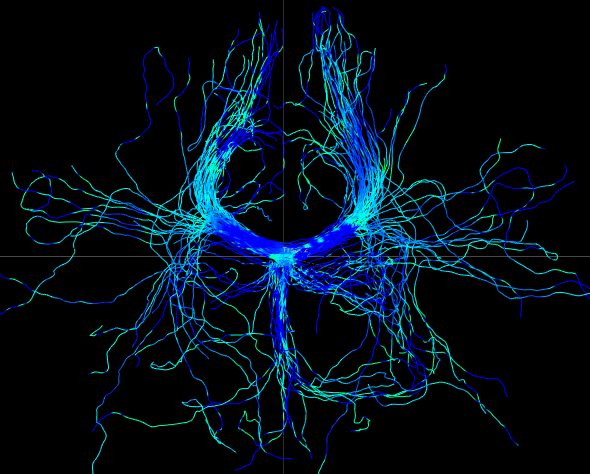} & \includegraphics[width=0.32\textwidth]{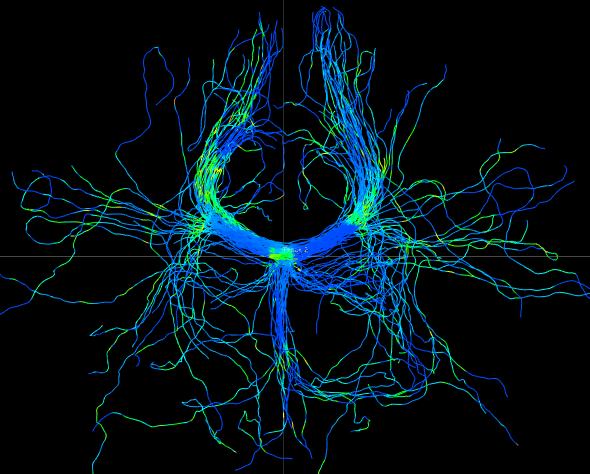}\\
  OO & OD &  total distortion \\ 
\includegraphics[width=0.32\textwidth]{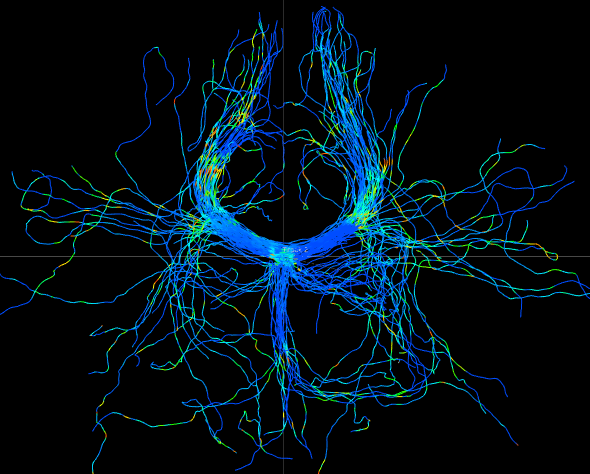}  & \includegraphics[width=0.32\textwidth]{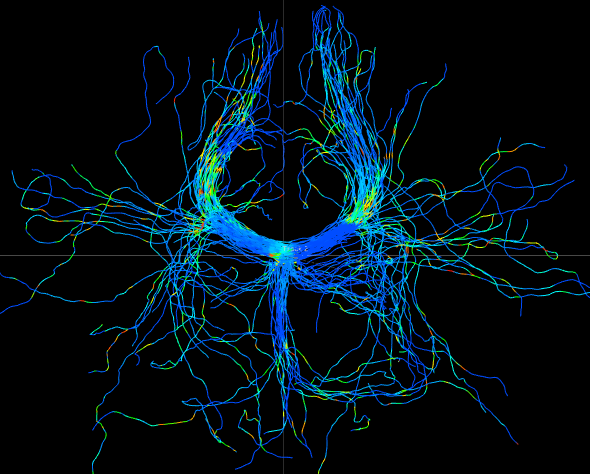} & \includegraphics[width=0.32\textwidth]{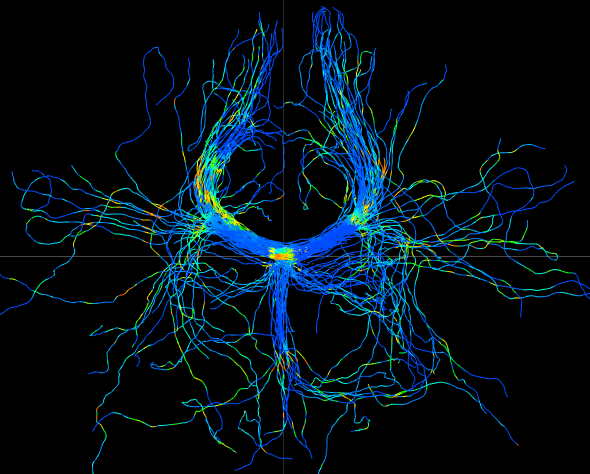}\\
    splay  & bend & twist 
\end{tabular}
  \caption{\label{fig:HCP_roi2_track}Fiber tracts cross a given ROI are colored by the six indices, respectively.}
\end{figure*}

\begin{figure*}[t!]
  \centering
\includegraphics[width=0.265\textwidth]{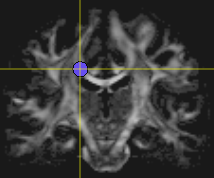} \\
ROI on OO map \\
\begin{tabular}{c  c  c }
\includegraphics[width=0.32\textwidth]{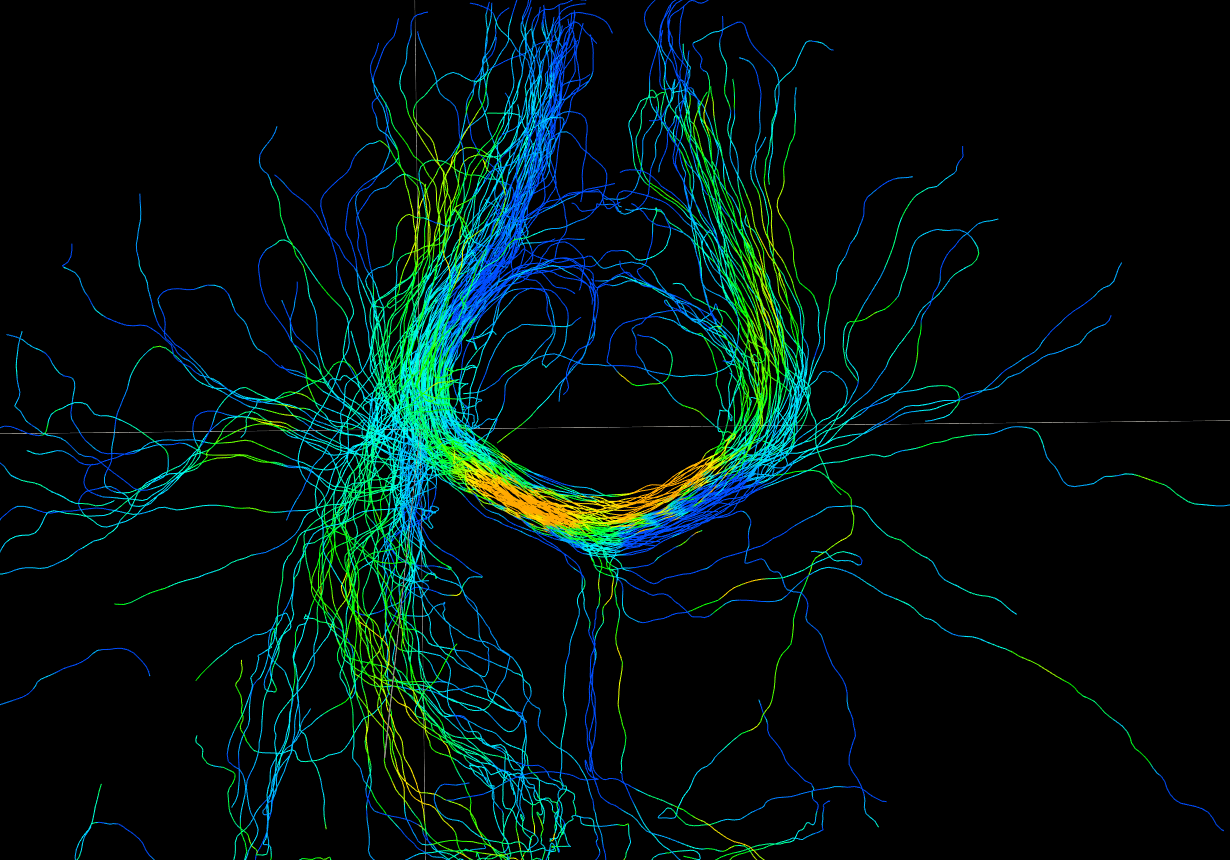}  & \includegraphics[width=0.32\textwidth]{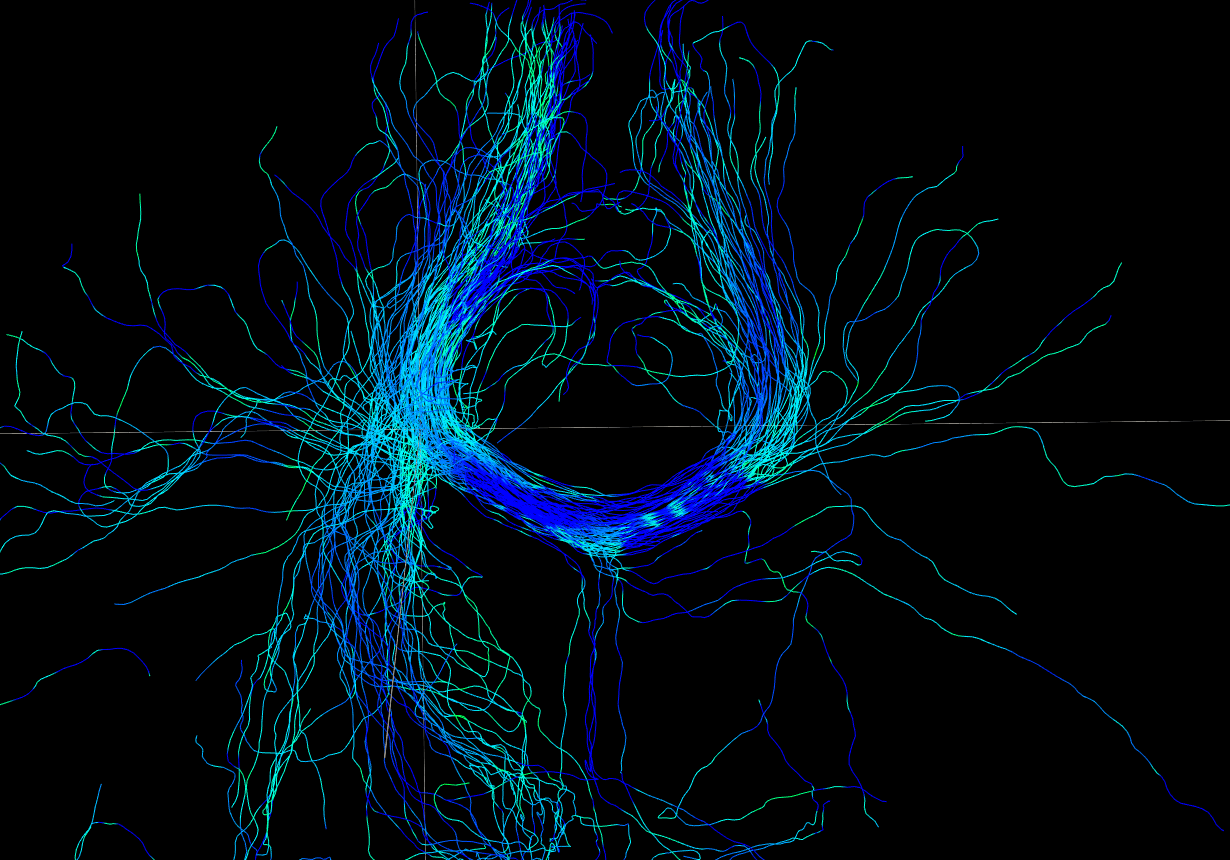} & \includegraphics[width=0.32\textwidth]{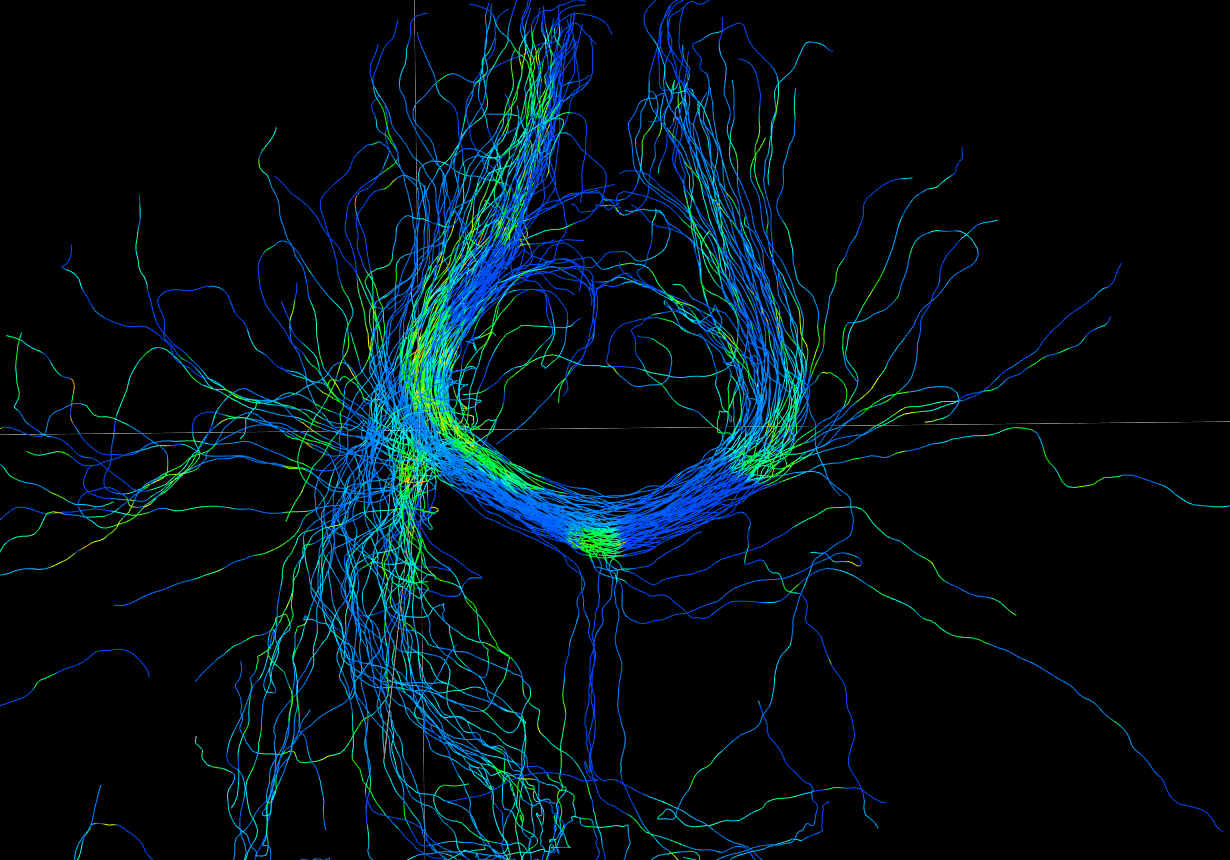}\\
  OO & OD &  total distortion \\ 
\includegraphics[width=0.32\textwidth]{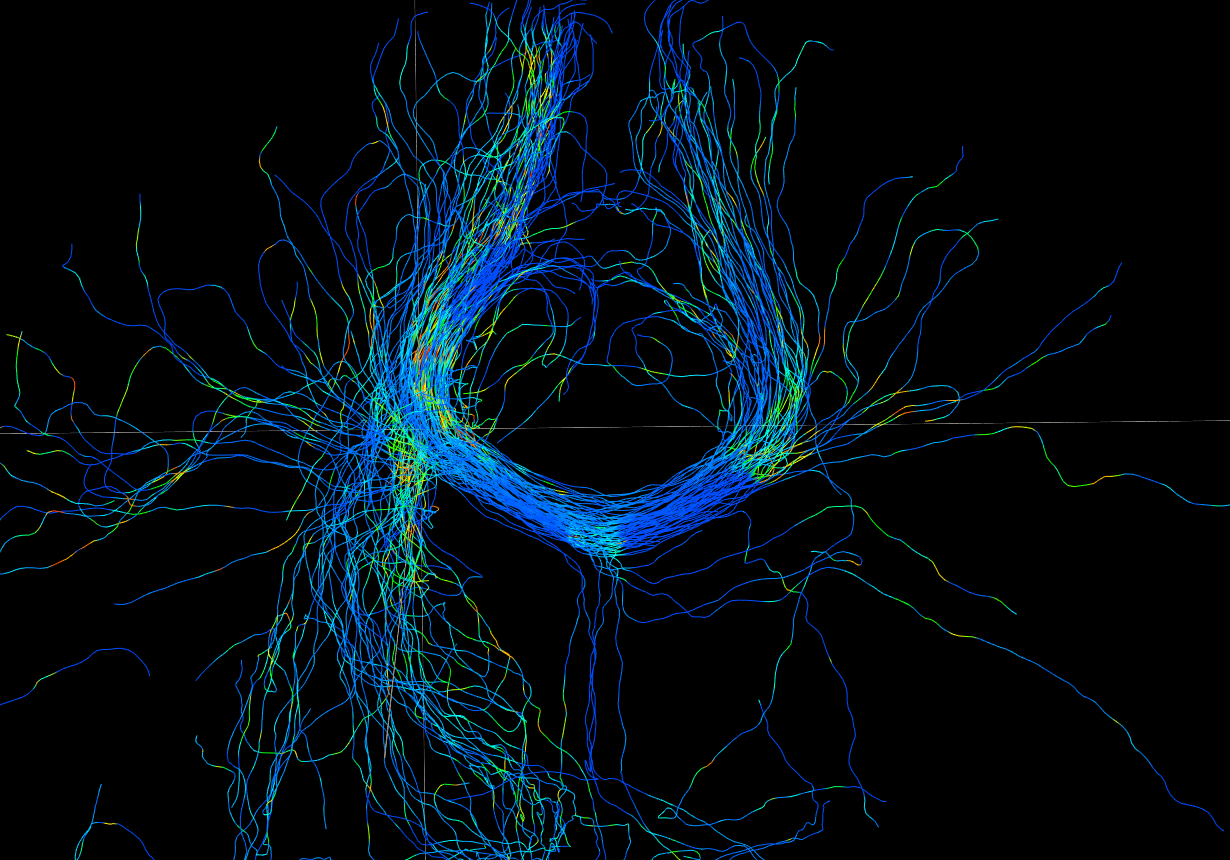}  & \includegraphics[width=0.32\textwidth]{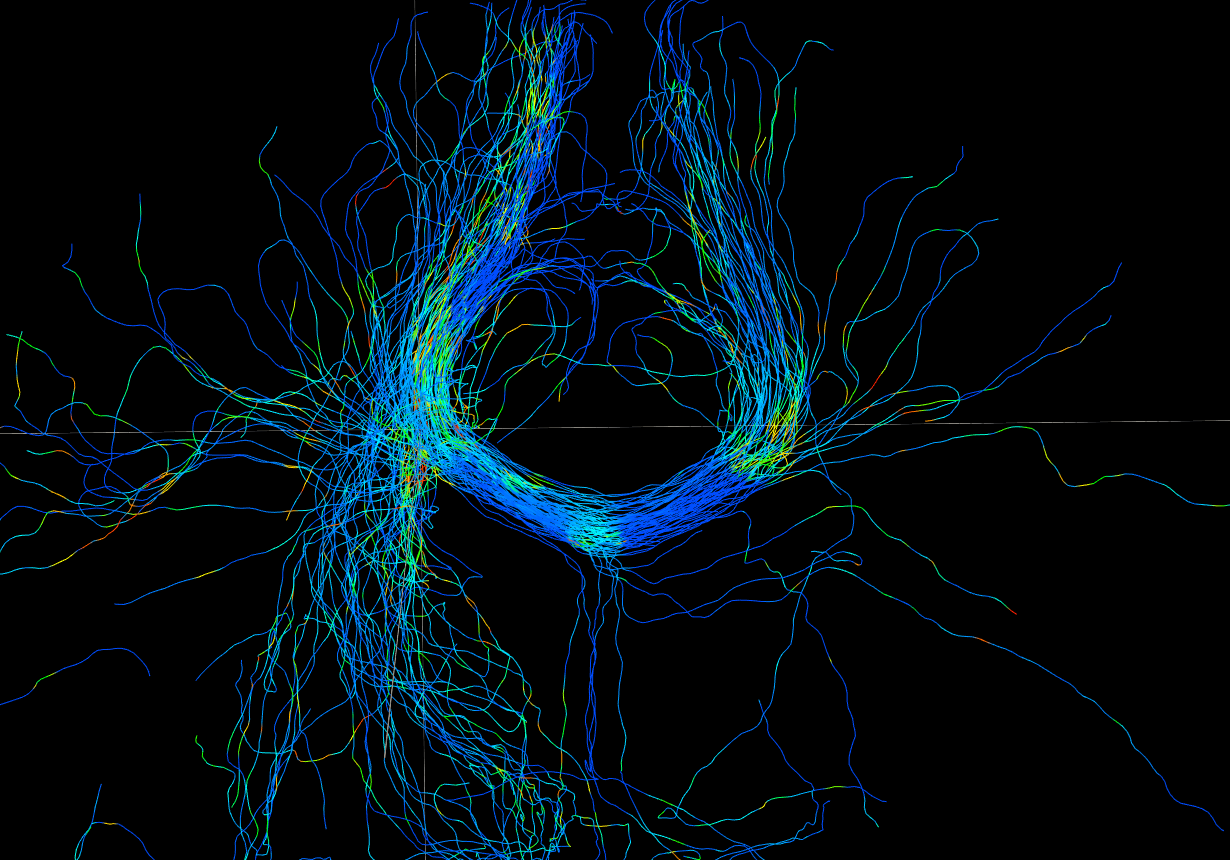} & \includegraphics[width=0.32\textwidth]{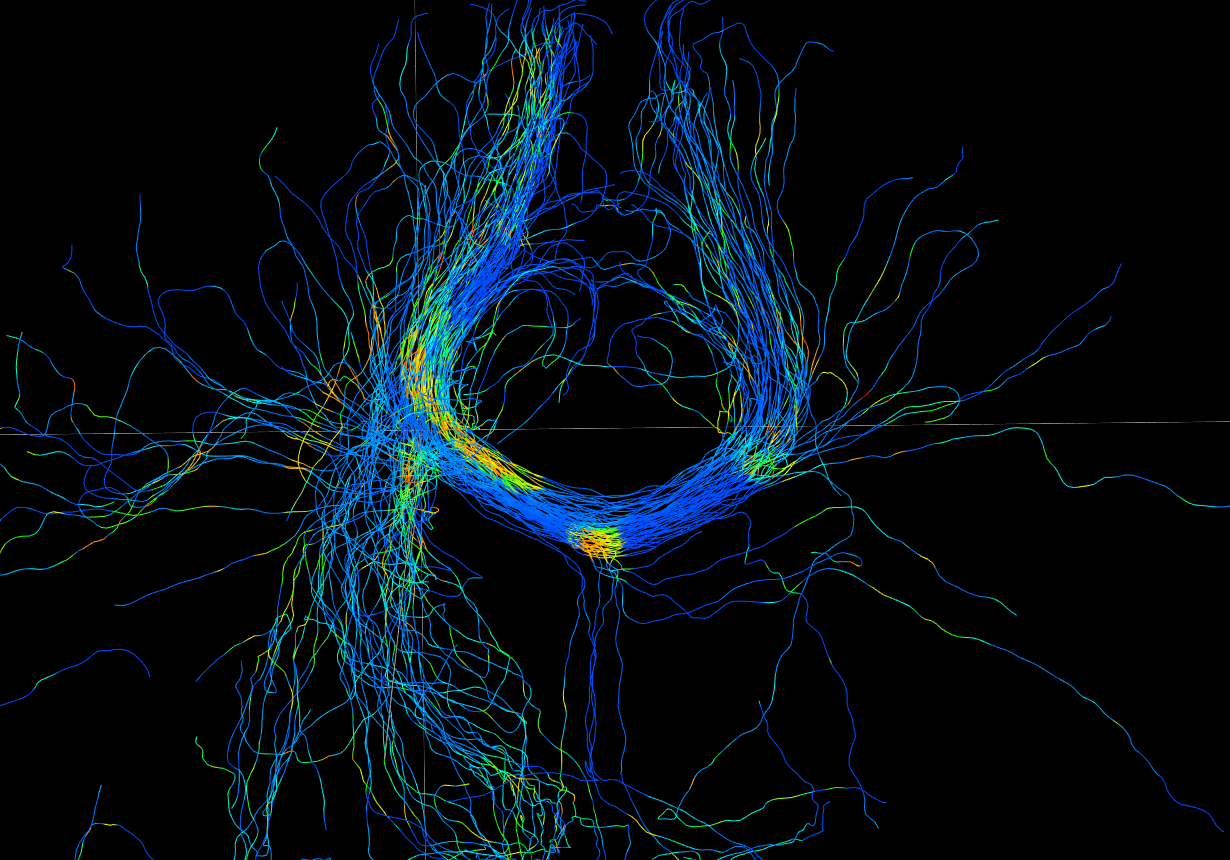}\\
    splay  & bend & twist 
\end{tabular}
  \caption{\label{fig:HCP_roi_track}Fiber tracts cross a given ROI colored by the six indices, respectively.}
\end{figure*}

We perform whole brain streamline tractography on the estimated fODF field using mrtrix~\citep{tournier_mrtrix_12}~\footnote{\href{http://www.mrtrix.org}{http://www.mrtrix.org/}}. 
The voxels with GFA larger than $0.3$ are used as seed voxels to generate $10000$ tracts by using \texttt{tckgen} in mrtrix. 
All other parameters are default parameters in mrtrix. 
The obtained fiber tracts are then visualized by using trackvis~\footnote{\href{http://trackvis.org}{http://trackvis.org}}. 
Fig.~\ref{fig:HCP_roi2_track} and~\ref{fig:HCP_roi_track} demonstrate the tracts respectively cross two given ball ROIs. 
The tracts are colored by using the proposed six scalar indices. 
Note that the proposed scalar indices are calculated based on estimated fODFs, not based on fiber tracts.
It can be seen that 
1) OO is high in areas with well aligned fibers, while OD is high in crossing areas and distortion areas; 
2) distortion indices are low when fibers are well aligned; 
3) the total distortion index is high in areas with highly curved fibers or crossing fibers. 
4) although splay, bend, twist indices may be separable (e.g., one is large while another one is close to zero) in synthetic data, 
in real data, these three types of distortions normally occur together, especially for bending and splaying. 
5) the ROI in Fig.~\ref{fig:HCP_roi2_track} is the crossing area of the Corpus Callosum and the Fornix, where all distortion indices have high values, especially for twist and total distortion indices. 
This finding agrees with Fig.~\ref{fig:HCP_region1}. 

\section{Discussion}
\label{sec:discussion}

\subsection{Effect of the Spatial Resolution on Directional Derivatives and Distortion Indices}

The definition of the spatial directional derivative in~\EEqref{eq:derivative_dir} is for continuous spatial domain. 
The unit of the spatial directional derivative $\frac{\partial \uu_i}{ \partial \Vv}$ in~\EEqref{eq:derivative_dir_v} is $[\si{mm^{-1}},\si{mm^{-1}},\si{mm^{-1}}]$, 
considering the numerator is a director with no unit and the unit of the denominator is $\si{mm}$. 
Thus, the four distortion indices have the unit of $\si{mm^{-1}}$. 
The spherical function field and peak field in diffusion MRI are obtained in a discrete integer lattice. 
In Algorithm~\ref{alg:derivative}, rotation matrices $\{\BM{R}_i\}$ are calculated based on directors in neighborhood voxels $\uu_1(\Vx+\VV{o}_i)$ and $\uu_1(\Vx-\VV{o}_i)$. 
Thus, these three rotation matrices are dependent on spatial resolution of the diffusion image. 
So are the central difference approximation of the spatial gradient in~\EEqref{eq:central_diff}, and the distortion indices calculated based on directional derivatives. 
Consider the twisting synthetic tensor image in Fig.~\ref{fig:distortion_exp} as an example, where the tensor from left to right rotates about the $x$-axis by the angle $\pi$. 
If there are $N+1$ tensors from left to right, then the spatial gradient along the $x$-axis is the rotation matrix with a rotation angle of $\pi/N$. 
With the local linear change assumption of rotation angles, finer spatial resolution will produce smaller rotation angles of central differences in the three rotation matrices, 
which results in smaller spatial gradients, directional derivatives and smaller distortion indices. 

An improved version of calculation of the spatial directional derivatives in Algorithm~\ref{alg:derivative} is to consider the spatial resolution of the image as the step size of the central difference. 
The image resolution should be used to normalize the rotation angles in the rotation matrices. 
We can approximate the spatial gradient using all directors within a given physical resolution, e.g., $3\times 3\times 3~\si{mm}$. 
If the image resolution is also $3~\si{mm}$ isotropic, then we just use the central difference described in Section~\ref{sec:derivative_dir}. 
If the image resolution is $1.5~\si{mm}$ isotropic, then we can use a mean of rotation matrices from two central differences. 
One rotation matrix is generated by $w\Vv(\Vx+\VV{o}_i)$, $w\Vv(\Vx-\VV{o}_i)$, then we keep the rotation axis, but scale the rotation angle by $2$, based on the local linear rotation angle assumption.
The other one is generated by $w\Vv(\Vx+2\VV{o}_i)$, $w\Vv(\Vx-2\VV{o}_i)$. 
The Riemannian mean is used to calculate the mean of rotation matrices~\citep{moakher_SIAM2002}.   
In this way, for the twisting synthetic tensor image in Fig.~\ref{fig:distortion_exp}, the rotation matrix representation of the central difference along the $x$-axis remains the same for different spatial resolutions of the synthetic image. 
Note that the local linear rotation change assumption only holds in a small local neighborhood, not for a large spatial scale. 
With the correct consideration of image resolution in calculation, 
the image resolution effect can be reduced in numerical calculation of the proposed distortion indices.

\subsection{DFA For General Spherical Functions Without Antipodal Symmetry}
\label{sec:DFA_vector}

Considering spherical functions obtained in diffusion MRI are normally antipodally symmetric, the detected principal directors and local orthogonal frames all have sign ambiguity. 
Thus, the proposed DFA is mainly for director data analysis. 
However, if the reconstructed spherical function in a voxel (e.g., an ODF) is not antipodally symmetric, the detected principal peak field is a traditional vector field. 
Then DFA can be modified for vector field analysis. 

The difference of two vectors, the spatial gradient, and the spatial derivative of a vector field are all well defined. 
Note that the definitions of OO and OD in Section~\ref{sec:order} work for a general spherical function $f(\uu)$ without requiring antipodal symmetry, although $\text{OO}(\Vn)$ is always antipodally symmetric by definition. 
If the detected peaks have no sign ambiguity, then the three orientations in the local orthogonal frame can all be traditional vectors. 
The first orientation is the principal peak at voxel $\Vx$. 
As described in Section~\ref{sec:frame}, after projecting all peaks onto the orthogonal plane, we obtain vectors $\{w(\Vy,\Vx) f(\uu_i,\Vy) (\uu_i-(\uu_i^T\uu_1) \uu_1)\}$ without sign ambiguity. 
Then, the second orientation can be set as the orientation in the orthogonal plan with maximal value among $|w(\Vy,\Vx) f(\uu_i,\Vy)|\|\uu_i-(\uu_i^T\uu_1) \uu_1)\|$, $\forall i$, or the mean orientation of the projected vectors in the orthogonal plane, 
and the third orientation is the cross product of the first and second orientations. 
Finally, the four orientational distortion indices in~\EEqref{eq:splay}, \EEqref{eq:bend}, \EEqref{eq:twist}, and~\EEqref{eq:distortion} can still be used for the spatial derivatives of the vector field, which are actually functions of Maurer-Cartan connections in the moving frame method. 

\section{Conclusion}
\label{sec:conclusion}

In this paper, we propose a unified mathematical framework, called Director Field Analysis (DFA), to analyze a spherical function field and its extracted peak field. 
See Fig.~\ref{fig:DFA} for an overview of the DFA pipeline. 
First, in DFA, we detect peaks from the spherical function field, and define the Orientational Order (OO) and the Orientational Dispersion (OD) indices in voxels or within spatial regions. 
Closed-form solutions of OO and OD are obtained for some specific spherical functions. 
We propose OO and OD as properties for general ODFs along peaks, independent of diffusion signal models. 
Second, we define a local orthogonal frame in each voxel exhibiting anisotropic diffusion, where the principal peak is its first axis, and the other two axes describe the local spatial change directions of principal peaks. 
Third, from the extracted local orthogonal frames in voxels, DFA estimates three distortion indices (splay, bend, twist) that are able to distinguish three types of distortions and a total orientational distortion index. 
To our knowledge, this paper is the first work to \emph{quantitatively} describe orientational distortion (splay, bend, and twist) in general spherical function fields from DTI or HARDI data. 
The experiments demonstrate the following: 
1) The proposed OO is another type of anisotropy index for spherical functions, and OD is more general and natural than the previous dispersion index proposed in NODDI~\citep{zhang_NODDI_NI2012}, which only works for Watson distributions. 
2) The proposed splay and bend indices can be seen as a generalization of the dispersion and curving indices in~\cite{Savadjiev_NI2010}, considering they have similar contrast in the same tensor field. 
The proposed four orientational distortion indices work not only for tensors but also for general spherical functions. 
3) The proposed distortion indices demonstrate good sensitivity for the three different types of orientational distortion in Fig.~\ref{fig:distortion}. 
4) Orientational distortion indices normally have large values in areas with fiber curving and crossing. 

Note that the proposed DFA and its related mathematical tools can be used not only for diffusion MRI data, but also for general director data. 
Moreover, there are many applications in which vector fields, like velocity fields, can be easily processed by using the modified DFA described in Section~\ref{sec:DFA_vector} to analyze their spatial features.  
Considering the proposed scalar indices are sensitive to different distortions of principal directions, 
these indices have potential in voxel-based analysis and tract-based analysis for group studies and longitudinal studies~\citep{smith:TBSS:NI2006,liu:NI2013}, 
which is a goal of future work.  
We will release the related codes and demos for DFA in DMRITool~\footnote{\label{fn:dmritool}\url{https://diffusionmritool.github.io}}, which is an open source toolbox for diffusion MRI data processing.

\section*{Acknowledgments}
The authors thank Dr.~Carlo Pierpaoli and Dr.~Elizabeth Hutchinson for useful discussions on the Fornix, and Ms. Liz Salak for editing the manuscript. 
This work was supported by funds provided by the Intramural Research Program of the \emph{Eunice Kennedy Shriver} National Institute of  Child  Health  and  Human  Development  (NICHD)  (ZIA-HD000266). 
The data were provided in part by the Human Connectome Project, WU-Minn Consortium 
(Principal Investigators: David Van Essen and Kamil Ugurbil; 1U54MH091657) funded by the 16 NIH institutes and centers that support the NIH Blueprint for Neuroscience Research; 
and by the McDonnell Center for Systems Neuroscience at Washington University.


\begin{thebibliography}{45}
\expandafter\ifx\csname natexlab\endcsname\relax\def\natexlab#1{#1}\fi
\providecommand{\url}[1]{\texttt{#1}}
\providecommand{\href}[2]{#2}
\providecommand{\path}[1]{#1}
\providecommand{\DOIprefix}{doi:}
\providecommand{\ArXivprefix}{arXiv:}
\providecommand{\URLprefix}{URL: }
\providecommand{\Pubmedprefix}{pmid:}
\providecommand{\doi}[1]{\href{http://dx.doi.org/#1}{\path{#1}}}
\providecommand{\Pubmed}[1]{\href{pmid:#1}{\path{#1}}}
\providecommand{\bibinfo}[2]{#2}
\ifx\xfnm\relax \def\xfnm[#1]{\unskip,\space#1}\fi
\bibitem[{Andrienko(2006)}]{andrienko_2006}
\bibinfo{author}{Andrienko, D.}, \bibinfo{year}{2006}.
\newblock \bibinfo{title}{Introduction to liquid crystals}.
\newblock \bibinfo{journal}{IMPRS school, Bad Marienberg} .
\bibitem[{Basser(1997)}]{basser_ANYAS1997}
\bibinfo{author}{Basser, P.J.}, \bibinfo{year}{1997}.
\newblock \bibinfo{title}{New histological and physiological stains derived
  from diffusion-tensor mr images}.
\newblock \bibinfo{journal}{Annals of the New York Academy of Sciences}
  \bibinfo{volume}{820}, \bibinfo{pages}{123--138}.
\bibitem[{Basser(2002)}]{basser_MRM2002}
\bibinfo{author}{Basser, P.J.}, \bibinfo{year}{2002}.
\newblock \bibinfo{title}{{Relationships between diffusion tensor and q-space
  MRI}}.
\newblock \bibinfo{journal}{Magnetic resonance in medicine}
  \bibinfo{volume}{47}, \bibinfo{pages}{392--397}.
\bibitem[{Basser et~al.(1994)Basser, Mattiello and LeBihan}]{Basser1994}
\bibinfo{author}{Basser, P.J.}, \bibinfo{author}{Mattiello, J.},
  \bibinfo{author}{LeBihan, D.}, \bibinfo{year}{1994}.
\newblock \bibinfo{title}{{MR} diffusion tensor spectroscropy and imaging}.
\newblock \bibinfo{journal}{Biophysical Journal} \bibinfo{volume}{66},
  \bibinfo{pages}{259--267}.
\bibitem[{Batchelor et~al.(2006)Batchelor, Calamante, Tournier, Atkinson, Hill
  and Connelly}]{batchelor_MRM2006}
\bibinfo{author}{Batchelor, P.}, \bibinfo{author}{Calamante, F.},
  \bibinfo{author}{Tournier, J.D.}, \bibinfo{author}{Atkinson, D.},
  \bibinfo{author}{Hill, D.}, \bibinfo{author}{Connelly, A.},
  \bibinfo{year}{2006}.
\newblock \bibinfo{title}{Quantification of the shape of fiber tracts}.
\newblock \bibinfo{journal}{Magnetic Resonance in Medicine}
  \bibinfo{volume}{55}, \bibinfo{pages}{894--903}.
\bibitem[{Cheng et~al.(2013)Cheng, Deriche, Jiang, Shen and
  Yap}]{JianCheng_NNSD_ISBI13}
\bibinfo{author}{Cheng, J.}, \bibinfo{author}{Deriche, R.},
  \bibinfo{author}{Jiang, T.}, \bibinfo{author}{Shen, D.},
  \bibinfo{author}{Yap, P.T.}, \bibinfo{year}{2013}.
\newblock \bibinfo{title}{{Non-Local Non-Negative Spherical Deconvolution for
  Single and Multiple Shell Diffusion MRI}}, in: \bibinfo{booktitle}{HARDI
  Reconstruction Challenge, International Symposium on Biomedical Imaging
  (ISBI'13)}.
\bibitem[{Cheng et~al.(2014)Cheng, Deriche, Jiang, Shen and Yap}]{cheng_NI2014}
\bibinfo{author}{Cheng, J.}, \bibinfo{author}{Deriche, R.},
  \bibinfo{author}{Jiang, T.}, \bibinfo{author}{Shen, D.},
  \bibinfo{author}{Yap, P.T.}, \bibinfo{year}{2014}.
\newblock \bibinfo{title}{{Non-Negative Spherical Deconvolution (NNSD) for
  estimation of fiber Orientation Distribution Function in single-/multi-shell
  diffusion MRI}}.
\newblock \bibinfo{journal}{NeuroImage} \bibinfo{volume}{101},
  \bibinfo{pages}{750--764}.
\newblock \DOIprefix\doi{10.1016/j.neuroimage.2014.07.062}.
\bibitem[{Cheng et~al.(2010)Cheng, Ghosh, Jiang and
  Deriche}]{Cheng_PDF_MICCAI2010}
\bibinfo{author}{Cheng, J.}, \bibinfo{author}{Ghosh, A.},
  \bibinfo{author}{Jiang, T.}, \bibinfo{author}{Deriche, R.},
  \bibinfo{year}{2010}.
\newblock \bibinfo{title}{{Model-free and Analytical {EAP} Reconstruction via
  Spherical Polar Fourier Diffusion MRI}}, in: \bibinfo{booktitle}{Medical
  Image Computing and Computer-Assisted Intervention (MICCAI'10)}, pp.
  \bibinfo{pages}{590--597}.
\newblock \DOIprefix\doi{10.1007/978-3-642-15705-9_72}.
\bibitem[{Cheng et~al.(2015)Cheng, Shen, Yap and Basser}]{cheng_MICCAI2015}
\bibinfo{author}{Cheng, J.}, \bibinfo{author}{Shen, D.}, \bibinfo{author}{Yap,
  P.T.}, \bibinfo{author}{Basser, P.J.}, \bibinfo{year}{2015}.
\newblock \bibinfo{title}{Tensorial spherical polar fourier diffusion mri with
  optimal dictionary learning}, in: \bibinfo{booktitle}{International
  Conference on Medical Image Computing and Computer-Assisted Intervention
  (MICCAI'15)}, \bibinfo{organization}{Springer}. pp.
  \bibinfo{pages}{174--182}.
\newblock \DOIprefix\doi{10.1007/978-3-319-24553-9_22}.
\bibitem[{Descoteaux et~al.(2007)Descoteaux, Angelino, Fitzgibbons and
  Deriche}]{Descoteaux2007}
\bibinfo{author}{Descoteaux, M.}, \bibinfo{author}{Angelino, E.},
  \bibinfo{author}{Fitzgibbons, S.}, \bibinfo{author}{Deriche, R.},
  \bibinfo{year}{2007}.
\newblock \bibinfo{title}{{Regularized, Fast and Robust Analytical Q-ball
  Imaging}}.
\newblock \bibinfo{journal}{Magnetic Resonance in Medicine}
  \bibinfo{volume}{58}, \bibinfo{pages}{497--510}.
\bibitem[{Duits and Franken(2009)}]{duits:2009}
\bibinfo{author}{Duits, R.}, \bibinfo{author}{Franken, E.},
  \bibinfo{year}{2009}.
\newblock \bibinfo{title}{Left-invariant diffusions on r3 s2 and their
  application to crossingpreserving smoothing on hardi-images}.
\newblock \bibinfo{journal}{CASA report, TU/e} , \bibinfo{pages}{23--27}.
\bibitem[{Duits and Franken(2011)}]{duits:IJCV2011}
\bibinfo{author}{Duits, R.}, \bibinfo{author}{Franken, E.},
  \bibinfo{year}{2011}.
\newblock \bibinfo{title}{Left-invariant diffusions on the space of positions
  and orientations and their application to crossing-preserving smoothing of
  hardi images}.
\newblock \bibinfo{journal}{International Journal of Computer Vision}
  \bibinfo{volume}{92}, \bibinfo{pages}{231--264}.
\bibitem[{Frank(2002)}]{frank_MRM2002}
\bibinfo{author}{Frank, L.R.}, \bibinfo{year}{2002}.
\newblock \bibinfo{title}{Characterization of anisotropy in high angular
  resolution diffusion-weighted mri}.
\newblock \bibinfo{journal}{Magnetic Resonance in Medicine}
  \bibinfo{volume}{47}, \bibinfo{pages}{1083--1099}.
\bibitem[{Helmer et~al.(2003)Helmer, Meiler, Sotak and
  Petruccelli}]{helmer_MRM2003}
\bibinfo{author}{Helmer, K.G.}, \bibinfo{author}{Meiler, M.R.},
  \bibinfo{author}{Sotak, C.H.}, \bibinfo{author}{Petruccelli, J.D.},
  \bibinfo{year}{2003}.
\newblock \bibinfo{title}{Comparison of the return-to-the-origin probability
  and the apparent diffusion coefficient of water as indicators of necrosis in
  rif-1 tumors}.
\newblock \bibinfo{journal}{Magnetic resonance in medicine}
  \bibinfo{volume}{49}, \bibinfo{pages}{468--478}.
\bibitem[{Johansen-Berg and Behrens(2009)}]{dMRI_book2009}
\bibinfo{author}{Johansen-Berg, H.}, \bibinfo{author}{Behrens, T.E.},
  \bibinfo{year}{2009}.
\newblock \bibinfo{title}{{Diffusion MRI: From quantitative measurement to In
  vivo neuroanatomy}}.
\newblock \bibinfo{publisher}{Elsevier}.
\bibitem[{Kindlmann(2004)}]{kindlmann_2004}
\bibinfo{author}{Kindlmann, G.}, \bibinfo{year}{2004}.
\newblock \bibinfo{title}{Superquadric tensor glyphs}, in:
  \bibinfo{booktitle}{Proceedings of the Sixth Joint Eurographics-IEEE TCVG
  conference on Visualization}, \bibinfo{organization}{Eurographics
  Association}. pp. \bibinfo{pages}{147--154}.
\bibitem[{Kindlmann et~al.(2007)Kindlmann, Ennis, Whitaker and
  Westin}]{Kindlmann_TMI2007}
\bibinfo{author}{Kindlmann, G.}, \bibinfo{author}{Ennis, D.B.},
  \bibinfo{author}{Whitaker, R.T.}, \bibinfo{author}{Westin, C.F.},
  \bibinfo{year}{2007}.
\newblock \bibinfo{title}{{Diffusion Tensor Analysis with Invariant Gradients
  and Rotation Tangents}}.
\newblock \bibinfo{journal}{IEEE Transactions on Medical Imaging}
  \bibinfo{volume}{26(11)}, \bibinfo{pages}{1483--1499}.
\bibitem[{Lasi{\v{c}} et~al.(2014)Lasi{\v{c}}, Szczepankiewicz, Eriksson,
  Nilsson and Topgaard}]{lasivc:FrontiersInPhysics2014}
\bibinfo{author}{Lasi{\v{c}}, S.}, \bibinfo{author}{Szczepankiewicz, F.},
  \bibinfo{author}{Eriksson, S.}, \bibinfo{author}{Nilsson, M.},
  \bibinfo{author}{Topgaard, D.}, \bibinfo{year}{2014}.
\newblock \bibinfo{title}{Microanisotropy imaging: quantification of
  microscopic diffusion anisotropy and orientational order parameter by
  diffusion mri with magic-angle spinning of the q-vector}.
\newblock \bibinfo{journal}{Frontiers in Physics} \bibinfo{volume}{2},
  \bibinfo{pages}{11}.
\bibitem[{Lessig et~al.(2012)Lessig, de~Witt and Fiume}]{lessig_2012}
\bibinfo{author}{Lessig, C.}, \bibinfo{author}{de~Witt, T.},
  \bibinfo{author}{Fiume, E.}, \bibinfo{year}{2012}.
\newblock \bibinfo{title}{Efficient and accurate rotation of finite spherical
  harmonics expansions}.
\newblock \bibinfo{journal}{Journal of Computational Physics}
  \bibinfo{volume}{231}, \bibinfo{pages}{243--250}.
\bibitem[{Liu et~al.(2013)Liu, Sachdev, Lipnicki, Jiang, Geng, Zhu, Reppermund,
  Tao, Trollor, Brodaty et~al.}]{liu:NI2013}
\bibinfo{author}{Liu, T.}, \bibinfo{author}{Sachdev, P.S.},
  \bibinfo{author}{Lipnicki, D.M.}, \bibinfo{author}{Jiang, J.},
  \bibinfo{author}{Geng, G.}, \bibinfo{author}{Zhu, W.},
  \bibinfo{author}{Reppermund, S.}, \bibinfo{author}{Tao, D.},
  \bibinfo{author}{Trollor, J.N.}, \bibinfo{author}{Brodaty, H.}, et~al.,
  \bibinfo{year}{2013}.
\newblock \bibinfo{title}{Limited relationships between two-year changes in
  sulcal morphology and other common neuroimaging indices in the elderly}.
\newblock \bibinfo{journal}{NeuroImage} \bibinfo{volume}{83},
  \bibinfo{pages}{12--17}.
\bibitem[{Michailovich et~al.(2011)Michailovich, Rathi and
  Dolui}]{michailovich:TMI2011}
\bibinfo{author}{Michailovich, O.}, \bibinfo{author}{Rathi, Y.},
  \bibinfo{author}{Dolui, S.}, \bibinfo{year}{2011}.
\newblock \bibinfo{title}{Spatially regularized compressed sensing for high
  angular resolution diffusion imaging}.
\newblock \bibinfo{journal}{IEEE transactions on medical imaging}
  \bibinfo{volume}{30}, \bibinfo{pages}{1100--1115}.
\bibitem[{Moakher(2002)}]{moakher_SIAM2002}
\bibinfo{author}{Moakher, M.}, \bibinfo{year}{2002}.
\newblock \bibinfo{title}{Means and averaging in the group of rotations}.
\newblock \bibinfo{journal}{SIAM journal on matrix analysis and applications}
  \bibinfo{volume}{24}, \bibinfo{pages}{1--16}.
\bibitem[{Moakher(2009)}]{moakher_2009}
\bibinfo{author}{Moakher, M.}, \bibinfo{year}{2009}.
\newblock \bibinfo{title}{The algebra of fourth-order tensors with application
  to diffusion MRI}. \bibinfo{publisher}{Springer}.
\newblock pp. \bibinfo{pages}{57--80}.
\bibitem[{{\"O}zarslan et~al.(2013){\"O}zarslan, Koay, Shepherd, Komlosh,
  {\.I}rfano{\u{g}}lu, Pierpaoli and Basser}]{ozarslan_NI13}
\bibinfo{author}{{\"O}zarslan, E.}, \bibinfo{author}{Koay, C.G.},
  \bibinfo{author}{Shepherd, T.M.}, \bibinfo{author}{Komlosh, M.E.},
  \bibinfo{author}{{\.I}rfano{\u{g}}lu, M.O.}, \bibinfo{author}{Pierpaoli, C.},
  \bibinfo{author}{Basser, P.J.}, \bibinfo{year}{2013}.
\newblock \bibinfo{title}{Mean apparent propagator (map) mri: a novel diffusion
  imaging method for mapping tissue microstructure}.
\newblock \bibinfo{journal}{NeuroImage} \bibinfo{volume}{78},
  \bibinfo{pages}{16--32}.
\bibitem[{Pajevic et~al.(2002)Pajevic, Aldroubi and Basser}]{pajevic_JMR2002}
\bibinfo{author}{Pajevic, S.}, \bibinfo{author}{Aldroubi, A.},
  \bibinfo{author}{Basser, P.J.}, \bibinfo{year}{2002}.
\newblock \bibinfo{title}{A continuous tensor field approximation of discrete
  dt-mri data for extracting microstructural and architectural features of
  tissue}.
\newblock \bibinfo{journal}{Journal of magnetic resonance}
  \bibinfo{volume}{154}, \bibinfo{pages}{85--100}.
\bibitem[{Pierpaoli and Basser(1996)}]{Pierpaoli1996}
\bibinfo{author}{Pierpaoli, C.}, \bibinfo{author}{Basser, P.},
  \bibinfo{year}{1996}.
\newblock \bibinfo{title}{{Toward a Quantitative Assessment of Diffusion
  Anisotropy}}.
\newblock \bibinfo{journal}{Magnetic Resonance in Medicine}
  \bibinfo{volume}{36}, \bibinfo{pages}{893--906}.
\bibitem[{Piuze et~al.(2015)Piuze, Sporring and Siddiqi}]{piuze_PAMI2015}
\bibinfo{author}{Piuze, E.}, \bibinfo{author}{Sporring, J.},
  \bibinfo{author}{Siddiqi, K.}, \bibinfo{year}{2015}.
\newblock \bibinfo{title}{{Maurer-Cartan forms for fields on surfaces:
  application to heart fiber geometry}}.
\newblock \bibinfo{journal}{IEEE transactions on pattern analysis and machine
  intelligence} \bibinfo{volume}{37}, \bibinfo{pages}{2492--2504}.
\bibitem[{Portegies et~al.(2015)Portegies, Fick, Sanguinetti, Meesters, Girard
  and Duits}]{portegies:PLOSOne2015}
\bibinfo{author}{Portegies, J.M.}, \bibinfo{author}{Fick, R.H.J.},
  \bibinfo{author}{Sanguinetti, G.R.}, \bibinfo{author}{Meesters, S.P.},
  \bibinfo{author}{Girard, G.}, \bibinfo{author}{Duits, R.},
  \bibinfo{year}{2015}.
\newblock \bibinfo{title}{{Improving fiber alignment in HARDI by combining
  contextual PDE flow with constrained spherical deconvolution}}.
\newblock \bibinfo{journal}{PloS one} \bibinfo{volume}{10},
  \bibinfo{pages}{e0138122}.
\bibitem[{Reisert and Kiselev(2011)}]{reisert_TMI11}
\bibinfo{author}{Reisert, M.}, \bibinfo{author}{Kiselev, V.},
  \bibinfo{year}{2011}.
\newblock \bibinfo{title}{{Fiber continuity: an anisotropic prior for ODF
  estimation}}.
\newblock \bibinfo{journal}{IEEE transactions on medical imaging}
  \bibinfo{volume}{30}, \bibinfo{pages}{1274}.
\bibitem[{Savadjiev et~al.(2010)Savadjiev, Kindlmann, Bouix, Shenton and
  Westin}]{Savadjiev_NI2010}
\bibinfo{author}{Savadjiev, P.}, \bibinfo{author}{Kindlmann, G.},
  \bibinfo{author}{Bouix, S.}, \bibinfo{author}{Shenton, M.},
  \bibinfo{author}{Westin, C.F.}, \bibinfo{year}{2010}.
\newblock \bibinfo{title}{{Local white matter geometry from diffusion tensor
  gradients}}.
\newblock \bibinfo{journal}{NeuroImage} \bibinfo{volume}{49(4)},
  \bibinfo{pages}{3175--3186}.
\bibitem[{Savadjiev et~al.(2007)Savadjiev, Zucker and
  Siddiqi}]{savadjiev_ICCV2007}
\bibinfo{author}{Savadjiev, P.}, \bibinfo{author}{Zucker, S.W.},
  \bibinfo{author}{Siddiqi, K.}, \bibinfo{year}{2007}.
\newblock \bibinfo{title}{On the differential geometry of 3d flow patterns:
  Generalized helicoids and diffusion mri analysis}, in:
  \bibinfo{booktitle}{2007 IEEE 11th International Conference on Computer
  Vision}, \bibinfo{organization}{IEEE}. pp. \bibinfo{pages}{1--8}.
\bibitem[{Scherrer et~al.(2015)Scherrer, Schwartzman, Taquet, Sahin, Prabhu and
  Warfield}]{scherrer:MRM2015}
\bibinfo{author}{Scherrer, B.}, \bibinfo{author}{Schwartzman, A.},
  \bibinfo{author}{Taquet, M.}, \bibinfo{author}{Sahin, M.},
  \bibinfo{author}{Prabhu, S.P.}, \bibinfo{author}{Warfield, S.K.},
  \bibinfo{year}{2015}.
\newblock \bibinfo{title}{Characterizing brain tissue by assessment of the
  distribution of anisotropic microstructural environments in
  diffusion-compartment imaging (diamond)}.
\newblock \bibinfo{journal}{Magnetic resonance in medicine} .
\bibitem[{Smith et~al.(2006)Smith, Jenkinson, Johansen-Berg, Rueckert, Nichols,
  Mackay, Watkins, Ciccarelli, Cader, Matthews et~al.}]{smith:TBSS:NI2006}
\bibinfo{author}{Smith, S.M.}, \bibinfo{author}{Jenkinson, M.},
  \bibinfo{author}{Johansen-Berg, H.}, \bibinfo{author}{Rueckert, D.},
  \bibinfo{author}{Nichols, T.E.}, \bibinfo{author}{Mackay, C.E.},
  \bibinfo{author}{Watkins, K.E.}, \bibinfo{author}{Ciccarelli, O.},
  \bibinfo{author}{Cader, M.Z.}, \bibinfo{author}{Matthews, P.M.}, et~al.,
  \bibinfo{year}{2006}.
\newblock \bibinfo{title}{Tract-based spatial statistics: voxelwise analysis of
  multi-subject diffusion data}.
\newblock \bibinfo{journal}{Neuroimage} \bibinfo{volume}{31},
  \bibinfo{pages}{1487--1505}.
\bibitem[{Sotiropoulos et~al.(2013)Sotiropoulos, Jbabdi, Xu, Andersson,
  Moeller, Auerbach, Glasser, Hernandez, Sapiro, Jenkinson
  et~al.}]{sotiropoulos_HCP_NI13}
\bibinfo{author}{Sotiropoulos, S.N.}, \bibinfo{author}{Jbabdi, S.},
  \bibinfo{author}{Xu, J.}, \bibinfo{author}{Andersson, J.L.},
  \bibinfo{author}{Moeller, S.}, \bibinfo{author}{Auerbach, E.J.},
  \bibinfo{author}{Glasser, M.F.}, \bibinfo{author}{Hernandez, M.},
  \bibinfo{author}{Sapiro, G.}, \bibinfo{author}{Jenkinson, M.}, et~al.,
  \bibinfo{year}{2013}.
\newblock \bibinfo{title}{{Advances in diffusion MRI acquisition and processing
  in the Human Connectome Project}}.
\newblock \bibinfo{journal}{NeuroImage} \bibinfo{volume}{80},
  \bibinfo{pages}{125--43}.
\bibitem[{Szczepankiewicz et~al.(2015)Szczepankiewicz, Lasi{\v{c}}, van Westen,
  Sundgren, Englund, Westin, St{\aa}hlberg, L{\"a}tt, Topgaard and
  Nilsson}]{szczepankiewicz:NI2015}
\bibinfo{author}{Szczepankiewicz, F.}, \bibinfo{author}{Lasi{\v{c}}, S.},
  \bibinfo{author}{van Westen, D.}, \bibinfo{author}{Sundgren, P.C.},
  \bibinfo{author}{Englund, E.}, \bibinfo{author}{Westin, C.F.},
  \bibinfo{author}{St{\aa}hlberg, F.}, \bibinfo{author}{L{\"a}tt, J.},
  \bibinfo{author}{Topgaard, D.}, \bibinfo{author}{Nilsson, M.},
  \bibinfo{year}{2015}.
\newblock \bibinfo{title}{Quantification of microscopic diffusion anisotropy
  disentangles effects of orientation dispersion from microstructure:
  applications in healthy volunteers and in brain tumors}.
\newblock \bibinfo{journal}{NeuroImage} \bibinfo{volume}{104},
  \bibinfo{pages}{241--252}.
\bibitem[{Tariq et~al.(2016)Tariq, Schneider, Alexander, Wheeler-Kingshott and
  Zhang}]{tariq:NI2016}
\bibinfo{author}{Tariq, M.}, \bibinfo{author}{Schneider, T.},
  \bibinfo{author}{Alexander, D.C.}, \bibinfo{author}{Wheeler-Kingshott,
  C.A.G.}, \bibinfo{author}{Zhang, H.}, \bibinfo{year}{2016}.
\newblock \bibinfo{title}{Bingham--noddi: Mapping anisotropic orientation
  dispersion of neurites using diffusion mri}.
\newblock \bibinfo{journal}{NeuroImage} \bibinfo{volume}{133},
  \bibinfo{pages}{207--223}.
\bibitem[{Tax et~al.(2016)Tax, Haije, Fuster, Westin, Viergever, Florack and
  Leemans}]{tax_NI2016}
\bibinfo{author}{Tax, C.M.}, \bibinfo{author}{Haije, T.D.},
  \bibinfo{author}{Fuster, A.}, \bibinfo{author}{Westin, C.F.},
  \bibinfo{author}{Viergever, M.A.}, \bibinfo{author}{Florack, L.},
  \bibinfo{author}{Leemans, A.}, \bibinfo{year}{2016}.
\newblock \bibinfo{title}{Sheet probability index (spi): Characterizing the
  geometrical organization of the white matter with diffusion mri}.
\newblock \bibinfo{journal}{NeuroImage} \bibinfo{volume}{142},
  \bibinfo{pages}{260--279}.
\bibitem[{Tournier et~al.(2007)Tournier, Calamante and
  Connelly}]{tournier_NI2007}
\bibinfo{author}{Tournier, J.D.}, \bibinfo{author}{Calamante, F.},
  \bibinfo{author}{Connelly, A.}, \bibinfo{year}{2007}.
\newblock \bibinfo{title}{{Robust determination of the fibre orientation
  distribution in diffusion MRI: non-negativity constrained super-resolved
  spherical deconvolution}}.
\newblock \bibinfo{journal}{NeuroImage} \bibinfo{volume}{35},
  \bibinfo{pages}{1459--1472}.
\bibitem[{Tournier et~al.(2012)Tournier, Calamante and
  Connelly}]{tournier_mrtrix_12}
\bibinfo{author}{Tournier, J.D.}, \bibinfo{author}{Calamante, F.},
  \bibinfo{author}{Connelly, A.}, \bibinfo{year}{2012}.
\newblock \bibinfo{title}{{MRtrix: Diffusion tractography in crossing fiber
  regions}}.
\newblock \bibinfo{journal}{International Journal of Imaging Systems and
  Technology} \bibinfo{volume}{22}, \bibinfo{pages}{53--66}.
\bibitem[{Tournier et~al.(2004)Tournier, Calamante, Gadian and
  Connelly}]{TournierNI2004}
\bibinfo{author}{Tournier, J.D.}, \bibinfo{author}{Calamante, F.},
  \bibinfo{author}{Gadian, D.}, \bibinfo{author}{Connelly, A.},
  \bibinfo{year}{2004}.
\newblock \bibinfo{title}{{direct estimation of the fiber orientation density
  function from diffusion-weighted MRI data using spherical deconvolution}}.
\newblock \bibinfo{journal}{NeuroImage} \bibinfo{volume}{23},
  \bibinfo{pages}{1176--1185}.
\bibitem[{Tuch(2004)}]{Tuch2004}
\bibinfo{author}{Tuch, D.S.}, \bibinfo{year}{2004}.
\newblock \bibinfo{title}{{Q-ball imaging}}.
\newblock \bibinfo{journal}{Magnetic Resonance in Medicine}
  \bibinfo{volume}{52}, \bibinfo{pages}{1358--1372}.
\bibitem[{Tuch et~al.(2002)Tuch, Reese, Wiegell, Makris, Belliveau and
  Wedeen}]{TuchMRM2002}
\bibinfo{author}{Tuch, D.S.}, \bibinfo{author}{Reese, T.G.},
  \bibinfo{author}{Wiegell, M.R.}, \bibinfo{author}{Makris, N.},
  \bibinfo{author}{Belliveau, J.W.}, \bibinfo{author}{Wedeen, V.J.},
  \bibinfo{year}{2002}.
\newblock \bibinfo{title}{{High Angular Resolution Diffusion Imaging Reveals
  Intravoxel White Matter Fiber Heterogeneity}}.
\newblock \bibinfo{journal}{Magnetic Resonance in Medicine}
  \bibinfo{volume}{48}, \bibinfo{pages}{577--582}.
\bibitem[{Van~Essen et~al.(2013)Van~Essen, Smith, Barch, Behrens, Yacoub,
  Ugurbil, Consortium et~al.}]{van:NI2013:HCP}
\bibinfo{author}{Van~Essen, D.C.}, \bibinfo{author}{Smith, S.M.},
  \bibinfo{author}{Barch, D.M.}, \bibinfo{author}{Behrens, T.E.},
  \bibinfo{author}{Yacoub, E.}, \bibinfo{author}{Ugurbil, K.},
  \bibinfo{author}{Consortium, W.M.H.}, et~al., \bibinfo{year}{2013}.
\newblock \bibinfo{title}{The wu-minn human connectome project: an overview}.
\newblock \bibinfo{journal}{Neuroimage} \bibinfo{volume}{80},
  \bibinfo{pages}{62--79}.
\bibitem[{Wu and Alexander(2007)}]{Wu_NI2007}
\bibinfo{author}{Wu, Y.C.}, \bibinfo{author}{Alexander, A.L.},
  \bibinfo{year}{2007}.
\newblock \bibinfo{title}{{Hybrid diffusion imaging}}.
\newblock \bibinfo{journal}{NeuroImage} \bibinfo{volume}{36},
  \bibinfo{pages}{617--629}.
\bibitem[{Zhang et~al.(2012)Zhang, Schneider, Wheeler-Kingshott and
  Alexander}]{zhang_NODDI_NI2012}
\bibinfo{author}{Zhang, H.}, \bibinfo{author}{Schneider, T.},
  \bibinfo{author}{Wheeler-Kingshott, C.A.}, \bibinfo{author}{Alexander, D.C.},
  \bibinfo{year}{2012}.
\newblock \bibinfo{title}{{NODDI: practical in vivo neurite orientation
  dispersion and density imaging of the human brain}}.
\newblock \bibinfo{journal}{Neuroimage} \bibinfo{volume}{61},
  \bibinfo{pages}{1000--1016}.

\end{thebibliography}
\end{document}